\documentclass[10pt]{article}

\usepackage[preprint]{tmlr}

\usepackage{bbm}
\usepackage{amssymb}   %
\usepackage{subcaption}
\usepackage{xcolor}
\usepackage{mattsstyle}
\usepackage{mathtools}  %

\usepackage{booktabs}
\usepackage{tabularx}
\usepackage{array}
\usepackage{longtable}
\usepackage{makecell}
\usepackage{enumitem}

\usepackage{tikz}
\usetikzlibrary{arrows.meta, positioning, shapes.geometric, matrix, backgrounds, fit}

\usepackage{placeins}

\definecolor{darkblue}{RGB}{0,51,102}

\newcommand{\KL}{\mathrm{KL}}
\newcommand{\E}{\mathbb{E}}
\newcommand{\R}{\mathbb{R}}

\newcommand{\Wass}{\mathsf{W}}
\def\Wassp#1{\Wass_{#1}}
\newcommand{\Wone}{\Wassp{1}}
\newcommand{\Wtwo}{\Wassp{2}}
\newcommand{\Sink}{\mathsf{S}}
\def\Seps#1{\Sink_{#1}}
\newcommand{\SW}{\mathsf{SW}}
\def\SWp#1{\SW_{#1}}
\definecolor{darkblue}{rgb}{0.1,0.1,0.6}

\title{Optimal Transport Meets Reinforcement Learning: A Survey}

\author{\name Yujie Zhu \email yujie.zhu@warwick.ac.uk \\
        \addr University of Warwick
        \AND
        \name Charles A. Hepburn \email charles.hepburn@warwick.ac.uk \\
        \addr University of Warwick
        \AND
        \name Matthew Thorpe \email matthew.thorpe@warwick.ac.uk \\
        \addr University of Warwick
        \AND
        \name Giovanni Montana \email g.montana@warwick.ac.uk \\
        \addr University of Warwick}

\begin{document}
\maketitle

\begin{abstract}
Reinforcement learning (RL) algorithms frequently compare probability distributions, such as state visitation distributions induced by policies and experts, action distributions from learned policies and offline datasets, or transition distributions from learned models and environments. However, commonly used divergences may become ineffective when these distributions overlap weakly, which is frequently encountered in imitation learning, offline RL, and deployment under distribution shift. Optimal transport (OT) offers an alternative by measuring the cost of \emph{moving} probability mass from one distribution to another under a ground cost that encodes task geometry. This survey covers how OT is used inside RL objectives and algorithms. For each method, we identify: the role OT plays, the distributions compared, the OT formulation used, and the treatment of temporal structure. Beyond categorising existing methods, we discuss the motivations behind different OT choices, practical considerations such as cost design and computational challenges, and highlight open problems including scalable trajectory-level transport, principled handling of mass mismatch, and theoretical analysis for OT-regularised RL.
\end{abstract}

\section{Introduction}

Reinforcement learning (RL) studies how an agent should act in an uncertain environment to maximise long-term reward. The agent interacts with the environment, observes states, takes actions, receives rewards, and updates its behaviour \citep{sutton1998reinforcement}. This framework covers problems from robotics and control \citep{kober2013reinforcement, schulman2017proximal, lillicrap2015continuous} to resource allocation \citep{ye2019deep, luong2019applications, cui2019multi} and dialogue systems \citep{li2016deep, su2016line, papangelis2019collaborative}, but it also exposes recurring difficulties: limited data \citep{levine2020offline, sun2023less, zhu2025uncertainty}, compounding error over time \citep{talvitie2017self, janner2019trust, jiang2024note}, and the need to generalise under distribution shift \citep{cobbe2019quantifying, wang2020improving, fujimoto2024assessing}.

Although RL is often presented as expected-return maximisation, many RL algorithms are really comparing and shaping \emph{distributions} \citep{zhang2020wasserstein, papagiannis2022imitation, dadashi2020primal}. For example, a policy induces a distribution over visited states and actions, namely its \emph{occupancy measure}. Informally, this is the policy's statistical ``footprint'': where it tends to spend time, and which actions it tends to take there. Several familiar RL problems can be phrased in terms of distribution comparisons. In \textcolor{black}{imitation learning (IL)}, the question is how far the learner's footprint is from the expert's \citep{xiao2019wasserstein, luo2023optimal, chang2023imitation}. In policy optimisation, the learning stability is influenced by how far the learned policy drifts from the policy from the previous iteration \citep{song2023provably, pacchiano2020learning, pfau2025wasserstein}. In offline RL, the issue is how far the learned policy deviates from the behaviour policy that generated the data \citep{wu2019behavior, omura2025offline, asadulaev2024rethinking}. In robust control, one asks how much the transition law may move away from a nominal model \citep{abdullah2019wasserstein, hou2020robust, queeney2024optimal}. In curriculum learning, one needs a principled way to move from one task distribution to another \citep{klink2022curriculum, huang2022curriculum}. Across all of these settings, progress depends on having a discrepancy that says not only whether two distributions differ, but how they differ in ways that matter for control, where we use control broadly to refer to sequential decision-making through actions that influence future states and optimise long-term performance.

One choice is to measure discrepancy using $f$-divergences such as Kullback-Leibler (KL) or Jensen-Shannon (JS). These compare densities pointwise: at each state or action, they ask how the two probability densities differ. This works well when distributions overlap substantially, but becomes problematic when they do not \citep{birrell2022f}. In \textcolor{black}{IL}, for instance, early in training the learner often visits states the expert never visited. A JS discriminator may then saturate and provide almost no gradient, and a KL term may be undefined or dominated by regions where one density is near zero \citep{arjovsky2017towards}. The learning signal degrades precisely when guidance is most needed.

Optimal transport (OT) takes a different approach. Rather than comparing densities pointwise, OT asks: what is the minimum effort required to \emph{move} probability mass from one distribution to another? A useful mental picture is to imagine reshaping one pile of sand into another. The amount of work depends both on how much sand must be moved and on how far it must move. In OT, that movement is quantified by a \emph{ground cost}, which specifies how expensive it is to transport mass between any two points \citep{santambrogio2015optimal}. The key consequence is that OT remains informative even when distributions have disjoint support. If a learner state is far from all expert states, OT quantifies \emph{how far} in the geometry defined by the ground cost rather than returning an uninformative or saturated value. This geometric perspective also exposes a modelling lever absent from $f$-divergences: the ground cost encodes domain knowledge about what it means for two states, actions, or trajectories to be close. A robotic manipulation task might use kinematic distance; a navigation task might use geodesic distance; and a high-dimensional observation space might use distance in a learned feature embedding. The choice of cost directly shapes what the algorithm treats as similar or different.

These properties have led to OT being adopted across many areas of RL. In \textcolor{black}{IL}, Wasserstein distances match learner and expert occupancies, either adversarially through dual critics \citep{xiao2019wasserstein} or through transport plans that assign rewards offline \citep{luo2023optimal}. In policy optimisation, Wasserstein constraints replace KL-based trust regions so that small updates respect action geometry rather than only likelihood ratios \citep{song2023provably}. In robustness, Wasserstein balls around nominal dynamics define ambiguity sets for worst-case optimisation \citep{yang2017convex}. In world modelling, Wasserstein losses on one-step transitions can be related to worst-case value error over Lipschitz test functions \citep{asadi2018equivalence}. In transfer and curricula, OT, or Gromov-Wasserstein when spaces differ, aligns distributions across domains and interpolates between task distributions in a geometry-aware way \citep{liu2024pearl, klink2022curriculum}. In offline RL, transport-based constraints keep learned policies close to the behaviour policy that generated the data \citep{wu2019behavior}. The technical details differ from section to section, but the recurring theme is simple: OT gives RL algorithms a way to compare distributions using geometry without requiring overlapping support.

Despite this breadth, the literature is fragmented. Methods developed for imitation may not cite related ideas in robustness, and techniques for trust regions may not connect to distributional critics. Without a unifying view, it is easy to treat OT as a black-box distance and overlook the modelling choices that determine success or failure: a mis-specified ground cost, poorly scaled entropic regularisation, an unstable dual critic, or a coupling that ignores temporal structure. 

We searched Google Scholar, IEEE Xplore, and arXiv using combinations of keywords including optimal transport, Wasserstein distance, reinforcement learning, imitation learning, and policy optimisation. \textcolor{black}{The survey primarily covers work published or publicly available up to April 2026. The reference list was last updated in September 2026, with cited preprints replaced by their published versions where available.}
\textcolor{black}{
Our inclusion criterion is \emph{core OT-RL} work; a small number of enabling or adjacent contributions from outside RL are retained where they are directly used by the surveyed methods, and are marked as such rather than reviewed in their own right (for example domain-adaptation objectives in Section~\ref{sec:transfer} and generative or causal OT tools in Section~\ref{sec:other-apps}).} 
We focus on works in which OT plays a substantive methodological role in RL or \textcolor{black}{IL}, such as shaping rewards, serving as an objective or constraint, or underpinning an optimisation procedure, while generally excluding purely application-oriented studies and OT research outside the RL setting. 
Related topics such as goal transfer, domain adaptation, and data filtering (Section~\ref{sec:transfer}) are included as important pre- or post-components of the RL pipeline, while \textcolor{black}{reinforcement learning from human feedback (RLHF)} and contextual bandits (Section~\ref{sec:robust-rlhf},~\ref{sec:other-apps}) are briefly discussed as related sequential decision-making problems employing similar OT-based ideas that may inform future OT-based RL research. 
Accordingly, rather than aiming to provide an exhaustive catalogue of OT-related RL research, this survey highlights the diverse algorithmic roles of OT in RL and captures the evolution of the field.

\begin{figure}[t]
\centering
\begin{tikzpicture}[
    rlbox/.style={rectangle, draw, rounded corners=4pt, minimum width=3.0cm,
                  minimum height=1.2cm, align=center, font=\small,
                  fill=gray!8, line width=0.8pt},
    otbox/.style={rectangle, draw, rounded corners=4pt, minimum width=2.6cm,
                  minimum height=0.9cm, align=center, font=\small,
                  draw=darkblue, text=darkblue, fill=blue!5, line width=0.8pt},
    otarr/.style={-, semithick, darkblue, dashed},
]

\definecolor{darkblue}{RGB}{0,51,102}

\node[rlbox, minimum width=3.4cm, minimum height=1.4cm] (hub)
    {\textbf{RL Pipeline}\\[-1pt]{\scriptsize Policy $\pi$\;\;\;Critic $Q$\;\;\;Dynamics $P$}};

\node[otbox, above=1.2cm of hub] (imit)
    {Imitation\\[-1pt]{\scriptsize $\Wass(\rho_\pi,\,\rho_{\pi_E})$\quad\S\ref{sec:il}}};

\node[otbox, above right=0.5cm and 1.6cm of hub] (robust)
    {Robustness\\[-1pt]{\scriptsize $\Wass(P,\,P_0)\!\le\!\delta$\quad\S\ref{sec:robust}}};

\node[otbox, below right=0.5cm and 1.6cm of hub] (offline)
    {Offline constraints\\[-1pt]{\scriptsize $\Wass(\pi(\cdot|s),\,\pi_{\beta}(\cdot|s))$\quad\S\ref{sec:offline}}};

\node[otbox, below=1.2cm of hub] (model)
    {Model learning\\[-1pt]{\scriptsize $\Wass(P_\phi,\,P)$\quad\S\ref{sec:modelbased}}};

\node[otbox, below left=0.5cm and 1.6cm of hub] (distrib)
    {Distributional critics\\[-1pt]{\scriptsize $\Wass(Z^\pi, \mathcal{T}^{\pi}Z^\pi)$\quad\S\ref{sec:policy-critics}}};

\node[otbox, above left=0.5cm and 1.6cm of hub] (trust)
    {Trust regions\\[-1pt]{\scriptsize $\Wass(\pi(\cdot|s),\,\pi_{\mathrm{old}}(\cdot|s))$\quad\S\ref{sec:policy-trust}}};

\draw[otarr] (hub) -- (imit);
\draw[otarr] (hub) -- (robust);
\draw[otarr] (hub) -- (offline);
\draw[otarr] (hub) -- (model);
\draw[otarr] (hub) -- (distrib);
\draw[otarr] (hub) -- (trust);

\end{tikzpicture}
\caption{Where optimal transport enters the RL pipeline. Each spoke represents a distinct role that OT plays, with the representative objective and section reference. Here $\Wass$ denotes a generic OT-based discrepancy, which may differ across applications, e.g., $\Wone, \Wtwo$ or an entropy-regularised formulation. Additional roles not shown include safety constraints (Sections~\ref{sec:robust-visitation},~\ref{sec:robust-uncertainty}), preference alignment (Section~\ref{sec:robust-rlhf}), transfer and curricula (Sections~\ref{sec:transfer},~\ref{sec:curricula}), and multi-policy/multi-agent diversity (Section~\ref{sec:extensions}).
}
\label{fig:pipeline}
\end{figure}
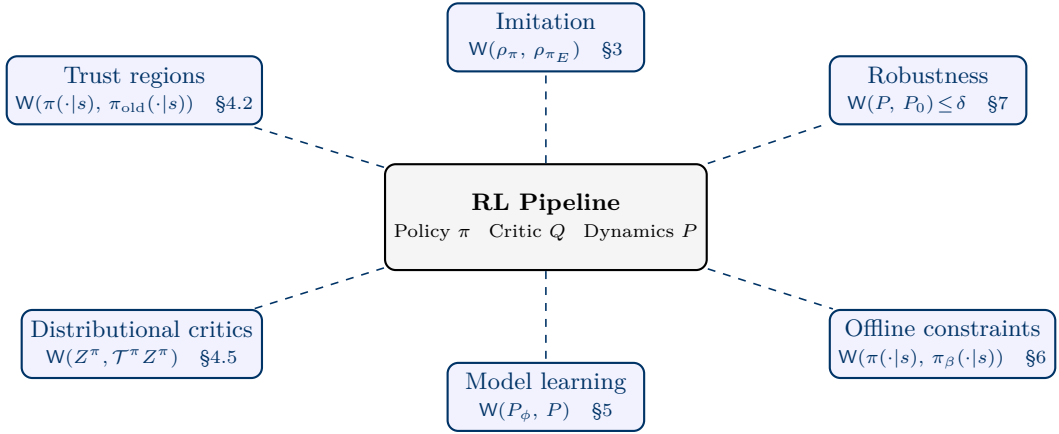

This survey therefore focuses on OT \emph{as it appears inside RL objectives and algorithms}. Throughout, we keep returning to four concrete questions: what OT is doing inside the objective, which distributions are compared, which OT formulation is used, and how temporal structure is handled. Those questions organise the survey (Figure~\ref{fig:pipeline})\textcolor{black}{\label{sec:intro-questions}} and make it easier to compare methods that are usually discussed in separate literatures. The recurring roles are reward signal, regulariser, trust-region constraint, ambiguity set, and model loss.

After preliminaries on RL and OT (Section~\ref{sec:preliminaries}), we cover imitation learning and reward inference (Section~\ref{sec:il}), policy optimisation and regularisation (Section~\ref{sec:policy}), model-based RL (Section~\ref{sec:modelbased}), offline RL (Section~\ref{sec:offline}), robustness, safety, and transfer (Section~\ref{sec:robust}), and other applications including multi-policy and multi-agent settings (Section~\ref{sec:extensions}). We conclude with practical guidance and open problems (Section~\ref{sec:discussion}).

\paragraph{Related surveys}
Compared with existing surveys on OT in machine learning \citep{montesuma2024recent, pereira2025survey}, which typically cover RL as only one of many application domains, our survey provides a dedicated and substantially more comprehensive overview of OT in RL. For example, \citet{montesuma2024recent} discusses RL briefly, focusing only on distributional RL, Bayesian RL, and policy optimisation with a single representative OT-based method in each category. In contrast, we review the diverse roles of OT across a broad range of RL settings, including imitation learning, policy optimisation, model-based RL, offline RL, robustness, transfer, and related applications. Existing survey efforts on OT within RL are limited. For instance, \citet{baheri2024synergy} focuses specifically on multi-agent RL and discusses the potential of OT at a high level, while our survey considers the broader RL landscape and categorises existing OT-based methods according to their algorithmic roles, with discussions of their underlying methodologies and algorithmic details.
Rather than serving solely as a catalogue of existing methods, our survey aims to help readers understand how OT is incorporated into RL algorithms, why different OT formulations are adopted, and when OT is beneficial. 
To this end, our survey primarily focuses on the methodological developments rather than empirical performance comparisons, as the latter are often highly dependent on the underlying environments, tasks, and experimental settings.
Accordingly, the survey is intended not only for researchers with expertise in RL or OT, but also for readers interested in the intersection of these two fields who have a basic mathematical background and seek a structured overview of the existing literature.

\section{Preliminaries}
\label{sec:preliminaries}

This section fixes the notation and core definitions used throughout the survey. Section~\ref{sec:rl-brief} introduces the Markov decision process framework, occupancy measures, and the distributional objects that OT will act on. Section~\ref{sec:ot} then reviews the OT concepts, including Wasserstein distances, entropic regularisation, Gromov-Wasserstein, and partial/unbalanced variants, that occur in later sections.

\subsection{Reinforcement learning in brief}
\label{sec:rl-brief}

Optimal transport enters RL through \emph{probability measures induced by a policy}: per-state action distributions $\pi(\cdot\mid s)$, discounted occupancies $\rho_\pi$, and trajectory laws $P_\pi$ or their pushforwards under an embedding map $\Phi$, i.e., the induced distributions after mapping trajectories through $\Phi$. This subsection fixes notation for these objects and gives the minimum intuition needed for the later sections. The key idea is simple: OT does not act directly on policy parameters, but on the distributions the policy generates, that is, \emph{where the policy places mass} and \emph{how that mass moves through the dynamics}.

\citet{sutton1998reinforcement} provides the foundation for RL. The environment is usually modelled as a discounted Markov decision process (MDP)
\[
\mathcal{M}=(\mathcal{S},\mathcal{A},P,R,\gamma,p_0),
\]
where $\mathcal{S}$ is the state space, $\mathcal{A}$ is the action space, $P(\cdot\mid s,a)$ is a transition kernel on $\mathcal{S}$, $R:\mathcal{S}\times\mathcal{A}\to\R$ is a reward function, $\gamma\in(0,1)$ is a discount factor, and $p_0$ is an initial-state distribution. A (possibly stochastic) policy is a conditional distribution $\pi(\cdot\mid s)$ on $\mathcal{A}$. A trajectory can be written as
\[
\tau=(s_0,a_0,r_0,s_1,a_1,r_1,\dots),
\qquad r_t := R(s_t,a_t),
\]
and is sampled by
\[
s_0\sim p_0,\quad a_t\sim \pi(\cdot\mid s_t),\quad s_{t+1}\sim P(\cdot\mid s_t,a_t).
\]
\textcolor{black}{Throughout the survey we adopt a \emph{deterministic-reward} convention, so that \(R:\mathcal{S}\times\mathcal{A}\to\R\) is a function and \(r_t:=R(s_t,a_t)\) is the realised reward. Nothing in the OT constructions below depends on this choice: when rewards are stochastic, one writes \(r_t\sim R(\cdot\mid s_t,a_t)\) for a reward kernel, replaces \(R(s,a)\) by the expected immediate reward \textcolor{black}{\(R(s,a)=\E[r\mid s,a]\)} in the Bellman equations~\eqref{eq:bellman} and in the objective~\eqref{eq:rl_obj}, and specifies the joint law of \((r_t,s_{t+1})\) given \((s_t,a_t)\). The one place where the distinction is substantive is the distributional Bellman operator of Section~\ref{sec:policy-critics}, where a \emph{random} immediate reward is required; we flag it there.} Here \(Z^\pi(s,a)\) denotes the random discounted return induced by rewards, transitions, and policy stochasticity.
It is often convenient to name the induced distribution on trajectories: let $P_\pi$ denote the trajectory law generated by $(p_0,\pi,P)$. We write $\E_{\pi}[\cdot]$ for expectations with respect to this induced joint distribution over trajectories (equivalently, $\E_{\tau\sim P_\pi}[\cdot]$), and specify the underlying sampling measure explicitly otherwise.
The discounted return from time $t$ is
\[
G_t=\sum_{i=t}^{\infty} \gamma^{\,i-t}\,r_i
=\sum_{i=t}^{\infty} \gamma^{\,i-t}\,R(s_i,a_i),
\]
so future rewards have less weight, by a factor $\gamma^{i-t}$, compared to the current reward.

The \emph{state-value} and \emph{action-value} functions under a fixed policy $\pi$ are
\[
V^\pi(s)=\E_{\pi}[G_0\mid s_0=s],\qquad
Q^\pi(s,a)=\E_{\pi}[G_0\mid s_0=s,a_0=a].
\]
The \emph{advantage} function measures the benefit of taking $a$ in $s$ relative to the average behaviour of $\pi$:
\[
A^\pi(s,a)=Q^\pi(s,a)-V^\pi(s).
\]
Intuitively, $A^\pi(s,a)>0$ signals that $a$ is better than what $\pi$ typically does in $s$, and $A^\pi(s,a)<0$ the opposite. The Bellman relations simply encode one-step consistency: today's value is immediate reward plus expected discounted future value,
\begin{align}
V^\pi(s) & =\E_{a\sim\pi(\cdot\mid s)}\!\Big[R(s,a)+\gamma\,\E_{s'\sim P(\cdot\mid s,a)}[V^\pi(s')]\Big], 
\label{eq:bellman}
\\
Q^\pi(s,a) & =R(s,a)+\gamma\,\E_{s'\sim P(\cdot\mid s,a),\,a'\sim\pi(\cdot\mid s')}\![Q^\pi(s',a')].
\end{align}

\paragraph{Occupancy measures: a distributional ``footprint'' of a policy.}
A central object for what follows is the \emph{discounted occupancy measure} (state-action visitation distribution)
\begin{equation}
\rho_\pi(s,a)=(1-\gamma)\sum_{t=0}^{\infty}\gamma^{\,t}\,\Pr(s_t=s,\ a_t=a\mid \pi).
\label{eq:occupancy}
\end{equation}
The prefactor $(1-\gamma)$ normalises $\rho_\pi$ to a probability measure under standard measurability assumptions, so that in discrete spaces $\sum_{s,a}\rho_\pi(s,a)=1$, and in general $\int \mathrm{d} \rho_\pi(s,a)=1$. \textcolor{black}{Equation~\eqref{eq:occupancy} is written with point probabilities and is therefore a discrete-space statement. In general state-action spaces the definition is taken at the level of measures: $\rho_\pi(B)=(1-\gamma)\sum_{t\ge 0}\gamma^{t}\,\Pr\big((s_t,a_t)\in B\mid\pi\big)$ for measurable $B\subseteq\mathcal{S}\times\mathcal{A}$, which is the form used whenever OT is applied to continuous occupancies below.} Informally, $\rho_\pi$ tells us how often the policy visits each state-action pair after averaging over time and discounting the distant future. A useful way to interpret $\rho_\pi$ (especially in continuous spaces) is via test functions: for any \textcolor{black}{bounded measurable $h$ (or, more generally, any $h$ that is nonnegative or $\rho_\pi$-integrable)},
\[
\int h(s,a)\, \mathrm{d} \rho_\pi(s,a)
=(1-\gamma)\,\E_\pi\!\Big[\sum_{t=0}^\infty \gamma^t h(s_t,a_t)\Big].
\]
Thus $\rho_\pi$ is the discounted time-average distribution of visited state-action pairs. Its state marginal is $\rho_\pi(s)=\sum_a \rho_\pi(s,a)$ and the conditional distribution recovers the policy: $\rho_\pi(a\mid s)=\pi(a\mid s)$ whenever $\rho_\pi(s)>0$. This is one reason occupancies are so useful: they package both \emph{where} the agent goes and \emph{what} it does there into a single object. The fixed-point identity
\[
\rho_\pi(s,a)=(1-\gamma)\,p_0(s)\,\pi(a\mid s)\;+\;\gamma\,\E_{\tilde s,\tilde a\sim\rho_\pi}\!\big[P(s\mid \tilde s,\tilde a)\,\pi(a\mid s)\big]
\]
expresses that discounted visits arise either from the start distribution (first term) or by flowing through the dynamics under $\pi$ (second term). It is useful to read this as a conservation law for probability mass: some mass enters at the initial distribution, and the rest is transported forward by the dynamics and the policy. This flow perspective is the starting point for several OT-based methods, particularly those that operate on occupancy measures (Section~\ref{sec:il}).

\paragraph{The flow constraint}
In finite MDPs, the occupancy measure is not an arbitrary distribution on $\mathcal{S}\times\mathcal{A}$: it must satisfy a conservation-of-mass constraint induced by the dynamics. Writing $\rho_\pi(s)=\sum_a\rho_\pi(s,a)$, one has the equivalent state-level balance equation
\begin{equation}
\rho_\pi(s)=(1-\gamma)p_0(s)+\gamma\sum_{\tilde s,\tilde a}\rho_\pi(\tilde s,\tilde a)\,P(s\mid \tilde s,\tilde a),
\label{eq:flow-constraint}
\end{equation}
and then $\rho_\pi(s,a)=\rho_\pi(s)\pi(a\mid s)$. This makes clear why many OT constructions in RL are naturally stated on occupancies: changing a policy changes its occupancy by \emph{pushing probability mass through the Markov dynamics}. In later sections, OT distances will be used either to compare two occupancies (e.g., learner vs.\ expert) or to constrain how far an update can move in occupancy space. For a reader seeing this for the first time, the important point is that OT is being used on a bona fide distribution that already encodes sequential behaviour.

The control objective can be written compactly in terms of $\rho_\pi$:
\begin{equation}
    J(\pi):=\E_\pi[G_0]=\frac{1}{1-\gamma}\,\E_{(s,a)\sim \rho_\pi}\![R(s,a)].
    \label{eq:rl_obj}
\end{equation}
Thus, many learning objectives reduce to expectations with respect to either $\rho_\pi$ (for occupancy-level matching and regularisation) or $\pi(\cdot\mid s)$ averaged over states drawn from $\rho_\pi$ (for policy-space constraints).

\paragraph{Finite-horizon empirical measures and ``bag-of-states'' viewpoints.}
For finite-horizon problems with horizon $T$, an undiscounted analogue is often used. Given a single rollout $\tau=(s_0, a_0, r_0, \dots,s_{T-1}, r_{T-1})$, the empirical \emph{state marginal} (a ``bag-of-states'' measure) is
\[
\hat p=\frac{1}{T}\sum_{t=0}^{T-1}\delta_{s_t},
\]
where $\delta_{x}$ denotes a Dirac mass at $x$. This empirical measure will serve as the basic object in trajectory-level transports used for offline reward labelling. The phrase ``bag of states'' is meant literally: we keep track of which states appeared, but not in which order. That simplification can be helpful computationally, but it also throws away temporal information. Later sections discuss when this is acceptable and how to reintroduce temporal structure (e.g., banded/masked couplings and windowed/intent-aware costs). OT on empirical trajectory marginals is not sequence-aware by itself; temporal structure must enter through the transported object, the cost, or the admissible couplings. In continuous spaces we reuse the same notation with the measure-theoretic interpretation understood; the empirical measure remains a sum of Dirac masses.

\paragraph{Distributional objects used later.}
Beyond $\rho_\pi$ and $\pi(\cdot\mid s)$, later sections also use: (i) the trajectory law $P_\pi$ and its pushforward $\mu_\pi=\Phi_{\#}P_\pi$ (formally defined in Section~\ref{sec:pre-pushforwards}) under behavioural embeddings $\Phi$ (Section~\ref{sec:policy-bgpg}); 
(ii) expert and behaviour policies $\pi_E$ and $\pi_{\beta}$ (imitation learning and offline RL); and (iii) return distributions $Z^\pi(s,a)$ in distributional RL (Section~\ref{sec:policy-critics}). The guiding principle is consistent: OT compares \emph{policy-induced measures} at the level appropriate to the application.

To summarise the notation that will recur:
\begin{itemize}
\item $\pi(\cdot\mid s)$: policy; $P(\cdot\mid s,a)$: dynamics; $p_0$: initial-state distribution. They generate $P_{\pi}$: trajectory law.
\item $V^\pi,Q^\pi,A^\pi$: value, action-value, and advantage under $\pi$.
\item $\rho_\pi$: discounted occupancy measure; $\rho_\pi(s)=\sum_a\rho_\pi(s,a)$; $\rho_\pi(a\mid s)=\pi(a\mid s)$.
\item $\hat p$: empirical state marginal of a single trajectory (finite-horizon).
\end{itemize}
A more comprehensive list of notation is provided in Table~\ref{tab:notation}.
These objects let us state, with precision, which distributions are compared by each OT construction (e.g., occupancies $\rho$, empirical marginals $\hat p$, or action-conditionals $\pi(\cdot\mid s)$) and where the transport signal enters the RL objective. 

\subsection{Optimal transport in brief}
\label{sec:ot}

We summarise the OT concepts used throughout the survey and fix the notation.
For background, we refer to \citet{peyre2019computational} for computational OT and to \citet{villani2008optimal,santambrogio2015optimal} for an accessible applied introduction and broader theoretical context.

\begin{figure}[t]
\centering
\begin{tikzpicture}[
    src/.style={circle, fill=blue!70, inner sep=2.6pt},
    tgt/.style={circle, fill=orange!80, inner sep=2.6pt},
    plan/.style={-{Stealth[length=2mm]}, semithick, darkblue},
    faint/.style={-{Stealth[length=2mm]}, semithick, gray!35},
    costbox/.style={draw=gray!50, rounded corners=2pt, fill=gray!5, inner sep=4pt},
]
\definecolor{darkblue}{RGB}{0,51,102}

\begin{scope}[shift={(-2.3,0)}]
\node[font=\small\bfseries] at (1.4,2.9) {(a) Cost prefers like-to-like};
\node[src] (x1a) at (0,1.7) {};
\node[src] (x2a) at (0,0.4) {};
\node[tgt] (y1a) at (2.8,1.7) {};
\node[tgt] (y2a) at (2.8,0.4) {};
\node[font=\scriptsize, left=3pt] at (x1a) {$x_1$};
\node[font=\scriptsize, left=3pt] at (x2a) {$x_2$};
\node[font=\scriptsize, right=3pt] at (y1a) {$y_1$};
\node[font=\scriptsize, right=3pt] at (y2a) {$y_2$};
\draw[plan] (x1a) -- (y1a);
\draw[plan] (x2a) -- (y2a);
\draw[faint] (x1a) -- (y2a);
\draw[faint] (x2a) -- (y1a);
\node[costbox, anchor=north] at (1.4,-0.45) {\scriptsize $
c(x_i,y_j)=2\mathbf{1}_{\{i \neq j\}}$};
\node[font=\scriptsize, gray!70, align=center] at (1.4,-1.95) {diagonal plan is cheap:\\match $x_1\!\to\!y_1$, $x_2\!\to\!y_2$};
\end{scope}

\begin{scope}[shift={(2.3,0)}]
\node[font=\small\bfseries] at (1.4,2.9) {(b) Cost prefers cross-match};
\node[src] (x1b) at (0,1.7) {};
\node[src] (x2b) at (0,0.4) {};
\node[tgt] (y1b) at (2.8,1.7) {};
\node[tgt] (y2b) at (2.8,0.4) {};
\node[font=\scriptsize, left=3pt] at (x1b) {$x_1$};
\node[font=\scriptsize, left=3pt] at (x2b) {$x_2$};
\node[font=\scriptsize, right=3pt] at (y1b) {$y_1$};
\node[font=\scriptsize, right=3pt] at (y2b) {$y_2$};
\draw[plan] (x1b) -- (y2b);
\draw[plan] (x2b) -- (y1b);
\draw[faint] (x1b) -- (y1b);
\draw[faint] (x2b) -- (y2b);
\node[costbox, anchor=north] at (1.4,-0.45) {\scriptsize $
c(x_i,y_j)=2\mathbf{1}_{\{i = j\}}$};
\node[font=\scriptsize, gray!70, align=center] at (1.4,-1.95) {crossed plan is cheap:\\match $x_1\!\to\!y_2$, $x_2\!\to\!y_1$};
\end{scope}
\end{tikzpicture}
\caption{The ground cost determines what OT counts as a good match. In both panels, the source and target marginals are the same, but the ground cost changes. Because OT minimises transport cost under that matrix, the optimal plan changes as well. This is why the ground cost is a modelling choice, not a numerical detail.}
\label{fig:ground-cost-matters}
\end{figure}
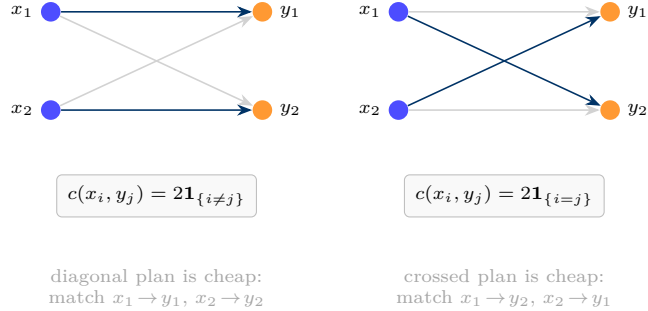

\paragraph{Why OT.}
Many RL problems involve comparing distributions whose supports overlap weakly: learner versus expert occupancies early in imitation learning, learned-policy actions versus logged actions in offline RL, or transition laws under shift. Pointwise divergences such as KL can be brittle under such mismatch, whereas OT remains informative by measuring the \emph{effort to move mass} under a ground cost that encodes task geometry. A minimal toy example is if $\mu=\delta_{x}$ and $\nu=\delta_{y}$, then $\Wone(\mu,\nu)=d(x,y)$ measures the distance between those two points, while \(\mathrm{KL}(\mu\|\nu)\) is infinite for \(x\ne y\). 
Generally, OT lifts the pointwise comparisons to the distribution level. As a result, OT can still provide informative gradients even when two distributions have little or no overlap in support, whereas likelihood-ratio divergences such as KL may become undefined or infinite in such settings. That is the practical reason why OT often continues to provide a useful learning signal under distribution shift and considers the geometry of the underlying space. However, the effectiveness of OT depends heavily on the choice of ground cost.
Figure~\ref{fig:ground-cost-matters} illustrates the key modelling choice in the later RL sections: even with the same source and target marginals, changing the ground cost changes the optimal transport plan, and therefore changes what is considered to be a good match.

\paragraph{Maps, couplings, and pushforwards.}
\label{sec:pre-pushforwards}
Let $\mu\in\mathcal{P}(\mathcal{X})$ and $\nu\in\mathcal{P}(\mathcal{Y})$ be probability measures on measurable spaces $\mathcal{X},\mathcal{Y}$, and let $F:\mathcal{X}\to\mathcal{Y}$ be a measurable map.
The \emph{pushforward} (image) of $\mu$ through $F$ is the measure $F_{\#}\mu\in\mathcal{P}(\mathcal{Y})$ defined by
\[
(F_{\#}\mu)(B)=\mu(F^{-1}(B)),\qquad \forall\,B\subseteq\mathcal{Y}\ \text{measurable}.
\]
A \emph{transport map} is a measurable $F$ such that $F_{\#}\mu=\nu$.
For instance, if $\mu$ is a distribution over states and $\Phi$ a feature map, then $\Phi_{\#}\mu$ is the induced distribution in feature space. 
A pushforward simply draws samples $x\sim \mu$, transforms them with $F$, and obtains the resulting distribution of outputs. 
The relaxation of a transport map is a \emph{coupling} (transport plan) between $\mu$ and $\nu$, which is a probability measure $\Gamma\in\mathcal{P}(\mathcal{X}\times\mathcal{Y})$ whose marginals are $\mu$ and $\nu$.
Whereas a transport map sends mass at each point to a single destination, a transport coupling can split mass across multiple destinations. This matters because in many practical problems the source mass cannot be matched by a deterministic one-to-one assignment.
Writing $P_{\mathcal{X}}:\mathcal{X}\times\mathcal{Y}\to\mathcal{X}$ and $P_{\mathcal{Y}}:\mathcal{X}\times\mathcal{Y}\to\mathcal{Y}$ for coordinate projections, the set of couplings is
\[
\Pi(\mu,\nu)
=\bigl\{\Gamma\in\mathcal{P}(\mathcal{X}\times\mathcal{Y}):\ (P_{\mathcal{X}})_{\#}\Gamma=\mu,\ (P_{\mathcal{Y}})_{\#}\Gamma=\nu\bigr\}.
\]
For a simple example, if $\mu=\delta_x$ and $\nu=\tfrac12\delta_{y_1}+\tfrac12\delta_{y_2}$, then no transport map exists. A coupling handles this immediately by sending mass $1/2$ from $x$ to each $y_i$.

\begin{figure}[t]
\centering
\begin{tikzpicture}[
    dot/.style={circle, fill, inner sep=2.5pt},
    learner/.style={dot, fill=blue!70},
    expert/.style={dot, fill=orange!80},
]

\definecolor{darkblue}{RGB}{0,51,102}

\def\lx{-5.5}
\def\ex{5.5}
\def\ey{2.5}
\def\rowh{0.7}

\node[font=\small\bfseries] at (\lx, 2) {Learner $\tau$};
\node[learner] (L1) at (\lx, 1.4) {};
\node[learner] (L2) at (\lx, 0.7) {};
\node[learner] (L3) at (\lx, 0) {};
\node[learner] (L4) at (\lx, -0.7) {};
\node[learner] (L5) at (\lx, -1.4) {};
\node[font=\small, left=5pt of L1] {$s_1$};
\node[font=\small, left=5pt of L2] {$s_2$};
\node[font=\small, left=5pt of L3] {$s_3$};
\node[font=\small, left=5pt of L4] {$s_4$};
\node[font=\small, left=5pt of L5] {$s_5$};

\node[font=\small\bfseries] at (-2.5, 2.8) {Expert $\tau^E$};
\node[expert] (E1) at (-1.05, \ey) {};
\node[expert] (E2) at (-0.35, \ey) {};
\node[expert] (E3) at (0.35, \ey) {};
\node[expert] (E4) at (1.05, \ey) {};
\node[font=\small, above=3pt of E1] {$s^E_1$};
\node[font=\small, above=3pt of E2] {$s^E_2$};
\node[font=\small, above=3pt of E3] {$s^E_3$};
\node[font=\small, above=3pt of E4] {$s^E_4$};

\node[font=\small\bfseries] at (0, -3.0) {Coupling $\Gamma$};

\node[font=\scriptsize] at (-1.05, 2.1) {$t'{=}1$};
\node[font=\scriptsize] at (-0.35, 2.1) {$t'{=}2$};
\node[font=\scriptsize] at (0.35, 2.1) {$t'{=}3$};
\node[font=\scriptsize] at (1.05, 2.1) {$t'{=}4$};

\node[font=\scriptsize, anchor=east] at (-1.5, 1.4) {$t{=}1$};
\node[font=\scriptsize, anchor=east] at (-1.5, 0.7) {$t{=}2$};
\node[font=\scriptsize, anchor=east] at (-1.5, 0) {$t{=}3$};
\node[font=\scriptsize, anchor=east] at (-1.5, -0.7) {$t{=}4$};
\node[font=\scriptsize, anchor=east] at (-1.5, -1.4) {$t{=}5$};

\fill[blue!35] (-1.4, 1.05) rectangle (-0.7, 1.75);
\fill[blue!12] (-0.7, 1.05) rectangle (0, 1.75);
\node[font=\scriptsize] at (-1.05, 1.4) {.15};
\node[font=\scriptsize] at (-0.35, 1.4) {.05};
\node[font=\scriptsize] at (0.35, 1.4) {.00};
\node[font=\scriptsize] at (1.05, 1.4) {.00};

\fill[blue!25] (-1.4, 0.35) rectangle (-0.7, 1.05);
\fill[blue!25] (-0.7, 0.35) rectangle (0, 1.05);
\node[font=\scriptsize] at (-1.05, 0.7) {.10};
\node[font=\scriptsize] at (-0.35, 0.7) {.10};
\node[font=\scriptsize] at (0.35, 0.7) {.00};
\node[font=\scriptsize] at (1.05, 0.7) {.00};

\fill[blue!25] (-0.7, -0.35) rectangle (0, 0.35);
\fill[blue!25] (0, -0.35) rectangle (0.7, 0.35);
\node[font=\scriptsize] at (-1.05, 0) {.00};
\node[font=\scriptsize] at (-0.35, 0) {.10};
\node[font=\scriptsize] at (0.35, 0) {.10};
\node[font=\scriptsize] at (1.05, 0) {.00};

\fill[blue!25] (0, -1.05) rectangle (0.7, -0.35);
\fill[blue!25] (0.7, -1.05) rectangle (1.4, -0.35);
\node[font=\scriptsize] at (-1.05, -0.7) {.00};
\node[font=\scriptsize] at (-0.35, -0.7) {.00};
\node[font=\scriptsize] at (0.35, -0.7) {.10};
\node[font=\scriptsize] at (1.05, -0.7) {.10};

\fill[blue!12] (0, -1.75) rectangle (0.7, -1.05);
\fill[blue!35] (0.7, -1.75) rectangle (1.4, -1.05);
\node[font=\scriptsize] at (-1.05, -1.4) {.00};
\node[font=\scriptsize] at (-0.35, -1.4) {.00};
\node[font=\scriptsize] at (0.35, -1.4) {.05};
\node[font=\scriptsize] at (1.05, -1.4) {.15};

\draw[thick] (-1.4, 1.75) rectangle (1.4, -1.75);
\draw (-0.7, 1.75) -- (-0.7, -1.75);
\draw (0, 1.75) -- (0, -1.75);
\draw (0.7, 1.75) -- (0.7, -1.75);
\draw (-1.4, 1.05) -- (1.4, 1.05);
\draw (-1.4, 0.35) -- (1.4, 0.35);
\draw (-1.4, -0.35) -- (1.4, -0.35);
\draw (-1.4, -1.05) -- (1.4, -1.05);

\node[font=\tiny, gray] at (1.85, 1.4) {$\to\!.20$};
\node[font=\tiny, gray] at (1.85, 0.7) {$\to\!.20$};
\node[font=\tiny, gray] at (1.85, 0) {$\to\!.20$};
\node[font=\tiny, gray] at (1.85, -0.7) {$\to\!.20$};
\node[font=\tiny, gray] at (1.85, -1.4) {$\to\!.20$};

\node[font=\tiny, gray] at (-1.05, -2.0) {$\downarrow$};
\node[font=\tiny, gray] at (-1.05, -2.3) {.25};
\node[font=\tiny, gray] at (-0.35, -2.0) {$\downarrow$};
\node[font=\tiny, gray] at (-0.35, -2.3) {.25};
\node[font=\tiny, gray] at (0.35, -2.0) {$\downarrow$};
\node[font=\tiny, gray] at (0.35, -2.3) {.25};
\node[font=\tiny, gray] at (1.05, -2.0) {$\downarrow$};
\node[font=\tiny, gray] at (1.05, -2.3) {.25};

\end{tikzpicture}
\caption{Offline reward labelling via optimal transport. An optimal coupling $\Gamma^*$ is computed between empirical state marginals of learner and expert trajectories, treating each as a uniform distribution over its states. This figure shows an example of a coupling, where each row sums to $1/T = 0.20$; each column sums to $1/T_E = 0.25$. The reward at each learner state $s_t$ is the negative transport cost to matched expert states under the optimal plan, $r(s_t) = -\sum_{t'=1}^{T_E} c(s_t, s^E_{t'})\,\Gamma^*_{t,t'}$. A near-diagonal structure emerges when trajectories are temporally aligned. The coupling in the figure is an example and is not necessarily the optimal coupling.}
\label{fig:otr}
\end{figure}
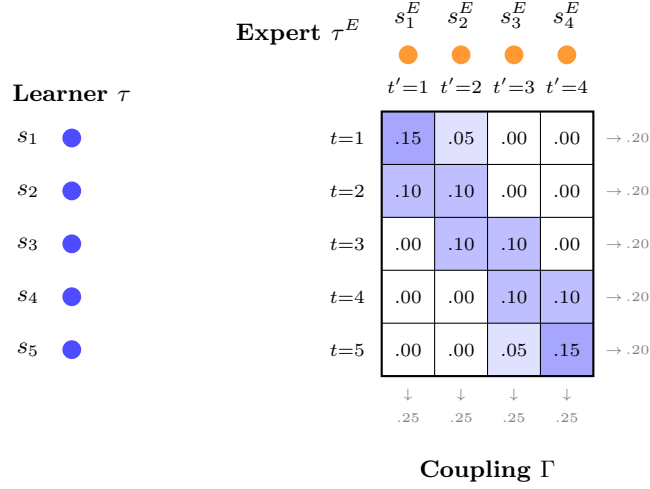

\paragraph{Monge and Kantorovich formulations.}
Given a ground cost $c:\mathcal{X}\times\mathcal{Y}\to\R_{\ge 0}$, the classical Monge problem \citep{monge1781memoire} seeks a transport map $F$ that pushes $\mu$ to $\nu$ and minimises transport cost \citep{santambrogio2015optimal,peyre2019computational}:
\[
\inf_{F:\,F_{\#}\mu=\nu}\ \int_{\mathcal{X}} c\bigl(x,F(x)\bigr)\,\mathrm{d}\mu(x).
\]
Transport maps may fail to exist (for example, when mass must be split), motivating the Kantorovich relaxation~\citep{kantorovitch1958translocation}:
\begin{equation}
\mathrm{OT}_c(\mu,\nu)
:=\inf_{\Gamma\in\Pi(\mu,\nu)}\ \int_{\mathcal{X}\times\mathcal{Y}} c(x,y)\,\mathrm{d}\Gamma(x,y).
\label{eq:kantorovich}
\end{equation}
The only conceptual change from Monge to Kantorovich is that we no longer insist on a unique destination for each source point. Instead, we optimise over couplings $\Gamma$, which can split mass and are therefore much more flexible. In discrete settings, $\Gamma$ can be viewed as a nonnegative matrix whose row/column sums equal the source/target histograms (see Figure~\ref{fig:otr}). In RL applications, $\mathcal{X}$ and $\mathcal{Y}$ may be action spaces, state spaces, or sets of trajectory features.

For discrete distributions, exact optimal transport can be formulated as a linear programming problem and solved using algorithms such as the network simplex \citep{dantzig1948programming} or interior-point methods \citep{karmarkar1984new} with cubic complexity $\mathcal{O}(N^3\log N)$ where $N$ is the number of particles from each distribution \citep{cuturi2013sinkhorn, kolouri2016transport}. 
Recent advances have substantially reduced the cost of large-scale OT. \textcolor{black}{For example, HiRef \citep{halmos2025hierarchical} is a hierarchical, low-rank refinement scheme for the bijective Monge (assignment) problem between two datasets of equal size, reported to run in log-linear time and linear space. These rates are conditional on a low-rank factorisation $C=AB^{\top}$ of the cost being available with small inner dimension.
Exactness is likewise conditional: the refinement invariant recovers the optimal assignment only given an \emph{optimal} low-rank solver at each level, and the authors observe that \textcolor{black}{the low-rank problems solved in practice are non-convex, so the corresponding numerical solvers are not guaranteed to attain a global optimum.}
The reported constants also depend on the low-rank subroutine, \textcolor{black}{and the practical hierarchy depends on the choice of a rank-annealing schedule whose cumulative ranks are constrained by the sample count.}
It is therefore best read as a scalable hierarchical scheme, exact under idealised subproblem solutions rather than a general-purpose exact solver, and the assumptions behind the computational scaling should be kept separate from those behind exact recovery.}
For continuous distributions, exact OT is an infinite-dimensional optimisation problem. In practice, continuous distributions are often approximated by empirical measures with samples, reducing the problem to a discrete OT problem.

The associated (general) Kantorovich dual problem is
\[
\mathrm{OT}_c(\mu,\nu) = \sup_{u,v}\lb\int_{\mathcal{X}} u(x)\,\mathrm{d}\mu(x)+\int_{\mathcal{Y}} v(y)\,\mathrm{d}\nu(y)\ :\ u(x)+v(y)\le c(x,y) \ \ \text{\textcolor{black}{for all }} \textcolor{black}{(x,y)\in\mathcal{X}\times\mathcal{Y}}\rb,
\]
where $u$ and $v$ are dual potentials \textcolor{black}{and the constraint is imposed pointwise, under the standard hypotheses that $c$ is lower semicontinuous and bounded below on Polish spaces}. Readers who mainly care about applications can think of the primal problem as ``find the best transport plan'' and the dual problem as ``find the most discriminating test functions consistent with the cost''. 

Equivalently, one may write
\[\mathrm{OT}_c(\mu,\nu) = \sup_{f}\ls\int f^c(x)\,\dd\mu(x)+\int f(y)\,\dd\nu(y)\rs\]
where $f^c(x)=\inf_y\{c(x,y)-f(y)\}$ is the $c$-transform of $f$. This single-potential formulation is often more convenient in optimisation and underpins several dual OT solvers used later in the survey.

\textcolor{black}{
Rewriting the $c$-transform yields a maximin formulation involving an optimisation over measurable functions $F:\mathcal{X}\to\mathcal{Y}$:}
\[\mathrm{OT}_c(\mu,\nu) = \textcolor{black}{\sup_f \inf_{F}} \E_{x\sim\mu}[c(x,F(x))-f(F(x))]+\E_{y\sim\nu}[f(y)],\]
\textcolor{black}{where the extrema are written as $\sup$ and $\inf$ because attainment is not guaranteed. Two points deserve emphasis. First, $F$ is an \emph{unconstrained measurable selector}: it is not required to satisfy $F_{\#}\mu=\nu$, and it need not induce an optimal coupling. Second, replacing the pointwise infimum $f^c(x)=\inf_y\{c(x,y)-f(y)\}$ by an infimum over measurable maps $F$ requires a measurable-selection argument together with integrability of $c(x,F(x))$; this is immediate for finitely supported measures, and holds more generally under the usual lower-semicontinuity and measurable-selection hypotheses \citep{villani2008optimal,santambrogio2015optimal}. The preceding $c$-transform identity is the safer general statement, and we treat the maximin form as its measurable realisation.} This form is used, for example, in dual formulations of partial OT for offline RL (Section~\ref{sec:offline-ot}). For most readers, the important point is not the algebra itself, but that these equivalent forms give different computational handles on the same OT problem.

\paragraph{Wasserstein distances and the $\Wone$ dual form.}
When $\mathcal{X}=\mathcal{Y}$ is equipped with a ground metric $d$ and $c(x,y)=d(x,y)^p$ for $p\in[1,\infty)$, the $p$-Wasserstein distance is
\[
\Wassp{p}(\mu,\nu)
=\Big(\inf_{\Gamma\in\Pi(\mu,\nu)}\int_{\mathcal{X}\times\mathcal{X}} d(x,y)^p\,\mathrm{d}\Gamma(x,y)\Big)^{1/p}.
\]
\textcolor{black}{This is finite, and defines a metric on the set $\mathcal{P}_p(\mathcal{X})$ of measures with finite $p$-th moment, $\int d(x,x_0)^p\,\mathrm{d}\mu(x)<\infty$ for some $x_0$; all Wasserstein distances below are understood on $\mathcal{P}_p$, and the $\Wtwo$ gradient flows of Section~\ref{sec:policy-wgf} on $\mathcal{P}_2$.}
For $p=1$ and $c(x,y)=\|x-y\|$, Kantorovich-Rubinstein duality yields
\begin{equation}
\Wone(\mu,\nu)=\sup_{\|g\|_{\mathrm{Lip}}\le 1}\Big\{\int g\,\mathrm{d}\mu-\int g\,\mathrm{d}\nu\Big\},
\label{eq:kr-dual}
\end{equation}
which underpins critic-based OT estimators, such as Wasserstein generative adversarial network (WGAN)-style training, used in several RL objectives. In RL terms, the dual test function $g$ is exactly the ``1-Lipschitz critic'' that can be trained from samples and then used as a shaping signal (e.g., imitation via occupancy matching).
This dual form is one of the main reasons $\Wone$ is so popular in machine learning: it replaces an optimisation over couplings by an optimisation over a critic function that can be parameterised by a neural network.

\paragraph{$\Wtwo$ structure and transport maps.}
With quadratic cost on $\R^d$, $c(x,y)=\|x-y\|^2$, the squared 2-Wasserstein distance is
\[
\Wtwo^2(\mu,\nu)=\inf_{\Gamma\in\Pi(\mu,\nu)}\int_{\R^d\times\R^d}\|x-y\|^2\,\mathrm{d}\Gamma(x,y).
\]
If $\mu$ is absolutely continuous (and mild regularity holds), Brenier's theorem implies that an optimal coupling is induced by a map $T=\nabla \varphi$ where the potential $\varphi$ is convex and $T_{\#}\mu=\nu$ \citep{brenier1991polar}. \textcolor{black}{Concretely, it suffices that $\mu,\nu\in\mathcal{P}_2(\R^d)$ with $\mu$ absolutely continuous with respect to Lebesgue measure; the map is then $\mu$-a.e.\ unique. These conditions are assumed wherever Brenier maps are invoked later. They are \emph{not} needed for the Kantorovich or Kantorovich-Rubinstein dualities, which we apply freely to empirical atomic measures.}
This observation motivates \emph{map-based} approximations to $\Wtwo$ in learning pipelines: one parameterises convex potentials, for example with input-convex architectures, and uses the induced gradient map. In practice, $\Wtwo$ structure is particularly convenient when distributions are approximately Gaussian or when one wants explicit transport maps.

\begin{figure}[t]
    \centering
    \includegraphics[width=0.4\linewidth]{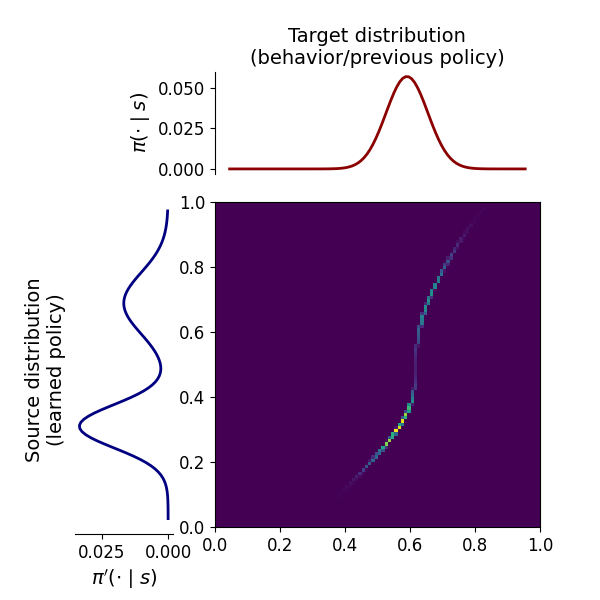}
    \includegraphics[width=0.4\linewidth]{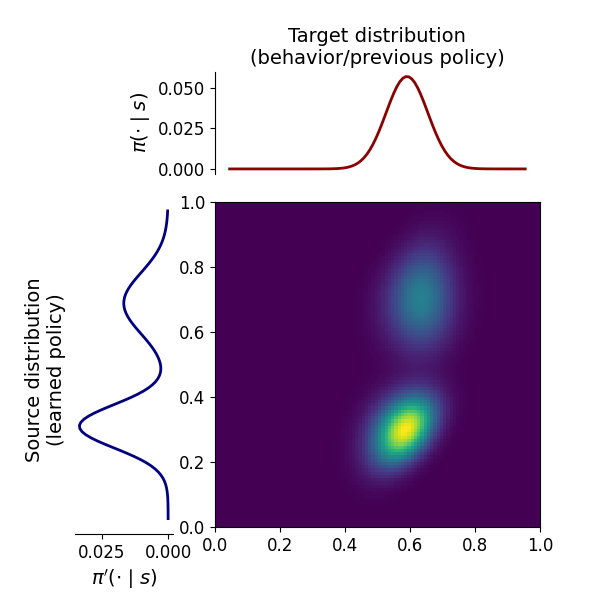}
    \caption{Example of optimal transport between two continuous distributions. The heatmap in the left figure shows the standard OT plan, and the right figure shows the entropy-regularised OT plan ($\varepsilon=0.01$), which is smoother. Brighter colour indicates more mass is transported.}
    \label{fig:ctsOT}
\end{figure}

\paragraph{Entropic regularisation, Sinkhorn iterations, and Sinkhorn divergence.}
For empirical (often discrete) measures, solving $\mathrm{OT}_c$ exactly can be costly \textcolor{black}{with complexity $\mathcal{O}(N^3\log N)$, where the number of atoms in $\mu$ and $\nu$ is $\mathcal{O}(N)$}. A standard remedy is entropic regularisation \citep{cuturi2013sinkhorn}:
\[
\mathrm{OT}_\varepsilon(\mu,\nu)
=\inf_{\Gamma\in\Pi(\mu,\nu)}
\Big\{\int c(x,y)\,\mathrm{d}\Gamma(x,y)\;+\;\varepsilon\,\KL\!\big(\Gamma\,\big\|\,\mu\otimes\nu\big)\Big\},
\]
where $\mu\otimes\nu$ is the product measure and $\KL$ is the Kullback-Leibler divergence.
The regularised objective is strictly convex and can be solved efficiently by Sinkhorn (matrix-scaling) iterations.
\textcolor{black}{Each dense Sinkhorn iteration costs $\mathcal{O}(N^2)$ time and $\mathcal{O}(N^2)$ storage \citep{cuturi2013sinkhorn}. For a target accuracy $\tau_{\mathrm{tol}}>0$ in transport cost, \citet{luo2023improved} establish a total complexity of $\mathcal{O}\big(N^2\|C\|_{\infty}^2\log N/\tau_{\mathrm{tol}}^2\big)$; this is an accuracy-dependent guarantee \emph{after rounding} the approximate plan onto $\Pi(\mu,\nu)$, and it requires the entropic coefficient $\varepsilon$ and the marginal stopping tolerance to be chosen as functions of $\tau_{\mathrm{tol}}$. It is therefore not the total cost for an arbitrary fixed $\varepsilon$ and an arbitrary stopping rule. Throughout the survey $\varepsilon$ denotes the entropic coefficient and $\tau_{\mathrm{tol}}$ a target numerical accuracy.}
When $c(x,y)=d(x,y)^p$, we write\textcolor{black}{, for the \emph{unrooted} regularised cost,}
\[
\Wassp{\varepsilon}^p(\mu,\nu)
=\inf_{\Gamma\in\Pi(\mu,\nu)}
\Big\{\int d(x,y)^p\,\mathrm{d}\Gamma(x,y)\;+\;\varepsilon\,\KL\!\big(\Gamma\,\big\|\,\mu\otimes\nu\big)\Big\}
\]
for the entropy-regularised Wasserstein distance. \textcolor{black}{Two conventions should be kept in mind. First, $\Wassp{\varepsilon}$ without a superscript always denotes this same unrooted quantity (that is, $\Wassp{\varepsilon}:=\Wassp{\varepsilon}^p$ with the $p$-dependence suppressed), so it is the unrooted cost that is debiased below; we never take a $p$-th root of a regularised cost. Second, $\Wassp{p}$ is reserved for the genuine $p$-Wasserstein \emph{distance}: an unregularised transport cost with $c=d^p$ equals $\Wassp{p}^p$, not $\Wassp{p}$. Where a method is described below as using a cosine dissimilarity or an arbitrary learned cost, the object computed is $\mathrm{OT}_c$ (or $\Wassp{\varepsilon}$) with that cost, and we call it a Wasserstein distance only when the cost is a genuine metric raised to a power.} Figure~\ref{fig:ctsOT} illustrates the effect of entropic regularisation on continuous transport plans, which leads to a smoother coupling.
Because $\Wassp{\varepsilon}$ is biased by the entropy term, one often switches to the debiased \emph{Sinkhorn divergence} \citep{genevay2018learning}
\[
\Seps{\varepsilon}(\mu,\nu)=\Wassp{\varepsilon}(\mu,\nu)-\tfrac12 \Wassp{\varepsilon}(\mu,\mu)-\tfrac12 W_\varepsilon(\nu,\nu)
\]
which satisfies $\Seps{\varepsilon}(\mu,\mu) = 0$ (whereas generally $\Wassp{\varepsilon}(\mu,\mu)\neq 0$). \textcolor{black}{Debiasing removes the self-transport offset and, under suitable conditions on $c$, yields a positive-definite divergence that metrises weak convergence \citep{feydy2019interpolating}. It does not, however, deliver a metric, and we therefore refer to $\Seps{\varepsilon}$ as a debiased divergence rather than a distance.} This debiasing matters in RL because otherwise the OT loss keeps exerting pressure even when agent and target distributions already match. However, entropy-regularisation without correction is widely used in existing works.
A useful way to read this correction is that it subtracts away the artificial self-transport cost created by entropic smoothing.
At a high level, entropic regularisation trades exactness for smoothness: instead of a sparse, nearly hard assignment between points, one obtains a softer plan that spreads some mass across multiple destinations. In practice, $\varepsilon$ should be chosen on the same scale as typical values of the ground cost $c$. If it is too small, the plan becomes sharp but Sinkhorn iterations slow down and become numerically fragile. If it is too large, the transport plan is oversmoothed, and the task geometry is washed out. Log-domain Sinkhorn is the standard stabilisation trick in the small-$\varepsilon$ regime. In RL pipelines, this entropic form is usually the default whenever OT must be computed repeatedly on minibatches, for example in behaviour regularisers, reward labelling, or trust-region penalties.

\paragraph{Heterogeneous spaces: Gromov-Wasserstein.}
When domains differ, a meaningful cross-space cost $c(x,y)$ may be unavailable (e.g.\ sim-to-real, different morphologies, or different sensors).
Gromov-Wasserstein (GW) \citep{memoli2011gromov} compares \emph{relations} rather than points: given intra-space dissimilarities $d_{\mathcal{X}}$ on $\mathcal{X}$ and $d_{\mathcal{Y}}$ on $\mathcal{Y}$, GW seeks a coupling $\Gamma\in\Pi(\mu,\nu)$ that best matches pairwise distances,
\[
\inf_{\Gamma\in\Pi(\mu,\nu)}
\iint\big|d_{\mathcal{X}}(x,x')-d_{\mathcal{Y}}(y,y')\big|^p\,\mathrm{d}\Gamma(x,y)\,\mathrm{d}\Gamma(x',y').
\]
GW is therefore the natural OT tool when structural alignment matters but a pointwise cross-domain cost is unavailable. In RL, this situation arises most clearly in cross-domain imitation and transfer, where within-domain geometry is trustworthy but pointwise cross-domain comparisons are not.
The simplest intuition is that GW asks not ``which point matches which point?'', but ``which neighbourhood structure in one domain resembles which neighbourhood structure in the other?''
For example, suppose one agent has three states arranged in a line, and another has three visually different states also arranged in a line. A pointwise cost between raw observations may be meaningless, but GW can still recognise that the two systems share the same internal geometry.

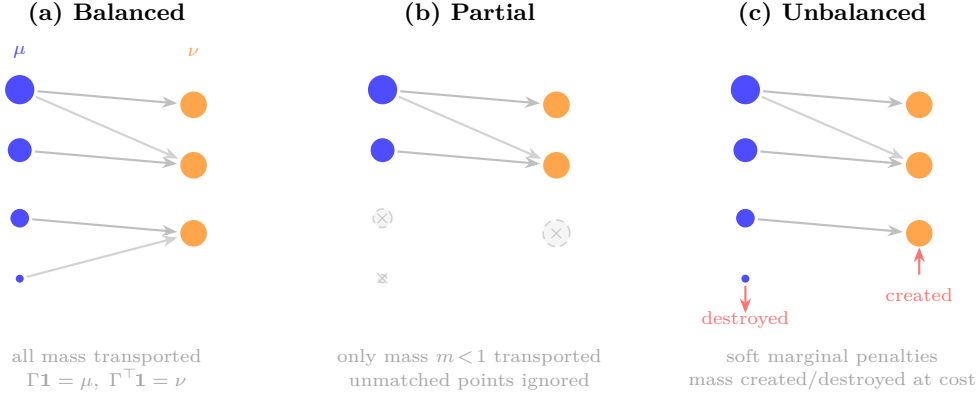
\begin{figure}[t]
\centering
\begin{tikzpicture}[
    dot/.style={circle, inner sep=0pt},
    src/.style={dot, fill=blue!70},
    tgt/.style={dot, fill=orange!70},
    arr/.style={-{Stealth[length=2mm]}, thick, gray!50, shorten >=1pt, shorten <=1pt},
    ghost/.style={dot, draw=gray!35, fill=gray!8, densely dashed, line width=0.6pt},
    lbl/.style={font=\scriptsize, align=center},
]

\definecolor{darkblue}{RGB}{0,51,102}

\begin{scope}[shift={(-4.8,0)}]
\node[font=\small\bfseries] at (1.15, 3.0) {(a) Balanced};

\node[font=\scriptsize, blue!70] at (0, 2.5) {$\mu$};
\node[font=\scriptsize, orange!70] at (2.3, 2.5) {$\nu$};

\node[src, minimum size=11pt] (a1) at (0, 2) {};
\node[src, minimum size=9pt] (a2) at (0, 1.2) {};
\node[src, minimum size=7pt]  (a3) at (0, 0.3) {};
\node[src, minimum size=3pt]  (a4) at (0, -0.5) {};

\node[tgt, minimum size=10pt]  (b1) at (2.3, 1.8) {};
\node[tgt, minimum size=10pt]  (b2) at (2.3, 1) {};
\node[tgt, minimum size=10pt] (b3) at (2.3, 0.1) {};

\draw[arr] (a1) -- (b1);
\draw[arr, gray!35] (a1) -- (b2);
\draw[arr] (a2) -- (b2);
\draw[arr] (a3) -- (b3);
\draw[arr, gray!35] (a4) -- (b3);

\node[lbl, gray!70] at (1.15, -1.7) {all mass transported\\[1pt]$\Gamma\mathbf{1}=\mu,\;\Gamma^\top\!\mathbf{1}=\nu$};
\end{scope}

\begin{scope}[shift={(0,0)}]
\node[font=\small\bfseries] at (1.15, 3.0) {(b) Partial};

\node[src, minimum size=11pt] (c1) at (0, 2) {};
\node[src, minimum size=9pt] (c2) at (0, 1.2) {};
\node[ghost, minimum size=7pt] (c3) at (0, 0.3) {};
\node[ghost, minimum size=3pt] (c4) at (0, -0.5) {};

\node[tgt, minimum size=10pt] (d1) at (2.3, 1.8) {};
\node[tgt, minimum size=10pt] (d2) at (2.3, 1) {};
\node[ghost, minimum size=10pt] (d3) at (2.3, 0.1) {};

\draw[arr] (c1) -- (d1);
\draw[arr, gray!35] (c1) -- (d2);
\draw[arr] (c2) -- (d2);

\node[font=\scriptsize, gray!50] at (0, 0.3) {$\times$};
\node[font=\scriptsize, gray!50] at (0, -0.5) {$\times$};
\node[font=\scriptsize, gray!50] at (2.3, 0.1) {$\times$};

\node[lbl, gray!70] at (1.15, -1.7) {only mass $m\!<\!1$ transported\\[1pt]unmatched points ignored};
\end{scope}

\begin{scope}[shift={(4.8,0)}]
\node[font=\small\bfseries] at (1.15, 3.0) {(c) Unbalanced};

\node[src, minimum size=11pt] (e1) at (0, 2) {};
\node[src, minimum size=9pt] (e2) at (0, 1.2) {};
\node[src, minimum size=7pt]  (e3) at (0, 0.3) {};
\node[src, minimum size=3pt]  (e4) at (0, -0.5) {};

\node[tgt, minimum size=10pt]  (f1) at (2.3, 1.8) {};
\node[tgt, minimum size=10pt]  (f2) at (2.3, 1) {};
\node[tgt, minimum size=10pt] (f3) at (2.3, 0.1) {};

\draw[arr] (e1) -- (f1);
\draw[arr, gray!35] (e1) -- (f2);
\draw[arr] (e2) -- (f2);
\draw[arr] (e3) -- (f3);

\draw[red!55, thick, -{Stealth[length=2mm]}, shorten <=1pt] (e4.south) -- ++(0,-0.4);
\node[font=\scriptsize, red!55, below] at (0, -0.8) {destroyed};

\draw[red!55, thick, {Stealth[length=2mm]}-, shorten >=1pt] (f3.south) -- ++(0,-0.4);
\node[font=\scriptsize, red!55, below] at (2.3, -0.5) {created};

\node[lbl, gray!70] at (1.15, -1.7) {soft marginal penalties\\[1pt]mass created/destroyed at cost};
\end{scope}

\end{tikzpicture}
\caption{Three OT formulations. (a)~Balanced OT enforces exact marginal constraints: all source mass reaches the target. (b)~Partial OT transports only a fraction $m$ of the total mass, leaving the rest unmatched ($\times$)---useful when data contain outliers or suboptimal demonstrations (Section~\ref{sec:offline}). (c)~Unbalanced OT allows mass creation and destruction at a cost (red arrows), softly penalising marginal deviations rather than enforcing them exactly. Dot sizes indicate relative mass.}
\label{fig:ot-variants}
\end{figure}

\paragraph{Mass mismatch: unbalanced and partial OT.}
Balanced OT enforces exact marginal constraints, which can be undesirable when demonstrations are partial, contain outliers, or reflect systematic mass mismatch.
\emph{Unbalanced OT} relaxes the marginal constraints by penalising deviations of the coupling's marginals from $\mu$ and $\nu$ (often using KL-type penalties). A common KL-penalised form is
\[
\inf_{\Gamma\in\mathcal{M}_{+}(\mathcal{X}\times\mathcal{Y})}
\Big\{\int c\,\mathrm{d}\Gamma
+\lambda_1\,\KL\big((P_{\mathcal{X}})_{\#}\Gamma\ \|\ \mu\big)
+\lambda_2\,\KL\big((P_{\mathcal{Y}})_{\#}\Gamma\ \|\ \nu\big)\Big\},
\]
where $\mathcal{M}_{+}$ denotes finite nonnegative measures on $\mathcal{X}\times\mathcal{Y}$.
\textcolor{black}{Because the marginals of $\Gamma$ are finite nonnegative measures rather than probability measures, $\KL$ here denotes the \emph{generalised} Kullback-Leibler divergence, which includes total-mass correction terms: for $\alpha,\beta\in\mathcal{M}_{+}$ with $\alpha\ll\beta$, $\KL(\alpha\|\beta)=\int\log\frac{\mathrm{d}\alpha}{\mathrm{d}\beta}\,\mathrm{d}\alpha-\alpha(\mathcal{X})+\beta(\mathcal{X})$, and $+\infty$ otherwise \citep{chizat2018scaling}. Ordinary probability-only KL is not a complete definition in this setting.}
As $\lambda_1,\lambda_2\to\infty$, one recovers the balanced formulation.
\emph{Partial OT} instead transports only a prescribed mass $m\le 1$ by restricting to couplings with total mass $m$ (and sub-marginal constraints):
\[
\Pi_m(\mu,\nu)
=\Big\{\Gamma\in\mathcal{M}_{+}(\mathcal{X}\times\mathcal{Y}):\ (P_{\mathcal{X}})_{\#}\Gamma\le \mu,\ (P_{\mathcal{Y}})_{\#}\Gamma\le \nu,\ \Gamma(\mathcal{X}\times\mathcal{Y})=m\Big\},
\]
and minimising $\int c\,\mathrm{d}\Gamma$ over $\Pi_m(\mu,\nu)$.
These variants are the natural OT tools when the dataset should not be matched in full (e.g.\ suboptimal demonstrations, incomplete coverage). In RL terms, they allow ``some mass to be ignored'' (partial) or ``mass to be created/removed at a price'' (unbalanced), which is often more faithful to real datasets. Figure~\ref{fig:ot-variants} contrasts the three formulations on a simple example.
The distinction is easiest to remember this way: partial OT says ``match only the best part of the data,'' whereas unbalanced OT says ``match everything you can, but allow mass mismatch if paying the penalty is cheaper.''

\paragraph{Barycentres and interpolation.}
\emph{$p$-Wasserstein barycentres} interpolate between distributions by minimising a weighted sum of Wasserstein distances,
\begin{equation}
\eta^\star = \argmin_{\eta}\ \sum_{k} \lambda_k\,\Wassp{p}^p(\eta,\mu_k),
\qquad \lambda_k\ge 0,\ \sum_k \lambda_k=1,
\label{eq:barycentre}
\end{equation}
and appear in curricula (interpolating task distributions) and in some multi-agent consensus formulations. They can be viewed as the OT analogue of ``averaging'' distributions in a geometry-aware way, and form the basis of displacement interpolation techniques widely used in graphics and imaging \citep{bonneel2011displacement}.

\paragraph{Sliced distances and scalable approximations.}
Several works in the survey use \emph{sliced} (or generalised sliced) Wasserstein distances \citep{kolouri2019generalized} as computationally efficient surrogates in high dimensions. The sliced Wasserstein distance compares one-dimensional projections:
\[
\SWp{p}^p(\mu,\nu)=\int_{\theta\in\mathbb{S}^{d-1}} \Wassp{p}^p\big((\operatorname{Proj}_{\theta})_{\#}\mu,(\operatorname{Proj}_{\theta})_{\#}\nu\big)\,\mathrm{d}\theta,
\]
where $\text{Proj}_{\theta}(x)=\langle x, \theta \rangle$ is the linear projection on direction $\theta$. It is estimated in practice by Monte Carlo sampling, and generalised variants replace the linear projection by more expressive transforms. 
Assume that $N$ observations in $\mathbb{R}^d$ are projected along $L$ directions, the computational complexity becomes $\mathcal{O}(LdN+LN\log N)$ \citep{nadjahi2021fast}.
The intuition is that 1D Wasserstein distances are cheap with a closed form, and averaging over many slices yields a tractable proxy for high-dimensional transport. Recent work refines this approximation by using sliced generalised Wasserstein geodesics to better capture transport structure beyond random projections \citep{mahey2023fastswgg}, and by learning differentiable approximations to generalised sliced transport plans to improve optimisation stability and computational efficiency \citep{chapel2025dgswplans}.

\paragraph{Which OT variant when?}
The survey repeatedly returns to a small set of modelling choices:
(i) \emph{$\Wone$ (dual critic):} when supports may be disjoint, and one wants a critic-like shaping signal (imitation, some robustness/regularisation);
(ii) \emph{entropy-regularised $\Wassp{\varepsilon}$:} when a smooth transport plan, stable gradients and improved computational efficiency are preferred (e.g. in large-scale applications);
(iii) \emph{Sinkhorn divergence $\Seps{\varepsilon}$:} when one wants entropic stability but also a debiased, distance-like quantity;
(iv) \emph{$\Wtwo$:} when closed forms are available with Gaussian distributions or transport maps are desired using a convex potential;
(v) \emph{GW:} when cross-domain pointwise costs are ill-posed but relational structure is meaningful (cross-domain imitation/transfer);
(vi) \emph{partial/unbalanced OT:} when demonstrations/data are incomplete, contaminated, or systematically mismatched in mass;
(vii) \emph{sliced OT:} when dimension makes full OT expensive and projections are adequate;
(viii) \emph{OT with windowed cost or masking:} when the temporal structure is crucial.

\paragraph{Conventions for the remainder of the survey.}
Couplings are denoted $\Gamma$ and $\Pi(\mu,\nu)$ indicates balanced marginal constraints unless stated otherwise; $\Wassp{p}$ denotes the $p$-Wasserstein distance, $\Wassp{\varepsilon}$ the entropy-regularised Wasserstein distance, and $\Seps{\varepsilon}$ the Sinkhorn divergence.
When we do not wish to specify a particular variant of Wasserstein distance, we use $\Wass$.
When we speak of a \emph{critic}, we mean a 1-Lipschitz potential in the $\Wone$ dual in Section~\ref{sec:il} and a value function in the actor-critic setting later; when we speak of a \emph{map}, we mean a $\Wtwo$-motivated transport map (typically $T=\nabla\varphi$ in Euclidean settings).
GW and mass-relaxed variants (unbalanced/partial) are invoked when spaces or masses do not match.

\paragraph{Cost design and numerical scaling.}
The ground cost $c(x,y)$ (or metric $d(x,y)$ when $c=d^p$) is the main modelling degree of freedom in OT-based RL. In RL applications, the generic points $x,y$ may be \emph{states} ($x=s,y=s'$), \emph{actions} ($x=a,y=a'$), \emph{state-action pairs} ($x=(s,a),y=(s',a')$), or learned \emph{trajectory/behaviour embeddings} (e.g.\ $x=\Phi(\tau)$). In simple control tasks, Euclidean or cosine distances on states/actions may suffice; in high-dimensional or visual domains, costs are typically computed in a learned feature space $f$, e.g.\ $c(s,s')=\|f(s)-f(s')\|$ (or a cosine dissimilarity), and analogously for actions when action geometry is non-trivial. Physics- or goal-aware metrics (task coordinates, kinematic distances, safety margins) should be preferred when available, and heterogeneous units must be normalised so that no single coordinate dominates $c$.

Numerically, entropic regularisation $\varepsilon$ should be scaled to the typical magnitude of $c$ (e.g.\ the median or a nearby percentile of pairwise costs on representative minibatches), and log-domain Sinkhorn stabilises computation for small $\varepsilon$. For $\Wone$ dual critics, approximate $1$-Lipschitzness is enforced (gradient penalty or spectral normalisation); for map-based $\Wtwo$ approximations, convex parameterisations help ensure valid transports. In later sections, we state explicitly the compared objects, any temporal constraints used when comparing trajectories, the representation used to define $c$ or $d$, and the numerical scaling/stabilisation choices (e.g.\ $\varepsilon$ and Lipschitz/convexity controls) implemented.

\section{Imitation learning and reward inference}
\label{sec:il}

\begin{figure}[t]
\centering
\begin{tikzpicture}[
    node distance=1.2cm and 1.8cm,
    rlbox/.style={rectangle, draw, rounded corners=3pt, minimum width=2cm,
        minimum height=0.7cm, align=center, font=\small, fill=gray!8},
    otbox/.style={rectangle, draw, rounded corners=3pt, minimum width=2cm,
        minimum height=0.7cm, align=center, font=\small,
        draw=darkblue, text=darkblue, fill=blue!5},
    arr/.style={-{Stealth[length=1.5mm]}, semithick, gray!60},
    otarr/.style={<->, semithick, darkblue, dashed},
    otlbl/.style={font=\scriptsize, text=darkblue, align=center, fill=white, inner sep=2pt}
]

\definecolor{darkblue}{RGB}{0,51,102}

\node[rlbox] (expert) {Expert demos\\[-1pt]{\scriptsize $\rho_{\pi_E}$}};
\node[otbox, right=2.5cm of expert] (ot) {OT comparison\\[-1pt]{\scriptsize $\Wass(\rho_\pi, \rho_{\pi_E})$}};
\node[rlbox, right=2.5cm of ot] (learner) {Learner rollouts\\[-1pt]{\scriptsize $\rho_\pi$}};

\node[rlbox, below=1.2cm of ot] (policy) {Policy\\[-1pt]{\scriptsize $\pi_{\theta}$}};

\draw[otarr] (expert) -- (ot);
\draw[otarr] (learner) -- (ot);

\draw[arr] (ot) -- (policy) node[midway, right, font=\scriptsize, text=gray!80] {reward / gradient};

\end{tikzpicture}
\caption{OT in imitation learning and reward inference (Section~\ref{sec:il}). OT compares learner and expert behaviour distributions via dual critics or primal couplings, yielding shaping rewards or pseudo-reward labels (Sections~\ref{sec:il-adversarial}--\ref{sec:il-otr}). Temporal/intent-aware extensions modify the coupling or ground cost (Section~\ref{sec:temporal}); cross-domain settings use GW (Section~\ref{sec:il-pretrain}).}
\label{fig:IL}
\end{figure}
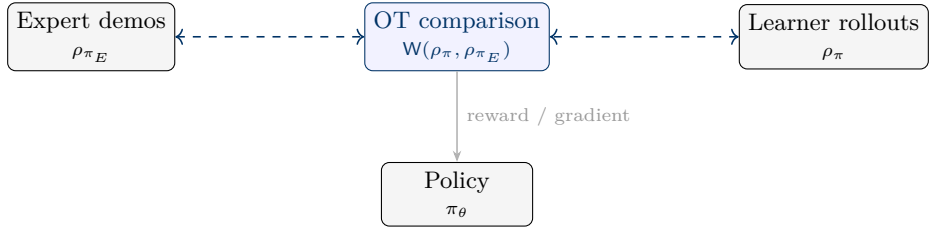

In many practical settings, the reward is unknown or hard to specify, yet demonstrations of desirable behaviour are available. 
Imitation learning seeks a policy from demonstrations, typically without access to rewards. Once imitation is written as occupancy matching, the main design choice becomes the discrepancy between the learner's state-action distribution and the expert's (Figure~\ref{fig:IL}). OT is attractive here because it replaces the geometry-blind divergences used in standard adversarial imitation with a discrepancy that remains informative under weak overlap and can incorporate task structure through the ground cost. In practical terms, OT helps answer the question ``how far is the learner from the expert, and in what sense?'' In this section, we start from the occupancy-matching view, then show how that OT signal appears in online reward shaping, offline pseudo-reward labelling, temporally aware matching, and cross-domain alignment.

To make it concrete, we consider a simple example for illustration. Take the robotic pick-and-place task as an example, where a robotic arm equipped with a gripper needs to pick up a block from the table and place it at a target position. The state contains the kinematic information (e.g. position, rotation and velocity) of both the robot and the block. The action consists of the end-effector displacement and the gripper command (open or close). For this task, the reward may naturally reflect the progress of moving the block towards its target position. However, it may be costly or impractical to annotate rewards, and they are assumed to be unavailable in IL. Given expert demonstrations of successful task executions, IL aims to learn a policy that reproduces the expert behaviour. One approach is to directly match the state-action occupancy distributions (the frequency each state-action pair appears in the demonstrations compared to the trajectories of the learned agent). 
Alternatively, one can first infer the reward function and then learn a policy that maximises the expected cumulative reward. With mixed-quality datasets in the offline RL setting, some OT-based methods shape the rewards by comparing the state visitation distributions between expert and unlabeled trajectories.

\subsection{Imitation as occupancy matching}
\label{sec:il-background}

It is useful to introduce three main categories of IL because OT will enter naturally only after this setup is clear.

\emph{Behaviour cloning} (BC) treats imitation as supervised learning on demonstration pairs $(s,a)$, fitting $\pi(\cdot\mid s)$ to the expert’s actions. It is simple and effective in-distribution but brittle under covariate shift: small action errors at test time move the agent into states that are rare or absent in the dataset, where the learned conditional $\pi(\cdot\mid s)$ is poorly specified, and errors compound over time. Intuitively, BC asks the right question one step at a time, but has no mechanism for correcting the distribution shift that those one-step mistakes create.

\emph{Inverse reinforcement learning} (IRL) instead posits an unknown \emph{cost} function $\ell:\mathcal{S}\times\mathcal{A}\to\R$ that explains expert behaviour and then seeks a policy that minimises this cost (we use $\ell$ to avoid clashing with the OT ground cost $c(x,y)$). A key object is the normalised discounted occupancy measure $\rho_\pi(s,a)$ (Section~\ref{sec:preliminaries}), which converts expectations under the trajectory law into expectations under a single distribution on $\mathcal{S}\times\mathcal{A}$. This is the step where the imitation problem becomes distributional: instead of asking whether two policies take the same action in the same state, we ask whether they induce the same long-run visitation pattern. In maximum-entropy RL/IRL, one adds the discounted \emph{causal entropy} \citep{bloem2014infinite}
\[
H(\pi)
\;=\;
\E_\pi\!\Big[\sum_{t=0}^{\infty}\gamma^t\big(-\log \pi(a_t\mid s_t)\big)\Big]
\;=\;
\frac{1}{1-\gamma}\,\E_{(s,a)\sim\rho_\pi}\!\big[-\log \pi(a\mid s)\big],
\]
which discourages overly deterministic policies and yields a well-posed optimisation problem.

A regularised maximum-entropy IRL problem can be written \citep{ho2016generative} as
\[
\mathrm{IRL}_{\psi}(\pi_E)
=\argmax_{\ell}\Big\{
-\psi(\ell)
+\min_{\pi}\big[-H(\pi)+\E_{(s,a)\sim\rho_\pi}[\ell(s,a)]\big]
-\E_{(s,a)\sim\rho_{\pi_E}}[\ell(s,a)]
\Big\},
\]
with a convex regulariser $\psi$ on costs. The inner minimisation is the entropy-regularised control problem for a fixed $\ell$:
\[
\pi^\star(\ell)\in\argmin_{\pi}\Big\{-H(\pi)+\E_{(s,a)\sim\rho_\pi}[\ell(s,a)]\Big\}.
\]
The standard max-ent IRL pipeline first finds a cost $\ell^\star$ via the outer maximisation and then outputs $\pi^\star(\ell^\star)$. Eliminating $\ell$ by convex duality yields the equivalent occupancy-matching form
\begin{equation}
\pi^\star\;\in\;\argmin_{\pi}\Big\{-H(\pi)+\psi^{\ast}\big(\rho_\pi-\rho_{\pi_E}\big)\Big\},
\label{eq:occ-match}
\end{equation}
which makes explicit that imitation reduces to matching \emph{occupancy measures} and that the choice of discrepancy $\psi^\ast$ (the convex conjugate of $\psi$) is central. This is exactly the point where OT can be inserted: once imitation is ``matching these two distributions,'' the next question is \emph{which} discrepancy should be used to define the match.

\emph{Adversarial imitation learning} can be viewed as an algorithmic instantiation of this occupancy-matching objective: it chooses a variational form for $\psi^\ast$ and optimises it using a learned discriminator/critic from samples of expert and learner behaviour. Generative Adversarial Imitation Learning (GAIL) is the canonical example, corresponding to an $f$-divergence choice (JS) for $\psi^\ast$ implemented via an adversarial discriminator \citep{goodfellow2014generative,ho2016generative}.

\subsection{Why optimal transport helps}
\label{sec:il-whyot}

The occupancy view makes clear why OT is useful in imitation learning. The learner must bring $\rho_\pi$ close to $\rho_{\pi_E}$, but early in training these distributions often overlap very little. As discussed in Section~\ref{sec:ot}, $f$-divergences become weak or unstable in exactly this regime, whereas OT still measures how much probability mass must move, and how far, under a task-dependent ground cost. That ground cost may live on $\mathcal{S}\times\mathcal{A}$, on $\mathcal{S}$ alone in learning-from-observation settings, or in a learned feature space. In each case it determines what the algorithm treats as a meaningful notion of behavioural similarity. Put differently, OT does not just say ``the learner is wrong''; it says \emph{how} the learner is wrong in the geometry of the task.

The rest of the section follows two questions: \emph{which distributions are being compared}, and \emph{how does the OT signal enter learning}? This leads to the following recurring patterns. A reader can think of them as different \textcolor{black}{ways in which OT enters the learning problem: as an online reward signal, as an offline relabelling step, through the coupling constraints, or through the representation used for matching.}
\begin{itemize}
\item \emph{Occupancy matching with OT discrepancies.} Replace the JS/KL discrepancy in adversarial IL by a Wasserstein objective (typically $\Wone$ via a 1-Lipschitz critic) or by a primal entropy-regularised objective, yielding geometry-aware shaping rewards.
\item \emph{Reward labelling from couplings (offline).} Compute a transport plan between empirical measures built from learner and expert trajectories and convert the resulting coupling into per-state (or per-transition) pseudo-rewards, enabling standard offline RL without an online min-max loop for reward learning in standard IRL. With multiple expert demonstrations, multi-marginal/barycentric aggregation can avoid the noise induced by naive concatenation \citep{sebag2023combining}.
\item \emph{Temporal and intent-aware matching.} Modify admissible couplings (e.g.\ banded/time-window constraints) and/or modify the ground cost (e.g.\ intent/successor-feature embeddings) to reduce spurious matches caused by treating trajectories as unordered bags of states.
\item \emph{Cross-domain matching.} When a pointwise cross-domain ground cost is ill-posed, use relational objectives such as Gromov-Wasserstein that match within-domain geometry rather than raw observations.
\end{itemize}

\subsection{Reward shaping in the online adversarial setting}
\label{sec:il-adversarial}

The max-ent IRL derivation in Section~\ref{sec:il-background} poses imitation as occupancy matching~\eqref{eq:occ-match}, where $\psi^\ast$ measures the gap between learner and expert occupancies \citep{ho2016generative}. Adversarial imitation learning turns this variationally into a learned critic and uses the critic output as a shaping reward for policy optimisation. GAIL is the canonical example: it chooses $\psi^\ast$ so that the induced discrepancy is an $f$-divergence, specifically JS, implemented with an adversarial discriminator \citep{ho2016generative}. OT-based variants keep this overall template but replace the underlying discrepancy with a Wasserstein objective, or with a tractable OT-inspired approximation, so that the reward signal remains informative under weak overlap and can reflect task geometry. The practical picture is simple: OT is used to turn the different behaviour between expert and learner into a dense reward-like signal that the policy can improve against.

A useful way to organise the OT-based occupancy-matching literature is by \emph{how the Wasserstein-type discrepancy is computed}.

\paragraph{Dual (critic-based) $\Wone$: Kantorovich potentials as shaping rewards.}
When the discrepancy is $\Wone$ on $\mathcal{S}\times\mathcal{A}$, Kantorovich-Rubinstein duality (Section~\ref{sec:ot}) expresses $\Wone(\rho_\pi,\rho_{\pi_E})$ as a supremum over $1$-Lipschitz critics $f$, suggesting a WGAN-style scheme \citep{arjovsky2017wasserstein}: train $f$ to maximise the dual gap between expert and learner occupancies, then use $\tilde r(s,a)=f(s,a)$ as a shaping reward. Wasserstein Adversarial Imitation Learning (WAIL) instantiates this idea and approximates the \(W_1\) dual by replacing the hard 1-Lipschitz constraint with a smooth penalty $\Omega_{d,\varepsilon}$ \citep{xiao2019wasserstein}:
\[
\Wone(\rho_\pi,\rho_{\pi_E})
\;\textcolor{black}{\approx}\;
\sup_{f:\mathcal{S}\times\mathcal{A}\to\R}
\Big\{
\E_{y\sim\rho_{\pi_E}}[f(y)]-\E_{x\sim\rho_{\pi}}[f(x)]
+
\E_{(x,y)\sim\rho_\pi\times\rho_{\pi_E}}\big[\Omega_{d,\varepsilon}(f,x,y)\big]
\Big\}.
\]
\textcolor{black}{\citet{xiao2019wasserstein} give two choices of regulariser; the $L_2$ form used throughout their experiments is
$$\Omega_{d,\varepsilon}(f,x,y)=-\frac{\big[f(y)-f(x)-d(x,y)\big]_+^2}{4\varepsilon},$$
which is nonpositive, vanishes exactly on pairs satisfying the Lipschitz inequality, and so penalises violations with sharpness set by $\varepsilon$; their entropic alternative is also nonpositive but does not vanish on the feasible set. The plus sign in the objective is therefore correct. The relation is an approximation rather than an identity: for finite $\varepsilon$ the penalised supremum is a regularised surrogate for the Kantorovich-Rubinstein dual.} The practical requirement in dual-$\Wone$ approaches is stable enforcement (or relaxation) of 1-Lipschitzness, commonly via weight clipping or gradient penalties \citep{arjovsky2017wasserstein,gulrajani2017improved}. Follow-up works explore alternative design choices and application settings (e.g.\ different shaping rewards, state-only matching or latent-space matching); \citet{zhang2020wasserstein,wu2023w, yang2026latent} are representative examples.

\paragraph{Primal entropic OT (Sinkhorn): coupling-derived shaping rewards.}
An alternative is to compute an OT plan directly in the primal using entropic regularisation and Sinkhorn iterations \citep{cuturi2013sinkhorn,genevay2018learning}. Sinkhorn Imitation Learning (SIL) \citep{papagiannis2022imitation} formulates imitation as minimising an entropy-regularised Wasserstein distance $\Wassp{\varepsilon}$ between expert and learner occupancy measures and uses the resulting coupling to derive a dense shaping reward. The main advantage of this route is conceptual as well as computational: one gets not just a scalar discrepancy, but an explicit soft matching between learner and expert samples. A key modelling choice is the ground cost: SIL defines a cosine dissimilarity in an adversarially learned feature space, so the cost matrix is both geometry-aware and discriminative. The feature-space parameters are trained by maximising $\Wassp{\varepsilon}$ between learner and expert occupancies, yielding a minimax objective analogous to GAIL but with a Sinkhorn-based discrepancy \citep{papagiannis2022imitation}. Relative to dual-$\Wone$ critics, the entropic primal objective provides smooth, minibatch-friendly estimates of the discrepancy and an explicit transport plan; in exchange, performance depends on scaling the entropic parameter and the feature-space cost appropriately.

\paragraph{Computational approximation: greedy primal couplings.}
Some methods try to reduce the cost of repeatedly solving OT subproblems.

Primal Wasserstein Imitation Learning (PWIL) avoids adversarial training and approximates the primal OT coupling by a greedy strategy that yields an upper bound on the true Wasserstein cost \citep{dadashi2020primal}. At each step, the agent’s current state-action is matched to nearby expert samples subject to remaining ``mass'', producing a non-stationary cost $c^{g}_{t,\pi}$ and a shaping reward $\tilde r_{t,\pi}=f(c^{g}_{t,\pi})$ for each state-action pair with a monotone decreasing function $f$ \citep{dadashi2020primal}. This preserves the geometric ``reward-from-transport'' intuition while avoiding a min-max loop, at the expense of a history-dependent reward induced by the evolving greedy coupling.

All three approaches pursue the same goal, namely to turn transport-based distribution matching into a usable reward signal, but they do so in different computational regimes. Dual-$\Wone$ critics are conceptually simple but live or die by Lipschitz control. Sinkhorn-primal methods give an explicit coupling and behave well on minibatches, but introduce the usual sensitivity to the entropic parameter. Greedy couplings such as PWIL avoid adversarial training altogether, at the price of a history-dependent reward.

\subsection{Offline reward labelling}
\label{sec:il-otr}

Instead of learning a shaping signal online through adversarial occupancy matching, one can compute pseudo-rewards \emph{offline}. The idea is to align trajectories in an unlabeled dataset with expert demonstrations using OT, then read per-time-step labels from the resulting coupling and pass the relabelled dataset to a standard offline RL algorithm. The important shift is that OT is now solved between \emph{empirical measures} built from finite trajectories, rather than between discounted occupancies \citep{luo2023optimal}. In effect, the transport plan is being used as a soft correspondence map: it says which parts of a learner trajectory resemble which parts of an expert trajectory.

Concretely, consider an offline dataset $\mathcal{D}$ consisting of transitions $(s,a,s')$ (with rewards unavailable or ignored) and a set of expert demonstrations. For a single episode in $\mathcal{D}$ we write
\[
\tau=\bigl((s_t,a_t,s_{t+1})\bigr)_{t=0}^{T-1},
\]
and denote by $(s_0,\dots,s_{T-1})$ the associated state sequence on which the OT problem is constructed. \textcolor{black}{Note that an episode written this way contains $T$ transitions but only $T$ state observations $(s_0,\dots,s_{T-1})$ within the window used for transport. The transition-pair measures introduced below are formed from the $T-1$ consecutive pairs inside that state sequence, which is why they are normalised by $1/(T-1)$ rather than $1/T$.} An expert demonstration may be available either as a full trajectory (including actions) or as an observation-only sequence; in both cases we denote the expert state sequence by $(s^E_0,\dots,s^E_{T_E-1})$.

Optimal Transport Reward labeling (OTR) \citep{luo2023optimal} constructs empirical state measures
\[
\hat p_{\pi}=\frac{1}{T}\sum_{t=0}^{T-1}\delta_{s_t},
\qquad
\hat p_{E}=\frac{1}{T_E}\sum_{t'=0}^{T_E-1}\delta_{s^E_{t'}},
\]
and, given a ground cost $c:\mathcal{S}\times\mathcal{S}\to\R_{\ge 0}$ (e.g.\ cosine distance on raw state vectors), computes an optimal coupling
\[
\Gamma^\ast
=\argmin_{\Gamma\in\Pi(\hat p_{\pi},\hat p_{E})}
\sum_{t=0}^{T-1}\sum_{t'=0}^{T_E-1} c(s_t,s^E_{t'})\,\Gamma_{t,t'},
\qquad
\Pi(\hat p_{\pi},\hat p_{E})
=\bigl\{\Gamma\in\textcolor{black}{\R_{+}^{T\times T_E}}:\ \Gamma\mathbf{1}=\tfrac{1}{T}\mathbf{1},\ \Gamma^\top\mathbf{1}=\tfrac{1}{T_E}\mathbf{1}\bigr\}.
\]
A per-step pseudo-reward is then assigned by aggregating transport costs under $\Gamma^\ast$:
\[
\tilde r_{\mathrm{OT}}(s_t)\;=\;-\sum_{t'=0}^{T_E-1} c\big(s_t,s^E_{t'}\big)\,\Gamma^\ast_{t,t'}.
\]
Computing $\Gamma^\ast$ exactly is a linear program; in practice, OTR solves an entropy-regularised OT problem and uses Sinkhorn iterations for efficiency \citep{luo2023optimal,cuturi2013sinkhorn}. 
After squashing through an exponential function (e.g. $s(r)=5\exp(Tr)$), rewards are rescaled and independent of the horizon.
These labels are attached to the offline transitions at the corresponding time steps (e.g.\ $(s_t,a_t,s_{t+1})\mapsto s(\tilde r_{\mathrm{OT}}(s_t))$) and then used by any downstream RL method \citep{haldar2023watch, wang2023offline, zare2023leveraging}.
\textcolor{black}{OTR demonstrates its ability to shape informative rewards from datasets with diverse behaviours by achieving more robust performance than competing baselines, without performance degradation across experiments \citep{luo2023optimal}.}

\paragraph{Learning from observation.}
When expert actions are unavailable (imitation learning from observation, ILfO), \citet{chang2023imitation} argue that state-only matching can be problematic in some environments depending on the state representation and dynamics, and treat trajectories as bags of \emph{transitions} rather than states, defining empirical measures $\hat p_{\pi}^{\rightarrow}=\frac{1}{T-1}\sum_{t}\delta_{(s_t,s_{t+1})}$ and $\hat p_{E}^{\rightarrow}=\frac{1}{T_E-1}\sum_{t'}\delta_{(s^E_{t'},s^E_{t'+1})}$ on $\mathcal{S}\times\mathcal{S}$ and solving OT with a ground metric $d$ on transition pairs to yield per-transition pseudo-rewards $\tilde r_{\mathrm{OT}}(s_t,s_{t+1})=-\sum_{t'} d\big((s_t,s_{t+1}),(s^E_{t'},s^E_{t'+1})\big)\,\Gamma^{\rightarrow\ast}_{t,t'}$. They also show that under a monotonic-improvement assumption, reusing pseudo-rewards from previously collected trajectories yields an upper bound on the current Wasserstein objective, enabling systematic off-policy reuse. 

In offline learning-from-observations (LfO) where interaction is disallowed, Distribution Correction Estimation (DICE) methods \citep{nachum2019dualdice} reformulate occupancy matching via a change of variables that avoids explicit policy evaluation, instead learning density-ratio corrections directly from data. They seek a policy whose \emph{state} occupancy $\rho_\pi(s)=\sum_a\rho_\pi(s,a)$ matches an empirical expert state occupancy estimated from expert state-only trajectories. Primal Wasserstein DICE (PW-DICE) \citep{yan2024pwdice} proposes to minimise the Kantorovich primal~\eqref{eq:kantorovich} on state occupancies, $\Wone(\rho_\pi(s),\rho_{\pi_E}(s))$, jointly over a feasible occupancy induced by $\pi$ and a coupling $\Gamma$. Feasibility is enforced via the Bellman-flow constraint~\eqref{eq:flow-constraint} on $\rho_\pi$, together with $\rho_\pi(s)=\sum_a\rho_\pi(s,a)$ and nonnegativity.
The main reason PW-DICE adopts the primal view is the flexibility of the ground cost. Dual-$\Wone$ methods usually enforce Lipschitzness through Euclidean gradient penalties, which quietly ties the geometry to Euclidean structure. The primal formulation instead allows arbitrary costs $c(s,s')$, including learned reachability-aware metrics \citep{yan2024pwdice}. On the optimisation side, PW-DICE adds pessimistic regularisers, implemented with KL-type penalties, to obtain a smooth convex dual objective, then recovers the policy by weighted behaviour cloning. The learned dual variables play a role analogous to rewards in offline reward-labelling methods from Section~\ref{sec:il-otr}. The same formulation also shows that, for particular hyperparameter choices, familiar $f$-divergence DICE variants appear as special cases, which gives a clean unifying view of Wasserstein and $f$-divergence occupancy matching in offline LfO \citep{yan2024pwdice}. 

Two limitations motivate the extensions in Section~\ref{sec:temporal}. First, state-marginal transport ignores order and can couple reversed or speed-mismatched rollouts. Second, the pseudo-rewards are context-dependent: $\tilde r_{\mathrm{OT}}(s_t)$ depends on the whole trajectory through $\Gamma^\ast$, so the same state may receive different labels across episodes. Put more intuitively, a bag-of-states OT problem can say that two trajectories contain similar ingredients without checking whether those ingredients appear in the same order. Temporal constraints (e.g.\ banded couplings and context-window costs) and intent-aware costs mitigate these issues by biasing the coupling toward locally consistent alignments and control-relevant geometry \citep{fu2024robot,bobrin2024align}.

\paragraph{Multiple expert demonstrations: concatenation vs.\ multi-marginal/barycentric aggregation.}
The construction above is often presented for a single demonstration, but practical IL usually offers a \emph{set} of expert trajectories $\{\tau_E^{(p)}\}_{p=1}^P$, often with clear multi-modality. A common heuristic is simply to pool all expert states into one empirical measure and then subsample. When demonstrations differ substantially, however, this can wash out modes and produce noisy pseudo-rewards. \citet{sebag2023combining} address this problem directly in sliced multi-marginal optimal transport imitation learning (SMMOTIL) by combining demonstrations in OT geometry, via sliced multi-marginal OT distances, rather than by concatenation. Their method keeps the reward-from-transport viewpoint of PWIL but changes how expert information is aggregated.
\citet{sebag2023combining} implement this idea through \emph{sliced} approximations, which reduce the complexity of transport computations by averaging one-dimensional Wasserstein distance over random projections and admit closed-form 1D solvers for discrete measures. Concretely, their multi-marginal variant computes a sliced multi-marginal Monge-Wasserstein discrepancy between the agent rollout and all expert rollouts jointly, which is theoretically equivalent to comparing the agent rollout to the sliced-Wasserstein barycentre of experts with $\Wtwo$. Empirically, this reduces instability and improves performance when expert demonstrations are diverse.

\begin{figure}[t]
\centering
\begin{tikzpicture}[
    dot/.style={circle, fill, inner sep=2.5pt},
    learner/.style={dot, fill=blue!70},
    expert/.style={dot, fill=orange!80},
]

\definecolor{darkblue}{RGB}{0,51,102}

\begin{scope}[shift={(-4.5,0)}]
\node[font=\small\bfseries] at (1.2, 2.4) {(a) Standard OT};

\node[font=\scriptsize] at (-0.8, 1.5) {Learner $\tau$};
\foreach \i/\x in {1/0, 2/0.6, 3/1.2, 4/1.8, 5/2.4} {
    \node[learner] (La\i) at (\x, 1.5) {};
    \node[font=\tiny, above=2pt] at (\x, 1.5) {$s_{\i}$};
}

\node[font=\scriptsize] at (-0.8, 0) {Expert $\tau^E$};
\foreach \i/\x in {1/0, 2/0.6, 3/1.2, 4/1.8, 5/2.4} {
    \node[expert] (Ea\i) at (\x, 0) {};
    \node[font=\tiny, below=2pt] at (\x, 0) {$s^E_{\i}$};
}

\draw[red!70, thick, shorten >=3pt, shorten <=3pt] (La1) -- (Ea5);
\draw[red!70, thick, shorten >=3pt, shorten <=3pt] (La2) -- (Ea3);
\draw[red!70, thick, shorten >=3pt, shorten <=3pt] (La3) -- (Ea1);
\draw[red!70, thick, shorten >=3pt, shorten <=3pt] (La4) -- (Ea2);
\draw[red!70, thick, shorten >=3pt, shorten <=3pt] (La5) -- (Ea4);

\node[font=\tiny, red!70, align=center] at (1.2, -0.8) {cross-temporal\\matches};
\end{scope}

\begin{scope}[shift={(2.5,0)}]
\node[font=\small\bfseries] at (1.2, 2.4) {(b) Temporally-aware OT};

\node[font=\scriptsize] at (-0.8, 1.5) {Learner $\tau$};
\foreach \i/\x in {1/0, 2/0.6, 3/1.2, 4/1.8, 5/2.4} {
    \node[learner] (Lb\i) at (\x, 1.5) {};
    \node[font=\tiny, above=2pt] at (\x, 1.5) {$s_{\i}$};
}

\node[font=\scriptsize] at (-0.8, 0) {Expert $\tau^E$};
\foreach \i/\x in {1/0, 2/0.6, 3/1.2, 4/1.8, 5/2.4} {
    \node[expert] (Eb\i) at (\x, 0) {};
    \node[font=\tiny, below=2pt] at (\x, 0) {$s^E_{\i}$};
}

\draw[darkblue, thick, shorten >=3pt, shorten <=3pt] (Lb1) -- (Eb2);
\draw[darkblue, thick, shorten >=3pt, shorten <=3pt] (Lb2) -- (Eb3);
\draw[darkblue, thick, shorten >=3pt, shorten <=3pt] (Lb3) -- (Eb1);
\draw[darkblue, thick, shorten >=3pt, shorten <=3pt] (Lb4) -- (Eb5);
\draw[darkblue, thick, shorten >=3pt, shorten <=3pt] (Lb5) -- (Eb4);

\node[font=\tiny, darkblue, align=center] at (1.2, -0.8) {$|t - t'| \le 3$\\temporally aligned};
\end{scope}

\end{tikzpicture}
\caption{Standard vs.\ temporally-aware OT matching between learner and expert trajectories. (a)~Without temporal constraints, the coupling can align temporally distant states, producing crossing matches that yield noisy pseudo-rewards. (b)~With a band constraint $|t-t'|\le 3$ (or context-window costs), matches are near-diagonal, reducing spurious alignments while tolerating modest speed variation.}
\label{fig:temporal}
\end{figure}
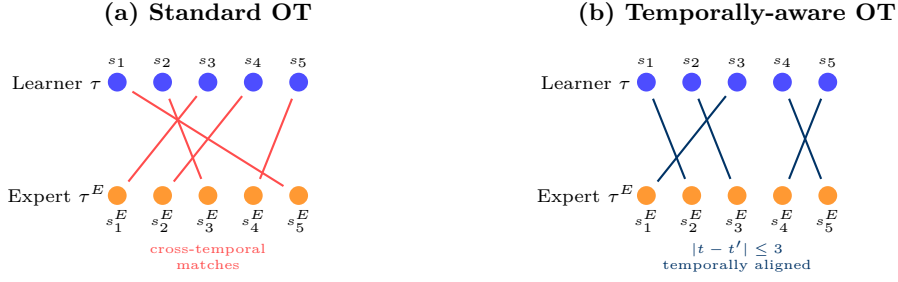

\subsection{Temporal and intent-aware extensions}
\label{sec:temporal}

The trajectory-level construction in the previous subsection treats rollouts as bags of states. This simplifies transport but discards ordering, so the OT plan can align temporally distant (or even reversed) parts of learner and expert trajectories (Figure~\ref{fig:temporal}).
The resulting pseudo-rewards can therefore be noisy and overly dependent on the global trajectory-level alignment. Two complementary fixes have been proposed: (i) enforce approximate time alignment by restricting the coupling to near-diagonal time indices and using windowed (context) costs, and (ii) define the ground cost in embeddings that reflect short-horizon dynamics (``intent''), so matches depend on where the agent is going rather than only what it looks like now.

\paragraph{Temporal masking and context smoothing.}
\textcolor{black}{The methods in this subsection address temporal structure in the coupling itself, and are used \emph{online}: the agent interacts with the environment while a fixed set of demonstrations supplies the transport target, in contrast to the offline relabelling pipeline of Section~\ref{sec:il-otr}.} A first strategy constrains the coupling to respect approximate time alignment. TemporalOT \citep{fu2024robot} proposes two coupled modifications: (i) a \emph{context} cost that compares short windows rather than single frames, and (ii) a \emph{band} constraint that restricts admissible couplings to near-diagonal indices.

Let $(o_0,\dots,o_{T-1})$ and $(o^E_0,\dots,o^E_{T-1})$ be learner and expert observation sequences of equal length, and let $f(o)$ be a fixed visual encoder (TemporalOT targets image-based tasks in partially observable settings; in fully observable MDPs, observations reduce to states). For a context length $k_c$, define the group-wise cosine dissimilarity cost \citep{fu2024robot}
\[
\hat c(o_i,o^E_j)=\frac{1}{k_c}\sum_{h=0}^{k_c-1}\Big(1-\frac{\langle f(o_{i+h}),\,f(o^E_{j+h})\rangle}{\|f(o_{i+h})\|_2\;\|f(o^E_{j+h})\|_2}\Big),
\]
with indices clipped when $i+h$ or $j+h$ exceeds the horizon. For a bandwidth $k_m$, define a binary mask
\[
M(i,j)=\mathbbm{1}\big(|i-j|\le k_m\big),
\]
and restrict the coupling between uniform empirical marginals $\hat p_{\pi}=\frac{1}{T}\sum_{i=0}^{T-1}\delta_{o_i}$ and $\hat p_E=\frac{1}{T}\sum_{j=0}^{T-1}\delta_{o^E_j}$ to the near-diagonal set
\[
\Pi_M(\hat p_{\pi},\hat p_E)=\Big\{\Gamma\in\Pi(\hat p_{\pi},\hat p_E):\ \Gamma_{ij}=0\ \text{whenever}\ M(i,j)=0\Big\}.
\]
TemporalOT then solves an entropic OT problem on this restricted set,
\[
\Gamma^\ast
=\argmin_{\Gamma\in\Pi_M(\hat p_{\pi},\hat p_E)}
\Big\{\sum_{i,j}\Gamma_{ij}\hat c(o_i,o^E_j)-\varepsilon\,\mathcal{H}(\Gamma)\Big\},
\]
which is equivalent to running Sinkhorn with forbidden entries assigned an infinite (or very large) cost. \textcolor{black}{Here $\mathcal{H}(\Gamma)=-\sum_{ij}\Gamma_{ij}\log\Gamma_{ij}$ is the \emph{coupling} entropy, a different
      object from the causal policy entropy $H(\pi)$ of Section~\ref{sec:il-background}. The $-\varepsilon\mathcal{H}(\Gamma)$ convention used here and the $+\varepsilon\KL(\Gamma\|\mu\otimes\nu)$ convention of Section~\ref{sec:ot} differ by a constant determined by the marginals, on $\Pi_M(a,b)$ as on $\Pi(a,b)$, so they select the same plans but report different values.} The resulting $\Gamma^\ast$ can be converted into a dense pseudo-reward by aggregating costs under the coupling, as in Section~\ref{sec:il-otr}; empirically, intermediate $k_m$ and $k_c$ reduce spurious matches while tolerating modest speed variation \citep{fu2024robot}.
The mask gives a temporally informed proxy reward by discouraging cross-time matches, which is not a full dynamic OT formulation.

\paragraph{Intent representations and short-horizon dynamics.}
The banded coupling above addresses \emph{time-index} ambiguity, but it does not resolve a second failure mode: even near-diagonal matches can be spurious if the ground cost compares only instantaneous observations. A complementary strategy is therefore to encode \emph{intent} directly in the representation used for the ground cost, so that similarity reflects short-horizon evolution (``where the agent is going'') rather than appearance alone.

Aligned Imitation Learning via Optimal Transport (AILOT) \citep{bobrin2024align} instantiates this by learning an intent embedding $\varphi:\mathcal{S}\to\R^d$ and defining the OT cost on \emph{paired} intent states separated by a fixed lookahead:
\[
c(s_i,s_{\min(i+k,T-1)},s^E_j,s^E_{\min(j+k,T_E-1)})
=\|\varphi(s_i)-\varphi(s^E_j)\|_2^2
+\|\varphi(s_{\min(i+k,T-1)})-\varphi(s^E_{\min(j+k,T_E-1)})\|_2^2,
\]
where $k$ probes near-future evolution. Given the entropic OT plan $\Gamma^\ast$ between the two trajectories, AILOT defines a dense intrinsic pseudo-reward by exponentially scaling the transport cost aggregated under $\Gamma^\ast$,
\[
\tilde r(s_i)=\alpha\exp\!\Big(-\kappa\,T\sum_{j=j_1}^{T_E-1}\Gamma_{ij}^\ast\,c(s_i,s_{\min(i+k,T-1)},s^E_j,s^E_{\min(j+k,T_E-1)})\Big),\]
\[j_1=\argmin_{j} c(s_0,s_{\min(k,T-1)},s^E_j,s^E_{\min(j+k,T_E-1)}),
\]
where restricting to the ``tail'' $j\ge j_1$ anchors alignment to the closest expert prefix for the given agent start state. Intuitively, the paired-window cost rewards states that not only look similar now but are poised to evolve similarly over the next few steps, improving robustness to nuisance variation and one-step ambiguities. In control terms, this aligns trajectories by \emph{intent} rather than appearance.

\subsection{Cross-domain imitation with Gromov-Wasserstein}
\label{sec:il-pretrain}

The OT machinery used for imitation can also serve cross-domain settings where expert and learner observations live in different spaces, so that any pointwise cross-domain cost is ill-posed. OT-based data filtering for downstream offline RL (e.g. \citealt{lyu2025cross}) is discussed in Section~\ref{sec:transfer}.

\paragraph{Relational matching via GW.}
The trajectory transport constructions above assume a shared observation space and a ground cost $c(s,s')$ that is meaningful across expert and learner domains. In many applications, this assumption fails (different sensors, viewpoints, or embodiments), so any pointwise cross-domain cost is ill-posed or dominated by nuisance variation.

The Gromov-Wasserstein (GW) distance (Section~\ref{sec:ot}) addresses this by matching \emph{relations} rather than points, seeking a coupling $\Gamma$ that preserves intra-domain pairwise distances.
\textcolor{black}{Gromov-Wasserstein Imitation Learning (GWIL)} applies GW to compare expert and learner state-action occupancy measures living in different spaces, and converts the resulting coupling into a pseudo-reward for policy optimisation \citep{fickinger2021cross}. In practice, GW objectives are computationally heavier than standard Wasserstein distances; entropic regularisation and associated solvers provide scalable approximations and are widely used in GW implementations \citep{peyre2016gromov}, building on regularised standard OT \citep{ferradans2014regularized}. Keypoint-guided OT \citep{gu2022keypoint}, which leverages partial correspondences between source and target, offers an alternative route when some cross-domain anchor pairs are available.

\subsection{Summary}
\label{sec:il-discussion}

Imitation learning can be expressed as matching \emph{policy-induced distributions}, most commonly discounted occupancies $\rho_\pi$ (learner) and $\rho_{\pi_E}$ (expert), or empirical trajectory measures in state/observation space. OT is useful in this setting because it remains informative under weak overlap: it measures how far mass must move under a task-dependent ground cost rather than comparing densities pointwise.

Across the methods reviewed in this section, OT enters the learning pipeline in several distinct roles. (i) \emph{Online occupancy matching:} adversarial methods replace the JS/KL-style discriminator with a Wasserstein objective (typically dual $\Wone$ with a 1-Lipschitz critic) or a primal entropic OT/Sinkhorn objective, producing geometry-aware \emph{shaping} rewards. (ii) \emph{Offline reward labelling:} coupling-based methods solve OT between empirical measures built from learner and expert trajectories (states or transitions) and convert the optimal plan $\Gamma^\ast$ into per-step \emph{pseudo-reward} labels, enabling standard offline RL without an online min-max loop. (iii) \emph{Temporal/intent structure:} to avoid spurious alignments caused by treating trajectories as unordered bags of states, extensions either restrict admissible couplings (e.g.\ near-diagonal bands and context windows) or redefine the ground cost in intent/dynamics-aware embeddings. (iv) \emph{Cross-domain imitation:} when a pointwise cross-domain cost is ill-posed, Gromov-Wasserstein provides a well-motivated alternative by matching relational structure under intra-domain metrics. \textcolor{black}{Table~\ref{tab:summary3} provides a summary and comparisons of reviewed methods, including the role of OT, the objects being compared, the OT formulation and its implementation.}
The main failure mode is consistent across these roles: a poor ground cost or weak temporal alignment can turn a transport plan into misleading rewards.

\textcolor{black}{Two related uses of OT are discussed in later sections.} OT also quantifies IRL reward ambiguity via Wasserstein dispersion (Section~\ref{sec:reward-ambiguity}), and OT-derived scores can filter or weight data before downstream offline RL (Section~\ref{sec:transfer}). 
In practice, the dominant failure modes are representational---poorly specified ground costs mis-shape behaviour and numerically unstable Lipschitz enforcement or overly diffuse entropic regularisation.

\section{Policy optimisation and regularisation}
\label{sec:policy}

Policy-gradient methods update a parameterised policy by ascending an estimate of the return gradient, and their stability depends heavily on what counts as a ``small'' policy change. Classical algorithms answer that question with density-ratio geometry: Trust Region Policy Optimisation (TRPO) \citep{schulman2015trust} uses a KL-based trust region, while Proximal Policy Optimisation (PPO) \citep{schulman2017proximal} clips likelihood ratios. OT offers a different answer. Instead of comparing policies only through likelihood ratios, it compares the cost of moving probability mass between their induced action or behaviour distributions. The intuition is that not all policy changes of the same probability size are equally meaningful: shifting mass from one action to a very similar one should count as a smaller change than shifting it to a qualitatively different action. 
Revisiting the previous robotic pick-and-place task example, a policy update (resulting in a different action for the given state) that changes the end-effector displacement slightly should be considered smaller than one that moves the end effector in the opposite direction, even if the probability changes are identical. KL-based trust regions treat both updates according to changes in action probabilities, whereas OT additionally accounts for the geometric distance between actions.

\begin{figure}[t]
\centering
\begin{tikzpicture}[
    node distance=1.2cm and 1.8cm,
    rlbox/.style={rectangle, draw, rounded corners=3pt, minimum width=2cm,
        minimum height=0.7cm, align=center, font=\small, fill=gray!8},
    otbox/.style={rectangle, draw, rounded corners=3pt, minimum width=2cm,
        minimum height=0.7cm, align=center, font=\small,
        draw=darkblue, text=darkblue, fill=blue!5},
    arr/.style={-{Stealth[length=1.5mm]}, semithick, gray!60},
    otarr/.style={<->, semithick, darkblue, dashed},
    otlbl/.style={font=\scriptsize, text=darkblue, align=center}
]

\definecolor{darkblue}{RGB}{0,51,102}

\node[rlbox] (policy) {Policy\\[-1pt]{\scriptsize $\pi_\theta$}};

\node[otbox, left=2.5cm of policy] (ref) {Reference\\[-1pt]{\scriptsize $\pi_{\mathrm{old}}/\pi_{\mathrm{ref}}$}};
\draw[otarr] (ref) -- (policy);
\node[otlbl, above=2pt] at ($(ref)!0.5!(policy)$) {$\Wass(\pi,\pi_{\mathrm{old}})$};
\node[otlbl, below=2pt] at ($(ref)!0.5!(policy)$) {trust region/\\gradient flow};

\node[rlbox, right=2.5cm of policy, yshift=1cm] (critic) {Critic\\[-1pt]{\scriptsize $Q_{\theta_q}, V_{\theta_v}$}};
\draw[arr] ([yshift=3mm]critic) -- ([yshift=3mm]policy);
\draw[arr] ([yshift=-3mm]policy) -- ([yshift=-3mm]critic);

\node[rlbox, right=2.5cm of policy, yshift=-1cm] (zdist) {Return distribution\\[-1pt]{\scriptsize $Z^\pi(s,a)$}};
\draw[arr] ([yshift=3mm]zdist) -- ([yshift=3mm]policy);
\draw[arr] ([yshift=-3mm]policy) -- ([yshift=-3mm]zdist);

\node[otlbl, above=2pt] at ($(zdist)!0.5!(policy)$) {$\Wass(Z^\pi, \mathcal{T}^{\pi}Z^\pi)$};
\node[otlbl, above=2pt] at ($(critic)!0.5!(policy)$) {$\Wass(Q_t, Q_{t+1})$};

\end{tikzpicture}
\caption{OT in policy optimisation and regularisation (Section~\ref{sec:policy}). Wasserstein trust regions constrain per-state action distributions or trajectory embeddings to stay close to a reference policy (Sections~\ref{sec:policy-trust}--\ref{sec:policy-wgf}). Wasserstein-trained critics use OT losses on return or Q-value distributions (Section~\ref{sec:policy-critics}). The blue dashed bidirectional arrow connects two distributions compared by OT, while grey solid arrows represent connections in the RL algorithm.}
\label{fig:policyoptimisation}
\end{figure}
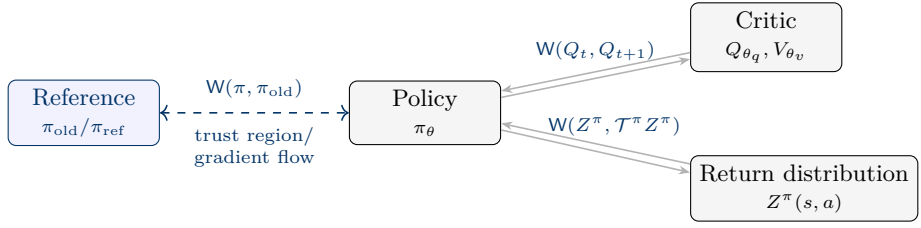

In this section, we follow that idea through four settings: trust regions on action distributions, regularisation in latent behaviour spaces, Wasserstein gradient-flow and natural-gradient updates, and critic training based on Wasserstein losses (Figure~\ref{fig:policyoptimisation}).

\subsection{Classical policy optimisation}
\label{sec:policy-setup}

We begin with the standard policy-optimisation setup, because the OT variants are easiest to understand as direct modifications of familiar trust-region and actor-critic updates. The goal of this subsection is not to re-derive policy gradients in full, but to isolate the place where ``geometry'' enters the update.

Let $\pi$ be the current policy and let $\rho_{\pi}$ denote the discounted state-action visitation measure induced by $\pi$.
The control objective~\eqref{eq:rl_obj}, recalled from Section~\ref{sec:preliminaries}, expresses $J(\pi)$ as an expectation under the discounted occupancy measure~$\rho_\pi$. The prefactor $\frac{1}{1-\gamma}$ is constant and can be suppressed when discussing optimisation templates. In policy-gradient methods, one considers a parameterised family $\{\pi_{\theta}\}$ and updates $\theta$ in the direction of the gradient $\nabla_\theta J(\pi_{\theta})$. The policy gradient theorem shows that this gradient can be expressed in terms of an \emph{advantage function} $A^{\pi}$ as
\[
\nabla_\theta J(\pi_{\theta})
\;=\;\frac{1}{1-\gamma}
\E_{s\sim\rho_{\pi_{\theta}},\,a\sim \pi_{\theta}(\cdot\mid s)}
\big[\nabla_\theta \log \pi_{\theta}(a\mid s)\,A^{\pi_{\theta}}(s,a)\big],
\]
where $A^{\pi}(s,a)$ measures how much better action $a$ is, in state $s$, than the policy’s default behaviour in that state \citep{sutton1998reinforcement}. In practice, one replaces $A^{\pi_{\theta}}$ by an estimate $\hat A$ obtained from rollouts and takes stochastic gradient steps.

It is helpful to introduce a candidate next policy $\pi'$ and ask how much expected advantage it gains under the \emph{state} distribution of the current policy $\pi$. This leads to the surrogate objective
\begin{equation}
    \label{eq:policy-surrogate}
    \E_{s\sim \rho_{\pi},\,a\sim \pi'(\cdot\mid s)}\!\big[A^{\pi}(s,a)\big],
\end{equation}
where $A^{\pi}$ is the advantage under the current policy $\pi$. The expression \eqref{eq:policy-surrogate} keeps the state distribution fixed at what $\pi$ currently visits, and chooses $\pi'$ so that it places more probability mass on actions with positive advantage and less on actions with negative advantage. This is why it is called a \emph{surrogate}: it is a local stand-in for the true return objective, using information that is reliable near the current policy. One can show that, for sufficiently small changes in the policy, maximising \eqref{eq:policy-surrogate} provides a first-order approximation to improving $J(\pi')$ itself \citep{schulman2015trust}.

If \eqref{eq:policy-surrogate} is maximised without any restriction, the resulting policy $\pi'$ can differ arbitrarily from $\pi$ after a single update. In practice, this often leads to instability: the advantage estimates are only reliable near the behaviour that generated the data, and large updates can push the agent into regions where those estimates are inaccurate. Trust-region methods address this by explicitly requiring that, in each state, the new action distribution $\pi'(\cdot\mid s)$ remains close to the old one $\pi(\cdot\mid s)$. A trust region is therefore best thought of as a local safety mechanism: improve the policy, but only by a step whose size is small enough that the local approximation still makes sense.

TRPO \citep{schulman2015trust} enforces this by constraining the average KL divergence between the two policies,
\[
\E_{s\sim \rho^{\pi}_{\nu}}\!\Big[\KL\big(\pi(\cdot\mid s)\,\|\,\pi'(\cdot\mid s)\big)\Big]\ \le\ \delta,
\]
for some step-size parameter $\delta>0$, where $\rho^{\pi}_{\nu}$ denotes the discounted state-visitation distribution induced by policy $\pi$ from initial-state distribution $\nu$. \textcolor{black}{The monotonic-improvement result of \citet{schulman2015trust} follows from improving a surrogate lower bound on $J(\pi')$ carrying a calibrated penalty on $\max_s \KL(\pi(\cdot\mid s)\|\pi'(\cdot\mid s))$. The constrained updates actually run are approximations to that bound: first a hard maximum-KL constraint, then the displayed average-KL constraint, which \citet{schulman2015trust} describe as a heuristic. Maximising~\eqref{eq:policy-surrogate} under the average constraint alone therefore carries no monotonicity guarantee, and the same caveat applies to the Wasserstein trust regions below whenever an expected-discrepancy constraint replaces a state-wise one.} PPO \citep{schulman2017proximal} takes a more heuristic approach: it modifies the objective so that, for each state and action, the contribution to the gradient is clipped when the ratio $\pi'(\cdot\mid s)/\pi(\cdot\mid s)$ becomes too large or too small. \textcolor{black}{Clipping changes the \emph{objective} rather than the feasible set: \textcolor{black}{it removes the direct incentive to push a ratio beyond the clipping range in the favourable direction for an individual sample,}
but it imposes no hard bound on the likelihood ratios or log-probability changes that optimisation can actually produce.} It plays a role similar to a KL-based trust region. \citet{schulman2017proximal} also propose the KL penalty with an adaptive coefficient as an alternative.

In actor-critic methods, a learned critic is used to approximate the value functions that appear in the advantage. With a parametric value function $V_{\theta_v}$ as critic, it is trained with the loss function
\[\mathcal{L}(\theta_v)=\mathbb{E}_\pi[(y_t-V_{\theta_v}(s_t))^2]\]
to predict the temporal-difference (TD) target
\[
y_t=r_t+\gamma\,V_{\bar{\theta}_v}(s_{t+1}),
\]
according to the Bellman equation~\eqref{eq:bellman}, where $V_{\bar{\theta}_v}$ is the target with parameters updated slowly to enhance learning stability. 
Advantage estimates $\hat A_t$ are then constructed, for example via generalised advantage estimation. The policy gradient is then approximated by replacing $A^{\pi}(s,a)$ with $\hat A_t$ in the surrogate objective~\eqref{eq:policy-surrogate}, so that the actor update is driven directly by the critic’s outputs. In this setting, stability depends both on the size of the policy step and on how smoothly the critic and its targets evolve.

In all of these algorithms, the notion of a ``small'' policy update is already present; OT simply changes the geometry used to measure it. TRPO makes the notion explicit with an average KL constraint, while PPO approximates it through clipped likelihood ratios. The remainder of the section keeps this basic optimisation template but replaces KL-style geometry with optimal transport on action distributions, behavioural embeddings, or value distributions.

\subsection{Wasserstein trust regions for policies}
\label{sec:policy-trust}

The most direct OT analogue of TRPO is to replace the KL trust region with a Wasserstein trust region on per-state action distributions.

Let $c:\mathcal{A}\times\mathcal{A}\to\R_{\ge 0}$ be a ground cost on the action space, and let $\Wass(\cdot,\cdot)$ denote the induced Wasserstein or entropy-regularised Wasserstein distance between probability measures on $\mathcal{A}$ \citep{cuturi2013sinkhorn,genevay2018learning}. Recall from Section~\ref{sec:preliminaries} that $\Wass(\mu,\nu)$ measures the minimal cost of transporting the mass of $\mu$ onto the mass of $\nu$ when moving a unit of mass from $a$ to $a'$ incurs cost $c(a,a')$. For a fixed state $s$, the discrepancy between $\pi(\cdot\mid s)$ and $\pi'(\cdot\mid s)$ can therefore be measured by $\Wass\big(\pi(\cdot\mid s),\pi'(\cdot\mid s)\big)$ and used in place of KL in the trust-region constraint.

The analogue of \eqref{eq:policy-surrogate} with an OT trust region is
\begin{equation}
    \label{eq:ot-trust-region}
    \max_{\pi'}\ \E_{s\sim \rho^{\pi}_{\nu},\,a\sim \pi'(\cdot\mid s)}[A^{\pi}(s,a)]
    \quad\text{s.t.}\quad
    \E_{s\sim \rho^{\pi}_{\nu}}\, \Wass\!\left(\pi'(\cdot\mid s),\,\pi(\cdot\mid s)\right) \le \delta.
\end{equation}
The constraint in \eqref{eq:ot-trust-region} requires that, on average over states visited by $\pi$, the new policy $\pi'$ remains inside a Wasserstein ball of radius $\delta$ around the old policy. 
A related problem is, for $\lambda\ge 0$, to consider a Wasserstein penalty:
\[
\max_{\pi'(\cdot\mid s)}\ \E_{a\sim \pi'(\cdot\mid s)}[A^{\pi}(s,a)]
\;-\;\lambda\,\Wass\!\left(\pi'(\cdot\mid s),\,\pi(\cdot\mid s)\right).
\]
In this formulation one explicitly balances improving $\bbE_{a\sim\pi^\prime(\cdot|s)}[A^\pi(s,a)]$ with staying close to $\pi$.

In discrete action spaces, these state-wise subproblems can be handled explicitly. Fix a state $s$ and denote $p(\cdot)=\pi(\cdot\mid s)$ and $p'(\cdot)=\pi'(\cdot\mid s)$ as probability vectors over the finite action set. \citet{song2023provably} consider the optimisation
\[
\max_{p'(\cdot)}\ \sum_{a} p'(a)\,A^{\pi}(s,a)
\quad\text{s.t.}\quad
\Wass\!\left(p'(\cdot),p(\cdot)\right)\le \delta,
\]
with $\Wass$ chosen as either a $1$-Wasserstein distance or an entropy-regularised approximation. They derive an explicit policy update conditional on the optimal dual variable, which itself is obtained from a one-dimensional optimisation problem and does not require specifying a particular parametric form for $\pi'$. The resulting update redistributes probability mass from low-advantage to high-advantage actions while paying a transport cost determined by $c(a,a')$.
For a reader used to KL trust regions, the key difference is that the update is not only asking ``how much probability should move?'', but also ``between which actions is that movement cheap enough to allow?''

\textcolor{black}{Their Wasserstein Policy Optimisation (WPO) and Sinkhorn Policy Optimisation (SPO) algorithms implement these per-state updates across batches of states.} \textcolor{black}{\citet{song2023provably} prove that WPO admits a monotonic-improvement guarantee with exact advantages, together with a performance lower bound that, when an inaccurate advantage estimate $\widehat A^{\pi}$ with error $\epsilon$ is used, carries an explicit degradation term $-2\epsilon/(1-\gamma)$ (their Theorem~4). Monotone improvement under approximate advantages therefore requires the improvement term to dominate that error, rather than following from boundedness alone. They further show that SPO converges to WPO as the entropic regulariser vanishes, and that both converge to the globally optimal policy \textcolor{black}{value} in the tabular setting under a decaying trust-region multiplier $\beta_k\to 0$, assuming finite state and action spaces, non-negative rewards, and an initial distribution supported on all states.}

Compared to a KL ball, a Wasserstein ball makes the notion of a ``small'' policy change depend explicitly on the ground cost: moving probability mass between similar actions incurs a low cost, while moving mass between very different actions incurs a high cost. This allows domain knowledge about the action space (for example, which torques or control inputs are close in effect) to enter directly through $c$. For a practitioner, this is the main attraction of Wasserstein trust regions: the constraint can reflect the physical or semantic structure of the action space, rather than only probability ratios.

In continuous action spaces, the constraint in \eqref{eq:ot-trust-region} becomes an infinite-dimensional problem over couplings between densities. \citet{terpin2022trust} study trust regions defined by optimal transport discrepancies (which include Wasserstein distances) and propose OT-TRPO, an extension of TRPO to continuous actions. 
When policies are Gaussian and the ground cost is quadratic, $\Wtwo$ has a closed-form expression in terms of means and covariances. Using convex duality, they show that the constrained optimisation problem can be reduced
to a one-dimensional convex optimisation over the Lagrange multiplier associated with the OT constraint, \textcolor{black}{and they derive a closed-form expression for the optimal \emph{nonparametric} policy update as a function of that multiplier, without ever forming a transport plan explicitly. This closed form should be kept separate from the parametric algorithm actually implemented. The Gaussian instantiation of OT-TRPO uses a policy with parametrised mean and \emph{fixed} variance, and updates it by gradient ascent on a surrogate built from the dual solution, rather than by an analytic joint update of means and covariances. Its per-iteration cost is accordingly modest: advantage estimation and \textcolor{black}{each evaluation of the scalar dual objective} are linear in the number of collected samples, and the parameter update is a gradient step, which \citet{terpin2022trust} report as leaving computational cost in line with the baselines they compare against.}

Beyond finite action spaces, \citet{song2023provably} also discuss how to extend Wasserstein trust regions to continuous actions without introducing a fixed hand-crafted discretisation of the action space. Their approach first solves the WPO/SPO target distribution on a finite sampled support set drawn from the current policy, then trains an Implicit Quantile Network (IQN; \citealt{dabney2018implicit}) to match this target via quantile regression. The IQN serves as a continuous generative model for the updated policy, allowing WPO/SPO-style updates to be applied in settings where the action space is large or continuous.

Wasserstein Policy Regularisation (WPR; \citealt{na2026semantic}) brings the same trust-region intuition to reinforcement learning from human feedback (RLHF), but with a different anchor point. Instead of keeping the policy close to the previous iterate, WPR keeps it semantically close to a reference policy under an entropy-regularised Wasserstein penalty. A dual-variable reformulation makes that penalty compatible with PPO-style clipped updates. A complementary distributionally robust optimisation (DRO) view of RLHF alignment appears in Section~\ref{sec:robust-rlhf}.

More recent work views Wasserstein trust regions through the lens of proximal updates and gradient flows. \citet{pfau2025wasserstein} derive a policy optimisation method by discretising a Wasserstein gradient flow of the expected-return functional (see Section~\ref{sec:policy-wgf}). Their gradient-flow derivation yields a qualitatively different update direction from standard policy gradients: actions are weighted by the gradient of the action-value function rather than by a score-function estimator. In practice, however, the trust region in their implemented algorithm is enforced via a KL penalty rather than a Wasserstein constraint, so the theoretical Wasserstein perspective and the practical algorithm differ.

Across these variants, the underlying move is the same: replace the KL geometry of TRPO with a Wasserstein geometry, and measure the size of an update by the transport cost of moving one action distribution into another under the chosen ground cost.

\subsection{Behaviour-guided regularisation in latent spaces}
\label{sec:policy-bgpg}

The per-state action view is not always the right one. In many applications, the relevant object is a whole trajectory, or a learned summary of it, rather than an instantaneous action distribution. Given a behavioural embedding map $\textcolor{black}{\Phi:\mathfrak{T}\to\mathcal{B}_{\mathrm{lat}}}$ that extracts such a summary, each policy $\pi$ induces a \emph{behavioural distribution} $\mu_\pi = \Phi_{\#}P_\pi$ on the latent space $\mathcal{E}$. The workflow is: choose the behavioural aspect that matters, encode trajectories into that space, and compare the resulting distributions there. A generic behaviour-regularised policy optimisation problem is then
\begin{equation}
\label{eq:behaviour-regularised}
\max_{\theta}\ \mathcal{L}(\theta)\;+\;\beta\,\mathcal{R}\big(\mu_{\pi_{\theta}}, \mu_{\mathrm{ref}}\big),
\end{equation}
where $\mathcal{L}(\theta)$ is a standard policy objective, $\mu_{\mathrm{ref}}$ is a reference behavioural distribution, $\mathcal{R}$ is a discrepancy measure, and the sign of $\beta$ determines whether the regulariser attracts (imitation, safety) or repels (exploration, diversity). Non-OT choices for $\mathcal{R}$ include moment penalties, \textcolor{black}{maximum mean discrepancy (MMD)}, or kernel-determinant objectives such as DvD \citep{parkerholder2020effective}. These approaches typically rely on simple moment comparisons or densities and do not exploit the full geometry of the empirical distributions when supports differ or embeddings are not injective.

\citet{pacchiano2020learning} propose to equip the family of behavioural distributions $\{\mu_\pi\}$ with a Wasserstein geometry and to use entropy-regularised Wasserstein distances $\Wassp{\varepsilon}$ (Section~\ref{sec:ot}) as behavioural regularisers. Their starting point is to regard a policy as a distribution over behaviours, and to define the dissimilarity between two policies $\pi$ and $b$ via $\Wassp{\varepsilon}(\mu_\pi,\mu_b)$ under a ground cost $c$ on $\mathcal{E}$. Because $\Wassp{\varepsilon}$ is differentiable and admits a dual formulation in terms of test functions on $\mathcal{E}$, it is amenable to stochastic-gradient methods.

In the Behaviour-Guided Policy Gradient (BGPG) algorithm, the policy objective is
\[
F(\theta)\;=\;\mathcal{L}(\theta)\;+\;\beta\,\Wassp{\varepsilon}\big(\mu_{\pi_{\theta}},\mu_b\big),
\]
with $\mathcal{L}(\theta)$ taken to be a standard surrogate such as \eqref{eq:policy-surrogate} and $b$ a reference policy. The Wasserstein term acts as a behavioural trust region in the latent space: it discourages updates that move the distribution of embeddings $\mu_{\pi_{\theta}}$ too far from $\mu_b$ in a single step. Crucially, the metric is defined on the behaviour space $\mathcal{E}$ rather than on actions, so the regulariser respects the geometry of the chosen behavioural summary. \citet{pacchiano2020learning} show that using smoothed Wasserstein distances is advantageous when the embedding map $\Phi$ is not injective, in which case likelihood-based divergences such as KL are difficult to apply.
The conceptual shift is that two policies are now considered close when they \emph{behave} similarly in the embedding space, even if their action probabilities differ sharply at individual time steps.

Algorithmically, BGPG uses the dual formulation of $\Wassp{\varepsilon}$ to learn \emph{behavioural test functions} on $\mathcal{E}$, which assign scores to embeddings and can be evaluated on trajectories. These test functions play a similar role to discriminators in adversarial methods, but are derived from the OT dual and updated jointly with the policy using stochastic gradient descent in a reproducing-kernel Hilbert space. The scores are then added as shaping terms to the task reward, steering policy optimisation towards or away from specific regions of behaviour space.

A companion algorithm, Behaviour-Guided Evolution Strategies (BGES), applies the same principle to evolution strategies. In BGES, parameter perturbations are sampled from a Gaussian distribution, and each perturbation induces a behavioural distribution over embeddings. The fitness function is augmented with an entropic Wasserstein term that measures the distance between the perturbed policy's behavioural distribution and a reference distribution, so that perturbations producing behaviours closer to the desired reference receive higher fitness scores.

What matters in BGPG and BGES is that OT no longer acts on actions but on \emph{behaviours}. Once the embedding map $\Phi$ and ground cost $c$ on $\mathcal{E}$ are fixed, the Wasserstein term quantifies how much the policy’s behavioural distribution must move to match a reference. This produces a family of regularisers that naturally handles non-injective embeddings and partial behavioural summaries, where action-level KL penalties are often a poor fit.

\subsection{Wasserstein gradient flows and natural gradients}
\label{sec:policy-wgf}

The trust-region and behaviour-guided methods above still start from a standard policy objective and then add OT as a penalty or constraint. A more ambitious viewpoint is to take OT as the underlying \emph{geometry} of policy space itself and derive update rules from that geometry.

To make this precise, it is helpful to think of a policy not only as a parameter vector, but as a probability measure. Depending on the context, this measure can be taken over actions (conditioned on state) or over whole trajectories. Let $\mathcal{X}$ denote the space on which we represent policies as measures (for example, a trajectory space $\textcolor{black}{\mathfrak{T}}$ or a state–action space $\mathcal{S}\times\mathcal{A}$), and let $\mathcal{P}(\mathcal{X})$ be the space of probability measures on $\mathcal{X}$. A functional
\[
F:\mathcal{P}(\mathcal{X})\to\R
\]
assigns a cost to each such measure; for policy optimisation, $F$ could be the negative expected return under the corresponding policy. A Wasserstein gradient flow of $F$ is then a curve $(\mu_t)_{t\ge 0}$ in $\mathcal{P}(\mathcal{X})$ that evolves according to the steepest descent of $F$ with respect to the $2$-Wasserstein metric. A convenient discrete-time characterisation of such flows is given by the Jordan–Kinderlehrer–Otto (JKO) scheme \citep{ambrosio2008gradient}:
\begin{equation}
\label{eq:jko}
\mu_{k+1}=\argmin_{\mu\in\mathcal{P}(\mathcal{X})}\Big\{F(\mu)+\frac{1}{2h}\,\Wtwo^2(\mu,\mu_k)\Big\},
\end{equation}
where $h>0$ is a step size and $\Wtwo$ is the $2$-Wasserstein distance with respect to a ground metric on $\mathcal{X}$. Each step chooses a new measure that decreases $F$ while staying close, in $\Wtwo$, to the previous measure $\mu_k$. \textcolor{black}{When~\eqref{eq:jko} is read as an update on an occupancy measure or a trajectory-policy law rather than on
      an unconstrained measure, the minimisation must be restricted to \emph{realisable} measures, or equivalently
      $F$ must be defined as $+\infty$ outside the feasible set: an arbitrary element of $\mathcal{P}(\mathcal{X})$
      need not satisfy the Bellman-flow constraint~\eqref{eq:flow-constraint} and so need not correspond to any
      policy.} This is the measure-valued analogue of an implicit gradient step, but with distance measured in Wasserstein space rather than in Euclidean space.
One can read~\eqref{eq:jko} as a proximal policy update in which ``do better'' means reducing the objective $F$, while ``do not move too far'' is measured by transport cost between the old and new policy-induced measures.

\citet{zhang2018policy} adapt this construction to RL by applying Wasserstein gradient flows on suitable spaces of policy-related probability measures. In their formulation, $\mu_k$ is the policy parameter distribution or energy-based policy distribution (e.g. per-state action distribution), and the minimisation in \eqref{eq:jko} is achieved by particles with the policy approximated by a parameterised state-conditioned neural network. The energy functional $F$ is defined based on the expected return modelled by a Q-network, and the Wasserstein term acts as a regulariser that discourages abrupt changes in the policy update. Conceptually, the picture is closer to moving a cloud of policy particles smoothly toward higher return than to taking an unconstrained gradient step in parameter space.
Earlier, \citet{richemond2017wasserstein} linked Wasserstein gradient-flow perspectives to the Fokker--Planck equation and Langevin-type dynamics in policy space, providing a continuous-time interpretation of policy updates as diffusion processes in Wasserstein geometry; however, their analysis is restricted to the bandit (single-step) setting under strong regularity assumptions.

For parametric policies $\pi_{\theta}$, the influence of Wasserstein geometry can be expressed in terms of a matrix that plays the role of a metric tensor on parameter space. Drawing on the general theory of Wasserstein information geometry \citep{li2018natural}, one can define a Riemannian metric on the parameter manifold by pulling back the $\Wtwo$ metric defined on the space of measures. In this framework, \citet{moskovitz2021efficient} introduce a \emph{Wasserstein information matrix} $G(\theta)$ for policies and define the Wasserstein natural gradient as
\[
g^{W} = G(\theta)^{-1}\,\nabla_{\theta}\mathcal{L}(\theta),
\]
where $\mathcal{L}(\theta)$ is a policy objective (for example, the surrogate in \eqref{eq:policy-surrogate}) and $\nabla_{\theta}\mathcal{L}(\theta)$ is its Euclidean gradient. Informally, $G(\theta)$ captures how small changes in parameters $\theta$ affect the $\Wtwo$ distance between nearby policy-induced measures; the direction $g^{W}$ is the steepest-ascent direction when distances between policies are measured in Wasserstein space rather than in parameter space.
This is the Wasserstein analogue of what the Fisher information matrix does for classical natural gradients: it rescales parameter directions according to how strongly they deform the policy distribution.

Computing $G(\theta)$ exactly is infeasible except in simple cases. \citet{moskovitz2021efficient} therefore propose low-rank and kernel-based approximations, leading to practical algorithms for Wasserstein natural gradients in both policy-gradient (WNPG) and evolution-strategy (WNES) settings. They report improvements in computational cost and performance, and highlight an advantage over Fisher natural gradients in settings where the optimal solution is deterministic. This connects back to the trust-region view: including a Wasserstein penalty in the objective, as in \eqref{eq:ot-trust-region}, can be seen as inducing a local metric, and Wasserstein natural gradients provide an explicit approximation to the corresponding Riemannian geometry.

The proximal formulation \eqref{eq:jko} also clarifies how Wasserstein gradient flows relate to the OT-based trust-region methods of Section~\ref{sec:policy-trust}. A JKO step is a proximal update in Wasserstein space. Methods such as WPO/SPO \citep{song2023provably} and Wasserstein policy optimisation \citep{pfau2025wasserstein} can therefore be read as discrete-time approximations to policy optimisation in that geometry. For WPO/SPO, the link is direct, because the algorithms use Wasserstein or Sinkhorn trust regions explicitly. For \citet{pfau2025wasserstein}, the link is looser: the derivation is Wasserstein, but the implemented algorithm enforces its trust region with KL regularisation. The broader lesson is that Wasserstein trust regions, behavioural regularisers, and Wasserstein natural gradients are not separate ideas so much as different ways of choosing a geometry on policy space and then asking updates to stay small in that geometry while still improving return. The same gradient-flow viewpoint reappears in the offline setting in Section~\ref{sec:offline}.

\subsection{Wasserstein-trained critics}
\label{sec:policy-critics}

So far, OT has acted on the policy side of actor-critic learning: it constrained the actor directly or supplied the geometry in which policy updates were defined. It also appears on the critic side. Since actor-critic methods plug critic-based advantage estimates into the surrogate objective \eqref{eq:policy-surrogate}, any Wasserstein loss used to train the critic indirectly changes the policy update as well.

Wasserstein distances appear in critic learning in two distinct ways. The first concerns returns in distributional RL, where the full return distribution $Z^\pi(s,a)$ is modelled rather than just its mean. The second concerns the value function, where critics model posterior distributions over Q-values updated via Wasserstein barycentres. 

\paragraph{Distributional RL in Wasserstein and Sinkhorn metrics.}
Instead of estimating the mean return for each state-action pair, distributional RL models the distribution of returns to capture the inherent randomness. Because return distributions are one-dimensional, Wasserstein distances can be computed through quantile comparisons, making OT a computationally efficient and geometrically meaningful loss for distributional critics. Distributional Bellman operators are contractive in suitable Wasserstein metrics under standard assumptions, motivating losses that measure the Wasserstein discrepancy between predicted and target return distributions. With the action-value distribution $Z^\pi$, the distributional Bellman operator is
\[\mathcal{T}^{\pi}Z^\pi(s,a)\overset{D}{=}R(s,a)+\gamma Z^\pi(s',a'), \quad s'\sim P(\cdot \mid s,a), \ a'\sim\pi(\cdot\mid s'),\]
and the distance between current and target Bellman return distributions is measured by
\[d_{\sup}(Z_{\theta_z},\mathcal T^\pi Z_{\bar{\theta}_z})=\sup_{s,a}\Wassp{p}(Z_{\theta_z}(s,a),(\mathcal T^\pi Z_{\bar{\theta}_z})(s,a)),\]
where $Z_{\bar{\theta}_z}$ is the target for stability in practice. \textcolor{black}{The deterministic-reward convention of Section~\ref{sec:rl-brief} is retained here: $Z^\pi$ is random through the transitions and the policy, so $R(s,a)$ may be a deterministic function without trivialising the operator. If rewards are themselves stochastic, let $(r,s')$ have conditional law $K(\cdot,\cdot\mid s,a)$ with $\E[r\mid s,a]=R(s,a)$ and $a'\sim\pi(\cdot\mid s')$; then $\mathcal{T}^\pi Z^\pi(s,a)$ has the law of $r+\gamma Z^\pi(s',a')$, and the deterministic convention is recovered when $r=R(s,a)$ almost surely.}
The practical point is that the critic is no longer asked only for the mean return, but for the whole distribution of plausible returns at each $(s,a)$.
In practice, \textcolor{black}{quantile-regression deep Q-network (QR-DQN)} \citep{dabney2018distributional} approximates the $\Wone$ loss via quantile regression on one-dimensional return distributions, providing a distributional alternative to scalar (mean-return) critics.
More recently, SinkhornDRL \citep{sun2024distributional} replaces the 1D quantile-based surrogate with a Sinkhorn divergence between particle-based distributional estimates, enabling multi-dimensional reward settings and improved numerical stability through entropic regularisation. \textcolor{black}{As discussed in Section~\ref{sec:ot}, debiasing removes the self-transport offset so that $\Seps{\varepsilon}(\mu,\mu)=0$ and the divergence remains geometry-aware; it does not make the resulting quantity a metric, and the contraction results reported for SinkhornDRL are tied to specific assumptions, costs and operators rather than following from a general metric property \citep{sun2024distributional}.} Both methods provide theoretical contraction guarantees and empirical improvements over non-distributional baselines. The relevant Bellman or projected Bellman operators have method-specific contraction or non-expansion guarantees, depending on the metric, projection, and representation. Distributional critics also enable OT-based off-policy evaluation, using Wasserstein distances between predicted and target return distributions \citep{qi2025distributional}.

Multi-dimensional return distributions are particularly relevant for safe RL with constraints, multi-objective RL, and exploration via learned joint reward distributions.

\paragraph{Wasserstein barycentres for Q-posteriors.}
Another way of using Wasserstein distances is to propagate uncertainty of Q-value estimates through Bellman updates instead of modelling return randomness.
\citet{metelli2019propagating} propose Wasserstein Q-learning (WQL), in which Q-values are represented as probability distributions (posteriors) rather than point estimates. For each state-action pair $(s,a)$, the critic maintains a distribution $Q(s,a)$ and updates it via a 2-Wasserstein barycentre between the previous estimate and a target distribution. More precisely, at time $t$ the new Q-posterior at $(s_t,a_t)$ is defined as the 2-Wasserstein barycentre
\[
Q_{t+1}(s_t,a_t)=\argmin_{Q}\Big\{(1-\alpha_t)\,\Wtwo^2\big(Q,Q_t(s_t,a_t)\big)+\alpha_t\,\Wtwo^2\big(\textcolor{black}{Q,\,r_{t}+\gamma V_t(s_{t+1})}\big)\Big\},
\]
where $\alpha_t$ is a learning rate, $V_t(s)$ is a value-posterior obtained as a 2-Wasserstein barycentre of $\{Q_t(s,a):a\in\mathcal{A}\}$, and $\textcolor{black}{r_{t}}+\gamma V_t(s_{t+1})$ denotes the distribution of the one-step reward plus the discounted next-state value. \textcolor{black}{We have re-indexed the reward as $r_t$, the reward of the transition \emph{starting} at time $t$, to match the convention $r_t=R(s_t,a_t)$ fixed in Section~\ref{sec:rl-brief}; \citet{metelli2019propagating} write $r_{t+1}$ under a next-step reward convention. Note also that $Q_t(s,a)$ denotes a \emph{posterior measure} over values here, not a scalar Q-value.} In Gaussian and particle-based approximations, this minimiser has a closed form, so that uncertainty is propagated across the Bellman update by 2-Wasserstein barycentres with modest computational overhead.
This update can be read as a distributional analogue of TD learning: instead of averaging two scalar targets, it takes a Wasserstein barycentre between the old Q-posterior and the new Bellman target distribution.

\citet{likmeta2023wasserstein} extend this idea to continuous-action actor-critic methods in Wasserstein Actor-Critic (WAC). They maintain Gaussian Q-posteriors and update them using $\Wtwo$ to target distributions, similarly to WQL. The actor is then updated according to an optimistic estimate of Q, derived from the upper tail of the Q-posteriors, so that Wasserstein-based uncertainty quantification guides exploration in action space. Recent work extends this to MMD barycentres as an alternative to Wasserstein for model-free uncertainty propagation \citep{roy2025wasserstein}.

Even without explicit distributional modelling, Wasserstein regularisation can improve the stability of standard critics. Wasserstein Adaptive Value Estimation for
Actor-Critic (WAVE) \citep{baheri2025wave} augments the usual TD loss with a small entropy-regularised Wasserstein term between empirical Q-value distributions at successive iterations:
\[
L(\theta)=L_{\mathrm{TD}}(\theta)+\lambda_k\,\Wassp{\varepsilon}\big(\hat{Q}_k,\hat{Q}_{k-1}\big),
\]
where $L_{\mathrm{TD}}$ is a standard TD error, $\hat{Q}_k$ and $\hat{Q}_{k-1}$ are empirical distributions of Q-values under the critic at iterations $k$ and $k-1$, and $\lambda_k$ is an adaptive regularisation weight based on recent performance. The transport term discourages abrupt changes in Q-values from one update to the next and leads to smoother critic and policy dynamics.

For policy optimisation, these methods modify not the policy update rule itself, but the critic that supplies the value estimates. Wasserstein distances govern how value distributions evolve with training, and hence indirectly regularise the gradients used for policy updates.

\subsection{Summary}
\label{sec:policy-discussion}

Across the methods in this section, OT acts on three kinds of policy-induced objects: action distributions, behavioural embeddings, and value or return distributions. The details differ, but the underlying move is the same. KL-style methods ask whether the new policy assigns similar probability to the same actions; OT asks how much probability mass must actually move, and through what geometry, to turn one distribution into another.

That shift leads to four main uses: trust regions on action distributions, regularisation of trajectory embeddings, gradient-flow or natural-gradient updates in Wasserstein geometry, and critic training based on return or Q-value distributions, \textcolor{black}{as summarised in Table~\ref{tab:summary4}}. OT in policy optimisation is therefore not a single algorithm but a family of ways to redefine what counts as a small, structured update. When the action geometry is weak or uninformative, KL/PPO-style updates are usually the simpler baseline.

\section{Model-based RL and Wasserstein world models}
\label{sec:modelbased}

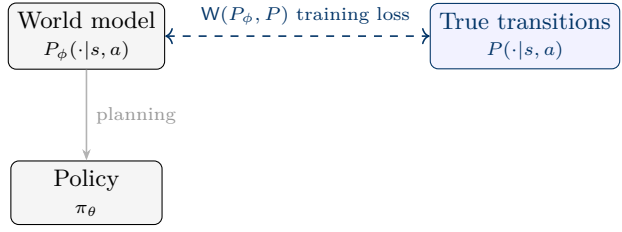
\begin{figure}[t]
\centering
\begin{tikzpicture}[
    node distance=1.2cm and 1.8cm,
    rlbox/.style={rectangle, draw, rounded corners=3pt, minimum width=2cm,
        minimum height=0.7cm, align=center, font=\small, fill=gray!8},
    otbox/.style={rectangle, draw, rounded corners=3pt, minimum width=2cm,
        minimum height=0.7cm, align=center, font=\small,
        draw=darkblue, text=darkblue, fill=blue!5},
    arr/.style={-{Stealth[length=1.5mm]}, semithick, gray!60},
    otarr/.style={<->, semithick, darkblue, dashed},
    otlbl/.style={font=\scriptsize, text=darkblue, align=center, fill=white, inner sep=2pt}
]

\definecolor{darkblue}{RGB}{0,51,102}

\node[rlbox] (model) {World model\\[-1pt]{\scriptsize $P_\phi(\cdot|s,a)$}};
\node[otbox, right=3.5cm of model] (true) {True transitions\\[-1pt]{\scriptsize $P(\cdot|s,a)$}};

\draw[otarr] (model) -- (true);
\node[otlbl, above=2pt] at ($(model)!0.5!(true)$) {$\Wass(P_\phi, P)$ training loss};

\node[rlbox, below=1.2cm of model] (policy) {Policy\\[-1pt]{\scriptsize $\pi_{\theta}$}};
\draw[arr] (model) -- (policy) node[midway, right, font=\scriptsize, text=gray!80] {planning};

\end{tikzpicture}
\caption{OT in model-based RL (Section~\ref{sec:modelbased}). A learned world model $P_\phi(\cdot\mid s,a)$ is trained to match the true transitions under a Wasserstein loss that upper-bounds worst-case value error. Bisimulation metrics embed this comparison into a fixed-point state metric (Section~\ref{sec:bisimulation}).}
\label{fig:modelbased}
\end{figure}

Model-based RL differs from model-free RL in that it relies on a learned, or nominal, transition model $P(\cdot\mid s,a)$ for planning, value expansion, or policy optimisation. This can greatly improve sample efficiency, but it introduces a new failure mode: if the learned world model is wrong, multi-step rollouts and Bellman backups can amplify small local errors into large control errors. OT is useful here because it measures model error in terms of how next-state distributions or trajectory distributions must move, under a chosen state-space geometry, rather than only through pointwise likelihood or regression losses (Figure~\ref{fig:modelbased}). The guiding question in this section is therefore not just ``is the model accurate?'', but ``is it accurate in the ways that matter for control?''
\textcolor{black}{Consider the robotic pick-and-place task again.} \textcolor{black}{The} \textcolor{black}{learned transition model predicts how the block and robot will move after executing a given action. Suppose two learned transition models predict the block position with errors of the same magnitude. A small error within the tolerance still allows a two-fingered gripper to successfully grasp the block, whereas an error of the same magnitude in the perpendicular direction may shift the block outside the feasible grasping region and cause the grasp to fail.} \textcolor{black}{These two errors are \emph{indistinguishable} under an isotropic Euclidean ground cost, which charges them equally: OT does not discover the grasp tolerance by itself. What separates them is the cost we choose. Encoding the task geometry, for example through an anisotropic quadratic cost $c(s,s')=(s-s')^{\top}M(s-s')$ that weights the grasp-sensitive direction more heavily, or through a reachability- or value-aware metric, makes the transport discrepancy penalise the consequential error more than the harmless one. The geometry must be \emph{encoded} in the ground cost rather than assumed to emerge from the use of OT, with $M$ positive semidefinite (positive definite if it is to induce a metric).}

This section focuses on OT as a tool for \emph{world modelling}. We first look at value-aware losses between true and learned next-state distributions $P(\cdot\mid s,a)$ and $P_\phi(\cdot\mid s,a)$. We then turn to latent-model objectives, where OT is used to learn compressed MDPs that preserve control-relevant structure, often with bisimulation-style guarantees. Robust control via Wasserstein ambiguity sets is closely related, but conceptually different: there OT hedges against transition uncertainty rather than fitting a model to data, and we discuss it in Section~\ref{sec:robust}.

\subsection{Model-based RL in brief}
\label{sec:modelbased-setup}

\textcolor{black}{Learning a transition model from observed trajectories is a form of system identification; planning then uses the learned model to evaluate or select actions, while RL optimises behaviour for long-term return.} A standard model-based pipeline learns $P_\phi(\cdot\mid s,a)$ from data and then uses the model for planning or policy optimisation, for example through model predictive control, Monte Carlo tree search, Dyna-style imagined rollouts, or value expansion. The basic challenge is \emph{compounding error}: one-step prediction may look accurate on average, yet the induced multi-step rollouts may still distort value estimates enough that the policy learns to exploit model flaws. A useful intuition is that small local errors can become strategically important when planning repeatedly feeds the model's own predictions back into itself. OT-based objectives address this by replacing purely pointwise prediction losses with discrepancies that reflect both state-space geometry and the control relevance of prediction errors.

\subsection{Value-aware model learning and Wasserstein losses}
\label{sec:modelbased-vaml}

In model-based RL, one learns an approximate transition model $P_\phi(\cdot\mid s,a)$ (and possibly a reward model $R_\phi$) from data and uses it for planning or to generate fictitious rollouts. 
A standard approach is maximum likelihood estimation (MLE) over a parametric model class \textcolor{black}{$\{P_\phi:\phi\in\Theta\}$ whose members admit densities $p_\phi(\cdot\mid s,a)$ with respect to a common reference measure on $\mathcal{S}$ (Lebesgue measure in continuous spaces, counting measure in discrete ones). Given observed next states $s'_1,\dots,s'_n$ from \textcolor{black}{$P(\cdot\mid s,a)$}, MLE minimises the average negative log-likelihood,}
\[
\textcolor{black}{\hat{\phi}\in\argmin_{\phi\in\Theta}\;-\frac{1}{n}\sum_{i=1}^{n}\log p_\phi(s'_i\mid s,a).}
\]
\textcolor{black}{This objective does not explicitly weight errors by their consequences for control. Likelihood losses can and do penalise different errors by different amounts; what they lack is any dependence on how much a given error changes the decisions or values downstream.} It can therefore prioritise discrepancies that matter little to the resulting decisions, while failing to prioritise errors that change which action is best.

\citet{farahmand2017value} argue that model-learning objectives should instead be \emph{value-aware}: not all aspects of the environment are helpful to find an optimal policy, and the decision problem is reflected by the reward function. Errors that significantly change the value function should be penalised more than errors that are irrelevant for control. In plain terms, the model should care most about getting the consequences of decisions right. One way to formalise this is to measure the worst-case discrepancy in value across a class of test functions. For a given state-action pair $(s,a)$, define
\[
\mathcal{E}(s,a)
\;=\;
\sup_{V\in\mathcal{V}}\bigg|\E_{s'\sim P(\cdot\mid s,a)}[V(s')]-\E_{\hat s'\sim P_\phi(\cdot\mid s,a)}[V(\hat s')]\bigg|^2,
\]
where $\mathcal{V}$ is a class of value functions. The value-aware loss then aggregates $\mathcal{E}(s,a)$ over $(s,a)$ drawn from the data distribution; intuitively, this loss is small if the learned model and the true dynamics produce similar expectations for all test functions in $\mathcal{V}$.
The role of the supremum is important: it asks for the \emph{worst} value function in the class, so the model is penalised according to the largest control-relevant mistake it could induce at $(s,a)$.

A natural choice for $\mathcal{V}$ is the class of Lipschitz functions with respect to a chosen state metric $d$.
By the Kantorovich-Rubinstein duality (Section~\ref{sec:ot}), the supremum over $1$-Lipschitz functions recovers the $\Wone$ distance, so that
\[
\mathcal{E}(s,a)
\;=\;
\Wone^{2}\!\big(P(\cdot\mid s,a),\;P_\phi(\cdot\mid s,a)\big).
\]
\citet{asadi2018equivalence} formalise this connection: when $\mathcal{V}$ consists of $C$-Lipschitz functions, the supremum equals $C^2\,\Wone^2$, so minimising the value-aware objective $\mathcal{E}(s,a)$ is equivalent, up to the constant $C^2$, to minimising the squared $\Wone$ distance between the true and learned next-state distributions. \textcolor{black}{Two qualifications matter. First, the constant $C$ is available only under a Lipschitz-MDP assumption (Lipschitz rewards and transitions with a contraction condition relating the Lipschitz modulus to $\gamma$); a bound on reward \emph{magnitude} alone does not establish the Lipschitz regularity of $V$ that the argument needs. Second, if $\mathcal{V}$ is a strictly smaller class than the full $C$-Lipschitz ball, 
the identity becomes an upper bound: $\mathcal{E}(s,a)\le C^2\Wone^2$, and minimising the transport term controls the value-aware loss without being equivalent to it.}
This provides a direct link between model-based RL and OT: training $P_\phi$ to minimise a Wasserstein distance with a value-relevant state metric is equivalent to training it to minimise the worst-case squared value error over a Lipschitz class of value functions. The key intuition is that the state metric weights model errors by their impact on downstream value.
This is the point at which OT becomes more than a generic discrepancy: once the metric reflects control similarity, Wasserstein model fitting becomes a way of fitting the model for planning rather than merely for prediction.

\subsection{Bisimulation metrics and Markov-structured transport}
\label{sec:bisimulation}

Bisimulation metrics formalise when two states are ``similar for control'': they compare both immediate rewards and the similarity of next-state distributions under each action. A classic definition is the (unique) fixed-point of a contraction mapping \citep{ferns2004metrics}. For a candidate state metric $d$, \textcolor{black}{with the transport term taken with respect to \emph{that same} metric, one form is}
\[
d(s,s')=\max_{a\in\mathcal{A}}\Big((1-\gamma)\,\big|R(s,a)-R(s',a)\big|+\gamma\,\textcolor{black}{\Wone[d]}\!\big(P(\cdot\mid s,a),\,P(\cdot\mid s',a)\big)\Big),
\]
with entropy-regularised variants used in practice for scalability \citep{kemertas2022approximate}. \textcolor{black}{We write $\Wone[d]$ to make explicit that the ground metric inside the operator is the candidate metric $d$
      itself. This is not a notational convenience but part of the definition: substituting an unrelated fixed
      ground metric yields a different operator, whose fixed point is not the bisimulation metric. With the $(1-\gamma)$ normalisation used in the display above, the resulting metric bounds optimal-value differences only after rescaling,}
\[
\textcolor{black}{\big|V^\ast(s)-V^\ast(s')\big|\;\le\;\frac{d(s,s')}{1-\gamma},}
\]
\textcolor{black}{and not by $d(s,s')$ itself; unnormalised variants of the fixed point absorb the factor into $d$ instead. Either way, such metrics motivate representation-learning objectives that collapse bisimilar states.}
One should read this fixed-point equation as balancing two questions at once: do the states give similar immediate reward, and do they lead to similar future-state distributions under every action?

Recent work connects these ideas to OT between \emph{stochastic processes} and shows how to incorporate Markov structure directly into the coupling class \citep{moulos2021bicausal,calo2024bisimulation}. Instead of arbitrary couplings over full trajectories, one restricts to \emph{Markovian couplings} between two Markov chains with transition kernels $\mu(\cdot\mid x)$ and $\nu(\cdot\mid y)$. Concretely, a Markovian coupling is a kernel $\kappa(\cdot\mid x,y)$ on $\mathcal{X}\times\mathcal{Y}$ such that its marginals agree with the original transitions:
\[
\sum_{y'} \kappa(x',y'\mid x,y)=\mu(x'\mid x),
\qquad
\sum_{x'} \kappa(x',y'\mid x,y)=\nu(y'\mid y).
\]
Unlike standard couplings, Markovian couplings ensure that the pairing at time $t+1$ depends only on the paired states at time $t$, respecting the sequential structure of MDPs.
This is exactly the extra structure missing from ordinary trajectory couplings: the transport itself has to respect the Markov evolution of the process.
Writing $\Pi_{x,y}$ for the set of such admissible couplings at the pair $(x,y)$, discounted OT costs can be characterised by Bellman-style equations of the form
\[
V^\star(x,y)=c(x,y)+\gamma\inf_{\kappa\in\Pi_{x,y}}\ \sum_{x',y'} \kappa(x',y'\mid x,y)\,V^\star(x',y').
\]
When the one-step cost $c$ is chosen to reflect reward mismatch, these dynamic-programming characterisations align closely with bisimulation fixed points and suggest practical solvers based on value iteration, including entropy-regularised variants with Sinkhorn-style updates \citep{calo2024bisimulation}.
The Bellman form is not accidental: it says that transport between processes should itself respect the recursive structure of control problems, not just match isolated snapshots of state.

Algorithmically, this viewpoint provides OT computations that respect temporal/Markov structure and clarifies why OT-like objectives in latent world models can support bisimulation-style guarantees: the transport discrepancy is computed over transitions in a way that is consistent with the dynamics, rather than over unordered state sets.

\subsection{Wasserstein auto-encoded and OT-based latent world models}
\label{sec:modelbased-latent}

The bisimulation perspective motivates a more ambitious goal: learning an entire latent MDP whose dynamics are trained via explicit OT objectives, a departure from reconstruction/likelihood-based latent world models such as \citep{corneil2018efficient,franccois2019combined,hafner2021mastering}. Wasserstein auto-encoded MDPs (WAE-MDPs) \citep{delgrange2023wasserstein} combine the framework of Wasserstein autoencoders with MDPs. They introduce an encoder $\phi$ that maps high-dimensional states to latent states, a decoder $\psi$ that reconstructs states from latent states, and a latent transition model $\bar P$. The key idea is to train this latent MDP by minimising a penalised optimal-transport distance 
between \emph{trace} distributions: the distribution of trajectories generated by the original policy in the real environment, and the distribution of trajectories generated by the distilled policy in the latent MDP and then decoded back to the original space. The intuition is that the latent model should not merely reconstruct observations; it should reproduce the behaviourally relevant trajectories of the original \textcolor{black}{environment}.

Concretely, WAE-MDPs augment the Variational MDPs (VAE-MDPs) objective with OT-based terms: a global Wasserstein penalty on trace (trajectory) distributions $L_{\xi_{\bar{\pi}}}$, together with local losses that encourage the latent reward $L_R$ and transition functions to approximate the original dynamics $L_P$. As part of the overall loss function, the local transition loss is defined as
\[L_P(\phi,\bar{P})=\E_{(s,a)\sim \xi_{\bar{\pi}}}\Wone\big(\phi_\#P(\cdot\mid s,a),\;\bar{P}(\cdot\mid\phi(s),a)\big),\]
where $\xi_{\bar{\pi}}$ is the stationary distribution of the policy $\bar{\pi}(\cdot|\phi(s))$. 
This term compares the real next-state distribution after encoding with the latent model's predicted next-state distribution. If it is small, then stepping once in latent space means roughly the same thing as stepping once in the real environment and then encoding.
The resulting latent model is shown to satisfy bisimulation-like properties: states that are functionally equivalent for control are mapped close together in latent space. This, in turn, enables formal verification and safety analysis to be performed in the latent MDP, with guarantees that transfer back to the original environment.
The high-level picture is that the latent state is not just a compressed observation; it is meant to be a compressed control state. A closely related extension, WAE-PCN \citep{delgrange2023waepcn}, combines Wasserstein autoencoded latent models with Pareto-conditioned networks for multi-objective RL.

Optimal Transport MDPs (OT-MDPs) \citep{ropke2025integrating} generalise and simplify this idea. They define a generic regularised OT objective between trajectories in the original MDP and trajectories in a learned latent MDP to learn the encoder, latent transition model and reward model:
\textcolor{black}{minimising over $\phi,\bar P,\bar R$, a regularised transport cost between the trace distribution induced by $\bar\pi$ in the original MDP and its counterpart in the latent MDP, together with the local penalties $L_{\xi_{\bar\pi}}+L_P+L_R$ weighted by $\lambda$. We describe the objective in words rather than displaying it, because the transport term is an OT cost between two \emph{trace distributions} and therefore involves a coupling of real and latent traces; writing it as an expectation over a single sampled trajectory would leave the law of the latent trace, and its coupling to the real one, unspecified.} The cost $c$ can be chosen with more freedom than the fixed reconstruction cost $c_{\text{WAE}}(\tau,\bar{\tau})=d_S(s,\psi(\bar{s}))+d_S(s',\psi(\bar{s}'))$ used by WAE-MDPs. OT-MDPs define a general cost of the form
\[c(\tau,\bar{\tau})=h\big(f(s),g(\bar{s})\big),\]
where $f$ and $g$ are feature maps on real and latent states and $h$ is a distance or divergence in feature space. In a specific instantiation, $c_{\text{value}}(\tau,\bar{\tau})=|V(s)-\bar{V}(\bar{s})|$ uses value functions directly, yielding a cost that prioritises control-relevant similarity.
The extra flexibility matters because it lets the modeller decide what two trajectories should count as ``the same'' for control purposes: visual similarity, feature similarity, or similarity in downstream value.
The OT-MDP framework then gives conditions on $c$ under which the learned latent model is bisimilar to the original MDP, and shows how planning tools such as model-based value expansion and Monte Carlo tree search can be run in the latent space. The OT objective is not just a training loss here; it is the quantity that controls how faithfully values and planning decisions transfer from the latent model back to the original environment. In practice, WAE-MDPs enforce 1-Lipschitzness with a gradient penalty, whereas OT-MDPs discretise the relaxed latent state \textcolor{black}{and instead encourage $1/2$-boundedness through another penalty, corresponding to the function-class constraint in the dual representation of total variation distance. Here $1/2$-boundedness means the magnitude of the function is bounded by $1/2$, i.e., $\|f\|_\infty\leq 1/2$.}

For control, WAE-MDPs and OT-MDPs use optimal transport to train world models that preserve the structure of the original MDP in a way that is relevant for value and safety, rather than merely matching raw observations. Policies are then optimised in these latent models using standard or OT-based policy-optimisation methods. The practical takeaway is that OT is being used as a filter on what kind of model fidelity matters: not pixel-level fidelity, but control-relevant fidelity. The closer the latent model is to the true environment in that Wasserstein sense, the more reliable the resulting policy improvements are.

\subsection{Summary}
\label{sec:modelbased-discussion}
OT enters model-based RL in three main ways: as a value-aware discrepancy between learned and true next-state laws, as the metric inside bisimulation-style state comparisons, and as an explicit trajectory-matching objective in latent world models. \textcolor{black}{The methods covered in this section are summarised in Table~\ref{tab:summary5}.}

Across these methods, the central modelling choice is the \emph{metric structure} used to compare next-state distributions or trajectory traces. In value-aware model learning, this choice appears as the state metric that defines $1$-Lipschitz test functions (hence the implied $\Wone$ geometry). In latent-model objectives, it appears as the trajectory feature maps $f,g$ and the cost $c$ used in the OT alignment. In high-dimensional observation spaces, these costs are typically defined in learned feature spaces rather than raw pixel space.

Ambiguity-set methods in Section~\ref{sec:robust} use the same geometry for a different purpose: hedging against shifted transition laws rather than fitting a model to data. The same modelling choices also reappear in offline settings, where learned dynamics are especially exposed to distributional shift and compounding error (Section~\ref{sec:offline}).

Several extensions are suggested by the OT tools reviewed in this survey. First, entropic OT/Sinkhorn discrepancies provide scalable minibatch objectives for matching conditional next-state laws $P(\cdot\mid s,a)$ via samples, and can be combined with learned feature costs to handle high-dimensional observations. Second, partial and unbalanced OT are well suited when transition data exhibit missing outcomes, outliers, or systematic mass mismatch (for example, logged data that omit rare but important transitions), and may yield more robust world models than strict balanced matching.

A newer direction makes the OT-flow connection operational for \emph{shifted dynamics}. Composite Flow Matching for RL with Shifted Dynamics Data (CompFlow) uses flow-matching constructions to model transition dynamics under distribution shift and estimates a \emph{dynamics gap} through Wasserstein distances between learned flow models. That gap is then used to guide data collection and exploration toward regions where source-domain data are least informative for the target dynamics \citep{kong2025compflow}.

\section{Offline policy optimisation and regularisation}
\label{sec:offline}

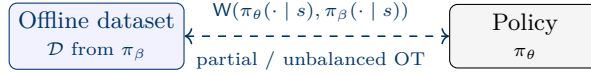
\begin{figure}[t]
\centering
\begin{tikzpicture}[
    node distance=1.2cm and 1.8cm,
    rlbox/.style={rectangle, draw, rounded corners=3pt, minimum width=2cm,
        minimum height=0.7cm, align=center, font=\small, fill=gray!8},
    otbox/.style={rectangle, draw, rounded corners=3pt, minimum width=2cm,
        minimum height=0.7cm, align=center, font=\small,
        draw=darkblue, text=darkblue, fill=blue!5},
    arr/.style={-{Stealth[length=1.5mm]}, semithick, gray!60},
    otarr/.style={<->, semithick, darkblue, dashed},
    otlbl/.style={font=\scriptsize, text=darkblue, align=center}
]

\definecolor{darkblue}{RGB}{0,51,102}

\node[otbox] (dataset) {Offline dataset\\[-1pt]{\scriptsize $\mathcal{D}$ from $\pi_\beta$}};
\node[rlbox, right=3.5cm of dataset] (policy) {Policy\\[-1pt]{\scriptsize $\pi_\theta$}};

\draw[otarr] (dataset) -- (policy);

\node[otlbl, above=2pt] at ($(policy)!0.5!(dataset)$) 
{$\Wass(\pi_\theta(\cdot\mid s),\pi_\beta(\cdot\mid s))$};
\node[otlbl, below=2pt] at ($(policy)!0.5!(dataset)$) {partial / unbalanced OT};

\end{tikzpicture}
\caption{OT in offline RL (Section~\ref{sec:offline}). The agent learns entirely from a fixed dataset $\mathcal{D}$ collected from the behaviour policy $\pi_\beta$. OT enters as a behaviour constraint $\Wass(\pi_\theta(\cdot\mid s),\pi_\beta(\cdot\mid s))$ that keeps the learned policy geometrically close to the data-generating policy with both induced distributions approximated by collected trajectories (Section~\ref{sec:offline-ot}). Partial and unbalanced OT relax mass conservation to handle suboptimal demonstrations.}
\label{fig:offline}
\end{figure}

Offline RL restricts the agent to a fixed dataset of transitions, with no environment interaction during learning (Figure~\ref{fig:offline}). The central difficulty is distributional shift: once the learned policy chooses actions that are poorly represented in the data, the critic must extrapolate into unsupported regions, and its value estimates can become badly biased. Standard remedies keep the learned policy close to the data-generating behaviour $\pi_{\beta}$, but 
density-ratio penalties such as $\KL(\pi\|\pi_{\beta})$ are a poor fit when supports differ, which could be too conservative and prevent further improvement beyond the behaviour policy, and lack geometric awareness. An OT cost $\Wass(\pi(\cdot\mid s),\pi_{\beta}(\cdot\mid s))$, by contrast, remains informative as long as a ground cost $c(a,a')$ is defined on $\mathcal{A}\times\mathcal{A}$ and encourages generalisation.
It therefore gives a geometric notion of staying close to the data: shifting mass between similar actions is cheap, while shifting it to qualitatively different actions is expensive. The intuition is the same as in policy trust regions, but the reason is different: here the goal is not mainly to stabilise optimisation, 
but to bias policy updates towards actions that remain geometrically close to what is in the dataset.

\textcolor{black}{Considering the robotic pick-and-place example, given an offline dataset containing trajectories generated by both expert and suboptimal policies, the learned policy should \textcolor{black}{improve upon the suboptimal behaviours so as to maximise return without deviating excessively} from the data distribution. During offline training, the learned policy may propose an action that is not present in the dataset. Suppose that, given the same state, actions of moving the robotic arm 4cm and 5cm forward are contained in the dataset. Then an action moving 4.5cm in the same direction could be regarded as an interpolation between observed actions, whereas an action moving 1 cm in the opposite direction lies much farther from the data-supported actions. OT captures this distinction through assigning lower transport cost to the former and higher cost to the latter according to the geometry of the action space.} \textcolor{black}{Transport cost measures proximity to the data, not quality: a small cost does not certify that an action is safe or high-value, and can still correspond to an unsupported action when the behaviour support is narrow. The geometry discourages deviation from the data; the value estimate decides whether a nearby action is preferable.}

\subsection{Standard offline RL framework}
\label{sec:offline-setup}

In offline RL \citep{lange2012batch,levine2020offline}, the agent learns from a fixed dataset $\mathcal{D} = \{(s_t, a_t, r_t, s_{t+1})\}$ generated by an unknown behaviour policy $\pi_{\beta}$ and cannot interact with the environment. Unlike imitation learning (Section~\ref{sec:il}), a reward function is available, and the objective is to \emph{improve} upon $\pi_{\beta}$, not merely replicate it. As $\pi$ deviates from $\pi_{\beta}$, the agent increasingly queries Q-values in regions of $\mathcal{S}\times\mathcal{A}$ that are poorly covered by $\mathcal{D}$, leading to overestimation, compounding bootstrapping errors, and eventual policy collapse. A good informal picture is that the critic only knows the terrain where data have been collected; once the actor leaves that terrain, its value estimates become guesses.

Two broad ideas dominate offline RL. The first is to \emph{constrain the policy}: keep the learned policy close enough to $\pi_{\beta}$ that the actor does not query the critic far outside the support of the data \citep{fujimoto2019BCQ,kumar2019bear,fujimoto2021td3bc,siegel2020keep,zhou2021plas,brandfonbrener2021OneStep,hepburn2025state}. Because $\pi_{\beta}$ is unknown, it must first be estimated, usually by supervised learning. A common formulation is
\begin{equation}
\label{eq:offline-constraint}
    \textcolor{black}{\pi^* \in \argmax_{\pi} \E_{s\sim \mathcal{D}}\left[\E_{a\sim\pi(\cdot\mid s)}\big[Q(s,a)\big] -\lambda\,D\big(\pi(\cdot\mid s),\, \pi_{\beta}(\cdot\mid s)\big)\right],}
\end{equation}
where $D$ denotes a divergence or regulariser, for example KL or MMD. \textcolor{black}{Here $Q$ is the fixed critic, the inner expectation evaluates actions sampled from the candidate policy, and $\lambda\ge 0$ controls deviation from the behaviour policy; we use $\lambda$ for this role throughout the section, reserving $\lambda_{\mathcal{P}}$ for the pessimism weight below.} OT provides a natural alternative for $D$: measuring deviation via $\Wass(\pi(\cdot\mid s),\pi_{\beta}(\cdot\mid s))$ makes ``small'' deviations depend on the geometry of $\mathcal{A}$ and remains meaningful under support mismatch. In that sense, offline OT regularisation is the data-constrained counterpart of the Wasserstein trust-region idea from Section~\ref{sec:policy-trust}. The difference is that the reference policy is now the dataset's behaviour policy rather than the previous policy iterate.

The second idea is \emph{pessimistic value estimation}: rather than constraining the actor directly, modify the Q-function so that out-of-distribution actions receive conservative values \citep{kumar2020cql,an2021uncertainty,kostrikov2022IQL}. The intuition is to make the critic suspicious of actions it has little evidence about. A general critic objective is
\begin{equation}
    \mathcal{L}(\theta_q)=\mathrm{TDLoss}(Q_{\theta_q}) + \textcolor{black}{\lambda_{\mathcal{P}}} \,\mathcal{P} \left(Q_{\theta_q}, \mathcal{D}\right),
\end{equation}
where $\mathcal{P}$ enforces pessimism outside the support of $\mathcal{D}$ \textcolor{black}{and $\lambda_{\mathcal{P}}\ge 0$ weights it}. OT-based policy constraints already induce a form of implicit pessimism, because they stop the actor from moving too aggressively into unsupported regions; OT-based ambiguity sets over transitions (Section~\ref{sec:robust}) can then make the critic targets themselves robust. Model-free and model-based versions both exist \citep{yu2020mopo,kidambi2020morel,yu2021combo,janner2022planning,rigter2022rambo,hepburn2024mbts_ml,wang2021offline,zhang2021autoregressive,diehl2021umbrella}, as do methods that combine explicit constraint with pessimistic estimation \citep{wu2019behavior,kostrikov2021offline,dadashi2021offline,beeson2024balancing}. The rest of this section focuses on OT-based behaviour constraints and on partial or unbalanced formulations that relax exact matching.

\subsection{OT-based policy constraints, partiality, and unbalanced formulations}
\label{sec:offline-ot}

The most common offline use of OT is as a \emph{behaviour regulariser}: the learned policy is discouraged from selecting actions that are far, in an action-geometry sense, from those supported by the dataset. OTR-style reward labelling (Section~\ref{sec:il-otr}) provides a complementary offline application, assigning pseudo-rewards via couplings rather than constraining the actor. In this section, the recurring question is therefore: should OT be used to keep the actor near the data, or to decide which parts of the data should matter most?

\paragraph{BRAC: a unifying framework.}
\citet{wu2019behavior} introduce \emph{behaviour-regularised actor critic} (BRAC), a generic actor-critic template that augments either the actor objective, the critic target, or both, with a divergence $D\big(\pi(\cdot\mid s),\pi_{\beta}(\cdot\mid s)\big)$. 
Writing $Q_{\theta_q}$ for the critic and $\pi_{\theta}$ for the actor, BRAC supports two modes: a \emph{value penalty} mode, which subtracts $\textcolor{black}{\lambda}\,D(\pi_{\theta}(\cdot\mid s'),\pi_{\beta}(\cdot\mid s'))$ inside the Bellman target, 
\[
\min_{\theta_q}\ 
\E_{(s,a,r,s')\sim\mathcal{D},\,a'\sim\pi_{\theta}(\cdot\mid s')}
\Big[\big(r+\gamma\big(Q_{\bar{\theta}_q}(s',a')-\textcolor{black}{\lambda}\,D(\pi_{\theta}(\cdot\mid s'),\pi_{\beta}(\cdot\mid s'))\big)-Q_{\theta_q}(s,a)\big)^2\Big],
\]
where $Q_{\bar{\theta}_q}$ is the target critic updated slowly to improve the stability of learning, 
and a \emph{policy regularisation} mode, which adds the penalty only to the actor update,
\begin{equation}
\label{eq:policyreg}
\max_{\theta}\ 
\E_{s\sim\mathcal{D}}\Big[ \E_{a\sim\pi_{\theta}(\cdot\mid s)}\big[Q_{\theta_q}(s,a)\big]\;-\;\textcolor{black}{\lambda}\,D\big(\pi_{\theta}(\cdot\mid s),\pi_{\beta}(\cdot\mid s)\big)\Big].
\end{equation}
The divergence $D$ is treated as a modular design choice; BRAC evaluates KL, kernel MMD, and Wasserstein distance. Replacing KL-type penalties by OT discrepancies (e.g.\ $\Wone$, $\Wtwo$, or Sinkhorn-based surrogates) makes the constraint \emph{geometric}: moving probability mass between similar actions is cheap, while moving mass between dissimilar actions is expensive. This is exactly the kind of notion of closeness one wants when the action space has structure that likelihood ratios alone cannot express.
The Wasserstein variant (BRAC-W) instantiates the Kantorovich-Rubinstein dual~\eqref{eq:kr-dual} on per-state action distributions, $\Wass(\pi(\cdot\mid s),\pi_{\beta}(\cdot\mid s)) = \sup_{\|g\|_{\mathrm{Lip}}\le 1} \E_{a\sim\pi}[g(a)] - \E_{a\sim\pi_{\beta}}[g(a)]$, with a gradient penalty enforcing the Lipschitz constraint \citep{gulrajani2017improved}. A central empirical finding is that simple behaviour regularisation is the essential ingredient for stable offline learning; many design complexities in prior work (adaptive penalty weights, large Q-ensembles) prove unnecessary. For the reader, BRAC is a useful anchor point: it is the clean baseline ``stay near the data'' framework, and the later OT methods mostly differ in how they define ``near'' or how much of the data they insist on matching.

The next three strands all depart from BRAC-style behaviour regularisation in different ways. Q-learning regularized by Direct Optimal Transport modeling (Q-DOT) replaces the adversarial Wasserstein critic with an explicit $\Wtwo$ transport map. Gradient-flow approaches interpret the update itself through Wasserstein geometry rather than as a penalty around a baseline policy. Partial and unbalanced OT methods relax exact mass matching so that suboptimal or contaminated datasets need not be copied wholesale.

\paragraph{$\Wtwo$-map regularisation (discriminator-free).}
When $\Wtwo$ geometry is desired, Brenier’s theorem (Section~\ref{sec:ot}) implies that (under absolute continuity of the source and with the squared Euclidean cost) the optimal transport map is $T=\nabla\psi$ for a convex potential $\psi$, so that $\Wtwo^2(\mu,\nu)=\int\|x-\nabla\psi(x)\|_2^2\,\mathrm{d}\mu(x)$.
Previous offline-RL methods that use Wasserstein regularisation typically estimate the discrepancy through a dual critic or another adversarial component. Q-DOT \citep{omura2025offline} instead aims for $\Wtwo$ geometry without a discriminator by parameterising a convex potential with input-convex neural networks (ICNNs). The learned policy is represented as the pushforward $\pi=\nabla\psi_{\#}\pi_{\beta}$ of the behaviour policy through the gradient of that potential, and the regularised actor objective becomes
\[\E_{(s,a)\sim\mathcal{D}}[Q_{\phi}(s,\nabla_a\psi(s,a))-\textcolor{black}{\lambda} \|a-\nabla_a\psi(s,a)\|_2^2].\]
This gives a discriminator-free alternative to adversarial Wasserstein penalties. In addition, the method trains a separate Advantage-Weighted Regression (AWR) policy to propose actions that need not appear explicitly in the offline dataset.
Both BRAC-W and Q-DOT evaluate OT objectives over minibatches rather than full distributions. The empirical Wasserstein distance is biased in finite samples \citep{bellemare2017cramer}. \textcolor{black}{\citet{fatras2019learning} study the resulting \emph{minibatch} objective, the expectation of the
        per-minibatch (Sinkhorn) loss over random minibatches, and show that averaging per-minibatch gradients
        gives an unbiased estimator of that objective. The qualification matters: the minibatch objective is
        not the full-distribution transport objective, it can lose distance properties such as vanishing at
        $\mu=\nu$, and the result is therefore a well-behaved surrogate rather than a correction that makes
        finite-minibatch gradients unbiased for the population Wasserstein distance.}

\paragraph{Value gradient flow without \textcolor{black}{an explicit behaviour penalty}.}
Consider the MaxEnt RL objective $\E_{\pi}[Q(s,a)-\alpha \log(\pi(a\mid s))]$, whose unique maximiser is the Boltzmann policy $\pi^*_Q(a\mid s)\propto\exp(Q(s,a)/\alpha)$. Viewing the KL divergence $\KL(\mu\|\pi^*_Q)$ as a free-energy functional over per-state action distributions $\mu$, one can apply the JKO scheme (see \eqref{eq:jko} in Section~\ref{sec:policy-wgf}) to obtain discrete-time $\Wtwo$ gradient flows. Because the resulting proximal steps are intractable in closed form, the distribution at each iteration is approximated by $N$ particles $\{a_i^{(l)}\}_{i=1}^N$, updated via 
\[
a_i^{(l+1)}=a_i^{(l)}+\textcolor{black}{h}\,\frac{1}{N}\sum_{j=1}^N\!\Big[k(a_j^{(l)}, a_i^{(l)})\,\nabla_{a_j}\tfrac{Q(s,a_j^{(l)})}{\alpha}\;+\;\nabla_{a_j}k(a_j^{(l)}, a_i^{(l)})\Big],
\]
where $k(\cdot,\cdot)$ is a positive-definite kernel and \textcolor{black}{$h>0$ a step size} \citep{xu2026reinforcement}. The first term steers particles toward high-$Q$ actions; the second is a repulsive term that maintains diversity. The temperature $\alpha$ controls the exploration--exploitation balance. \textcolor{black}{Two clarifications. First, the displayed rule is a Stein variational gradient descent (SVGD)-style update:
        the exact JKO proximal step is replaced by a descent direction restricted to a reproducing-kernel Hilbert
        space (RKHS) determined by $k$, so the particle dynamics approximate, rather than realise, exact JKO
        minimisation or an exact $\Wtwo$ gradient flow. Second, the method is unregularised only in the sense that
        no explicit divergence penalty against $\pi_\beta$ is added: departure from the reference distribution is
        still controlled, through a finite transport budget on the flow. OT therefore enters as the
        \emph{geometry of the update} rather than as \textcolor{black}{an explicit} constraint, with no trajectory-level temporal structure.}

\paragraph{Partial OT for suboptimal demonstrations (PPL).}

Expert data can be expensive, noisy, or simply suboptimal. In that setting, forcing the learner to match all observed behaviour is undesirable. 
\textcolor{black}{Partial Policy Learning (PPL) \citep{asadulaev2024rethinking} uses a critic-based cost to favour useful actions in suboptimal offline data. Its OT motivation is the one-sided \emph{incomplete transport} formulation of \citet{gazdieva2023extremal}. For probability measures $\mu$ and $\nu$, this takes the form}
\[
\textcolor{black}{\inf_{\Gamma\geq0}\left\{\int c(x,y)\,\mathrm{d}\Gamma(x,y):
(P_{\mathcal X})_\#\Gamma=\mu,\quad (P_{\mathcal Y})_\#\Gamma\leq w\nu\right\},\qquad w\geq1}
\]
\textcolor{black}{where $P_{\mathcal{X}}(x,y) = x$ and $P_{\mathcal{Y}}(x,y) = y$ are the projections onto the marginals. This shares the fixed-mass and sub-marginal structure of partial OT in Section~\ref{sec:ot}, but retains the full source mass and allows unmatched mass only on the scaled target side. Both marginals are matched exactly when $w=1$, and increasing $w$ relaxes the target constraint. For map-induced plans $\Gamma=(\mathrm{id},T)_\#\mu$, the constraint becomes $T_\#\mu\leq w\nu$.}

\textcolor{black}{PPL adopts a deterministic actor $\pi_\theta:\mathcal S\to\mathcal A$ and the cost $-Q_k(s,a)$. A pushforward $(\pi_\theta)_\#d_{\mathcal D}$ is an aggregate action distribution under dataset states $s\sim d_{\mathcal D}$, not the conditional behaviour distribution $\pi_\beta(\cdot\mid s)$; these should not be directly equated or compared by measure domination. At the distributional level, the incomplete-transport motivation compares the policy-induced state-action measure $(\mathrm{id},\pi_\theta)_\# d_{\mathcal D}$ with the state-action distribution induced by the dataset $\mathcal D$.
Motivated by this formulation, PPL introduces the state-conditioned neural optimisation, whose core objective is}
\[
\textcolor{black}{\sup_{\omega:\,f_\omega\leq0}\ \inf_\theta
\left\{
\E_{s\sim d_{\mathcal D}}\!\left[-Q_k(s,\pi_\theta(s))-f_\omega(s,\pi_\theta(s))\right]
+w\,\E_{(s,a)\sim\mathcal D}\!\left[f_\omega(s,a)\right]
\right\}.}
\]
\textcolor{black}{Here $Q_k$ is held fixed during the actor update, and $f_\omega:\mathcal S\times\mathcal A\to\mathbb R$ is a learned potential. We use the nonpositive-potential convention of \citet{gazdieva2023extremal}, also used in PPL's released implementation. The neural objective encourages high-value actions while using dataset actions to train the potential, and it is not asserted to be an exact reformulation of a conditional domination constraint on stochastic policies. The parameter $w$ controls the target relaxation, and excessive relaxation can be detrimental when demonstrations are already near-optimal.}

\paragraph{Unbalanced OT for constrained policy optimisation.}
Partial OT relaxes matching by transporting only part of the mass. Unbalanced OT relaxes it differently: mass may be created or destroyed, but doing so incurs a penalty. This is useful when the right policy should not merely select a subset of the data, but should also be allowed to reweight it more flexibly. \citet{givchi2021distributionally} cast policy optimisation in exactly this way, as occupancy-measure matching with negative reward as the transport cost and Bregman penalties, KL in their implementation, on the mismatch between projected marginals. With baseline policy $\pi'$ representing either the previous iterate or an expert policy, the distributionally constrained objective is
\[\max_{\rho_\pi}\E_\pi[r]-\epsilon_1D_{\psi_1}(\rho_\pi\mathbf{1}\mid\rho_{\pi'}\mathbf{1})-\epsilon_2D_{\psi_2}(\rho_\pi^T\mathbf{1}\mid\rho_{\pi'}^T\mathbf{1}),\]
where $\rho_\pi\mathbf{1}$ represents the projected state marginal and $\rho_\pi^T\mathbf{1}$ represents the projected action marginal of the occupancy measure. Assuming that the set of discounted occupancy measures is a compact convex polytope and adding an entropy term, \citet{givchi2021distributionally} optimise the resulting problem with Dykstra-style Bregman projections \citep{dykstra1983algorithm}. Considering large or continuous state and action spaces, this is extended to off-policy optimisation with sample-based estimates and duality. While introduced in an online setting, the formulation is directly applicable to offline constraints when data coverage is partial or when strict matching to empirical marginals is undesirable. From a computational standpoint, \citet{fatras2021unbalanced} show that minibatch estimation interacts favourably with unbalanced OT: unlike balanced OT, where subsampling introduces bias and artefacts, unbalanced losses with minibatches yield significantly more robust gradient estimates, supporting the scalability of such formulations.

\subsection{Summary}
\label{sec:offline-discussion}

OT-based offline RL methods can be read as making the ``stay close to data'' principle \emph{geometric}. The distance between the learned policy and the behaviour policy is not only a matter of likelihood ratios (KL), but of how much probability mass must be transported in the action space under a chosen cost. Partial and unbalanced OT extend this idea to settings where the dataset contains suboptimal behaviour or outliers, relaxing mass conservation so that the learner matches only the best-supported portion of behaviour. \textcolor{black}{Table~\ref{tab:summary6} summarises technical characteristics of the methods discussed in this section.}

Several OT tools discussed elsewhere in this survey suggest extensions in offline RL. OT-based transition ambiguity sets (Section~\ref{sec:robust-ambiguity}) and OT-based world models (Section~\ref{sec:modelbased}) can be combined with offline learning to provide robust/pessimistic targets under model misspecification. OT-based data filtering and reweighting (e.g., through OTR-style scores in Section~\ref{sec:transfer}) provide a complementary mechanism for reducing distribution shift by curating the dataset before policy learning. Wasserstein/Sinkhorn penalties can also be applied to learned behavioural embeddings, yielding offline analogues of behaviour-guided regularisation (Section~\ref{sec:policy-bgpg}). Establishing systematic guidelines for choosing costs and partiality/unbalanced parameters from data, and providing theory comparable to KL-regularised offline frameworks, remain open directions.

\section{Robustness, safety, and transfer under distribution shift}
\label{sec:robust}

\begin{figure}[t]
\centering
\begin{tikzpicture}[
    node distance=1.2cm and 1.8cm,
    rlbox/.style={rectangle, draw, rounded corners=3pt, minimum width=2cm,
        minimum height=0.7cm, align=center, font=\small, fill=gray!8},
    otbox/.style={rectangle, draw, rounded corners=3pt, minimum width=2cm,
        minimum height=0.7cm, align=center, font=\small,
        draw=darkblue, text=darkblue, fill=blue!5},
    arr/.style={-{Stealth[length=1.5mm]}, semithick, gray!60},
    otarr/.style={<->, semithick, darkblue, dashed},
    otlbl/.style={font=\scriptsize, text=darkblue, align=center, fill=white, inner sep=2pt}
]

\definecolor{darkblue}{RGB}{0,51,102}

\node[rlbox] (transitions) {Transitions\\[-1pt]{\scriptsize $P_0(\cdot|s,a)$}};
\node[otbox, below=1.2cm of transitions] (ambiguity) {Ambiguity set\\[-1pt]{\scriptsize $\mathcal{U}_\delta(s,a)$}};
\draw[otarr] (transitions) -- (ambiguity);
\node[otlbl, right=2pt] at ($(transitions)!0.5!(ambiguity)$) {$\Wass(P, P_0) \le \delta$};

\node[rlbox, right=4.5cm of transitions] (policy) {Policy\\[-1pt]{\scriptsize $\pi_{\theta}$}};
\node[otbox, below=1.2cm of policy] (target) {Visitation /\\[-1pt]preference target};
\draw[otarr] (policy) -- (target);
\node[otlbl, right=2pt] at ($(policy)!0.5!(target)$) {$\Wass(\rho_\pi, \rho_{\mathrm{safe}})$};

\draw[arr] (transitions) -- (policy) node[midway, above, font=\scriptsize, text=gray!80] {robust Bellman};

\end{tikzpicture}
\caption{
OT in robustness, safety, and transfer (Section~\ref{sec:robust}). A Wasserstein ambiguity set $\mathcal{U}_\varepsilon$ around the nominal transitions defines worst-case Bellman updates (Section~\ref{sec:robust-ambiguity}). OT-based visitation regularisation and preference alignment shape where the policy visits or how it aligns with human feedback (Sections~\ref{sec:robust-visitation}--\ref{sec:robust-rlhf}).}
\label{fig:robustness}
\end{figure}
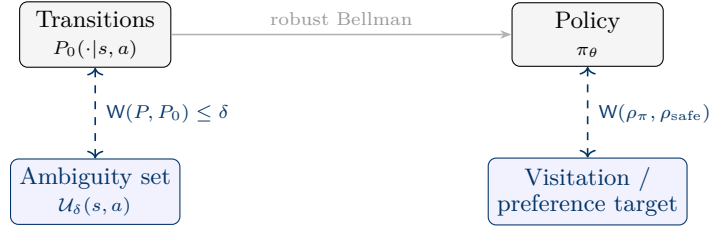

Robustness and safety become central once a learned policy is deployed outside the distribution on which it was trained (Figure~\ref{fig:robustness}). Risk criteria such as variance, entropic risk, or \textcolor{black}{conditional value at risk (CVaR)} summarise different aspects of return uncertainty\textcolor{black}{, with CVaR specifically tail-focused,} but they say little about \emph{how} uncertainty enters through the dynamics or \emph{which} perturbations should count as plausible. 
OT gives a geometric alternative: uncertainty is represented as bounded transport of probability mass under a chosen ground metric, and robustness is expressed through worst-case or stay-close objectives in that geometry. The central question is which environment shifts should be treated as plausible during training.

\textcolor{black}{For example, consider deploying a trained pick-and-place policy on another robotic arm with slightly different dynamics. The same action at the same state may induce a different distribution of next states from that of the training arm. OT can measure the discrepancy between transition distributions in a geometric sense to quantify admissible dynamics shifts for robust control. Similar ideas can be used for other types of distribution shift in different settings.}

This geometric viewpoint appears in several distinct places. It gives Wasserstein ambiguity sets for distributionally robust control, visitation-level penalties for safety shaping, uncertainty scores derived from transport between value-induced action distributions, DRO formulations for preference shift in RLHF, and alignment tools for transfer across tasks or domains. 
OT-based curriculum generation via distributional interpolation is discussed separately in Section~\ref{sec:curricula}.

\subsection{Distributionally robust control with Wasserstein ambiguity sets}
\label{sec:robust-ambiguity}

A standard distributionally robust RL perspective starts from a nominal transition law $P_0(\cdot\mid s,a)$, estimated from data or provided by a simulator, and then asks what happens if the true environment is nearby but not identical. Wasserstein ambiguity sets answer that question geometrically: for each $(s,a)$, nature may replace $P_0(\cdot\mid s,a)$ by any $P(\cdot\mid s,a)$ lying inside a Wasserstein ball of radius \textcolor{black}{$\delta$ under the chosen metric.}
In plain language, the agent is trained not against one model of the world, but against all models that can be reached from the nominal one by a bounded amount of transport.

\begin{figure}[t]
\centering
\begin{tikzpicture}

\definecolor{darkblue}{RGB}{0,51,102}

\draw[thick, rounded corners, fill=gray!5] (-3.8,-2.6) rectangle (3.8,2.6);
\node[font=\small, gray!70, anchor=north west] at (-3.7, 2.5) {$\mathcal{P}(\mathcal{S})$};

\fill[blue!6] (0,0) circle (1.9cm);
\draw[darkblue, thick, dashed] (0,0) circle (1.9cm);
\node[font=\small, darkblue] at (1.55, 1.55) {$\mathcal{U}_\delta$};

\draw[darkblue, thick, |<->|] (0,0) -- (1.34,1.34);
\node[font=\scriptsize, darkblue, fill=white, inner sep=1pt] at (0.8, 0.57) {$\delta$};

\filldraw[darkblue] (0,0) circle (3pt);
\node[font=\scriptsize, darkblue, right=4pt] at (0,0) {$P_0(\cdot\mid s,a)$};

\filldraw[red!80] (1.34,-1.34) circle (3pt);
\node[font=\scriptsize, red!80, right=4pt, align=left] at (1.34,-1.34) {$P^\ast$\ (worst case)};

\filldraw[gray!55] (-0.8, 0.9) circle (2pt);
\filldraw[gray!55] (0.5,-0.6) circle (2pt);
\filldraw[gray!55] (-1.0,-0.4) circle (2pt);
\filldraw[gray!55] (0.6, 0.9) circle (2pt);

\end{tikzpicture}
\caption{The Wasserstein ambiguity set for distributionally robust RL. The nominal transition $P_0(\cdot\mid s,a)$ (dark-blue dot) sits at the centre of a Wasserstein ball $\mathcal{U}_\delta = \{P : \Wassp{p}(P, P_0) \le \delta\}$ inside the space of transition distributions $\mathcal{P}(\mathcal{S})$. Grey dots denote other candidate distributions inside the ball. The worst-case is represented by $P^\ast$ (red dot). Two modelling choices shape this game: the ground metric $d$ determines \emph{which} perturbations are considered admissible (through the $\Wassp{p}$ geometry), while the radius $\delta$ controls \emph{how far} models may move away from the nominal one, determining the conservatism of the resulting policy.}
\label{fig:ambiguity}
\end{figure}
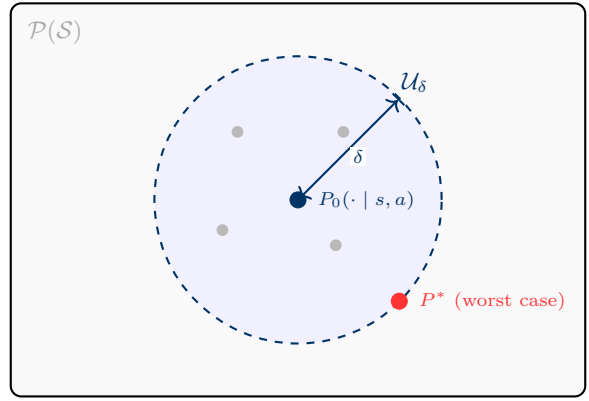

A canonical formulation (Figure~\ref{fig:ambiguity}) introduces, for each $(s,a)$, a Wasserstein ball around the nominal next-state distribution:
\[
\mathcal{U}_{\delta}(s,a)
=\Big\{P(\cdot\mid s,a)\in\mathcal{P}(\mathcal{S}):\ \Wassp{p}\!\big(P(\cdot\mid s,a),\,P_0(\cdot\mid s,a)\big)\le \delta\Big\},
\]
and collects these into an ambiguity set over transition kernels,
\[
\mathcal{U}_{\delta}
=\Big\{P:\ P(\cdot\mid s,a)\in\mathcal{U}_{\delta}(s,a)\ \ \forall (s,a)\in\mathcal{S}\times\mathcal{A}\Big\}.
\]
The distributionally robust control problem then replaces the Bellman expectation under $P_0$ with a worst-case expectation over $P\in\mathcal{U}_\delta$:
\[
V^\star(s)
=\max_{a\in\mathcal{A}}
\min_{P(\cdot\mid s,a)\in\mathcal{U}_{\delta}(s,a)}
\Big\{R(s,a)+\gamma\,\E_{s'\sim P(\cdot\mid s,a)}[V^\star(s')]\Big\}.
\]

The resulting control problem is minimax: the agent chooses actions to maximise return, while nature chooses the most damaging transition law inside the Wasserstein ball. The radius \textcolor{black}{$\delta$} controls how pessimistic this game becomes, and the ground metric determines which perturbations count as nearby enough to be adversarially relevant.
\citet{yang2017convex} \textcolor{black}{shows} how Kantorovich duality turns these robust Bellman updates into a finite-dimensional convex program when the nominal distribution has finite support, thereby avoiding direct optimisation over an infinite-dimensional ambiguity set. \citet{clement2021first} then develop first-order methods aimed at better scalability.

One concrete example is Wasserstein Robust Reinforcement Learning ($\text{WR}^2$L) \citep{abdullah2019wasserstein}, which imposes an expected Wasserstein constraint around a reference transition model $P_{0}(\cdot\mid s,a)$:
\[
\textcolor{black}{\max_{\theta}\ \min_{\phi\,\in\,\Phi_{\delta}(\theta)}\; \E_{\tau\sim p_{\theta}^{\phi}(\tau)}\!\Big[\sum_{t=0}^{T-1}\gamma^tR(s_t,a_t)\Big],
\qquad
\Phi_{\delta}(\theta)=\Big\{\phi:\ \E_{(s,a)\sim d_{\pi_\theta,\phi}}\!\big[ \Wtwo^{2}\big(P_{\phi}(\cdot\mid s,a),\,P_{0}(\cdot\mid s,a)\big) \big] \le \delta\Big\},}
\]
where the trajectory density $p_{\theta}^{\phi}(\tau)$ and state-action visitation distribution $d_{\pi,\phi}$ are induced by policy $\pi_\theta$ under transition model $P_\phi$. They approximate the $\Wtwo$ term using curvature information (via a Hessian-of-expectation approximation) and propose a corresponding optimisation procedure. Conceptually, this treats robustness as a \emph{budgeted} transport of mass in next-state space, with the metric encoding which perturbations are plausible.

The ambiguity-set idea extends naturally to stochastic control with uncertain exogenous disturbances, where duality-based reformulations again yield tractable Bellman operators with out-of-sample guarantees \citep{yang2017convex,yang2020wasserstein,kordabad2022safe,kandel2020safe,hou2020robust,queeney2024optimal}. More general distributionally robust stochastic optimisation results based on Wasserstein distance are established by \citet{gao2023distributionally}.

\subsection{Safety control via visitation regularisation}
\label{sec:robust-visitation}

A complementary approach targets safety more directly by shaping \emph{where} the policy visits. Let $\rho_{\mathrm{safe}}$ 
\textcolor{black}{denote a reference risk distribution that reflects the desired safety preference} (e.g., low mass near obstacles or high mass around safe corridors). One can then incorporate a transport-based penalty that encourages the policy to move its induced state visitation distribution toward (or away from) $\rho_{\mathrm{safe}}$ in the geometry induced by a chosen cost on states. Here OT is not modelling uncertainty in the environment; it is modelling how far the policy's visitation distribution is from a safer one.

In discrete state and action settings, this idea can be implemented as a regularisation term in value-based updates. \citet{baheri2023risk} modifies the standard Q-learning update by subtracting a state-level OT cost that measures how far the policy's visitation distribution has drifted from the safety prior:
\[
Q(s,a)\leftarrow Q(s,a)+\alpha\big[R(s,a)-\lambda\,\textcolor{black}{C(s)}+\gamma\max_{a'}Q(s',a')-Q(s,a)\big],
\]
\textcolor{black}{where $C(s)$ is described as an OT cost associated with state $s$ without a specified state-wise decomposition. Here $\lambda>0$ tunes the safety--return trade-off. In a follow-up study, \citet{shahrooei2024optimal} explicitly define pairwise transport contributions $C(s,s')=\Gamma^*_{s,s'}\|s-s'\|^2$ where $\Gamma^*$ is the optimal transport plan for $\Wtwo^2(\rho_{\pi},\rho_{\mathrm{safe}})$. These contributions enter
their update as an additive reward-shaping term.}
Experiments in \citet{shahrooei2024optimal} report practical safety gains on small discrete tasks.
This illustrates a common pattern in OT-based safety: rather than imposing hard constraints, one introduces a geometry-aware \emph{soft} mechanism that shifts probability mass away from risky decisions.

This only works as well as the safety prior $\rho_{\mathrm{safe}}$ is specified: if that prior is poor, the OT penalty can push learning in the wrong direction.

\subsection{Action uncertainty scores via entropic OT}
\label{sec:robust-uncertainty}

The visitation-based methods above assume a pre-specified risk distribution \textcolor{black}{$\rho_{\mathrm{safe}}$}. When such a prior is unavailable, one can instead read safety off the \emph{learning dynamics themselves}. Safety constraints can also be enforced implicitly by discouraging actions with unpredictable outcomes. Without requiring a reference risk distribution, \citet{shahrooei2025safety} quantify the uncertainty of discrete actions $\{a_i\}_{i=1}^N$ and encourage the agent to select actions with more predictable outcomes. The total uncertainty in a state $s$ is measured by a Wasserstein distance between action distributions derived from current and target Q-value estimates. 
The intuition is to look at how much the value-induced distribution over actions changes from one update to the next: if the mass has to be rearranged heavily, then the agent is still unsure about which actions should be preferred.
For each state $s$, they define the uncertainty score 
\[U(s,a_i)=\frac{\Delta(s,a_i)}{W_p(Q_s,T_t)}\]
\textcolor{black}{where $Q_s=\sum^N_{i=1}q_i^s\delta_{a_i}$ and $T_t=\sum^N_{i=1}p_i^t\delta_{a_i}$ are two discrete probability distributions over the available actions, with nonnegative normalised weights $q_i^s,p_i^t\ge 0$, $\sum_i q_i^s=\sum_i p_i^t=1$, obtained by normalising the corresponding value profiles across actions.} 
The first is induced by the current Q-values at state $s$, and the second by the \textcolor{black}{state-action-reward-state-action (SARSA)} target values after the latest transition. The OT coupling between these two action distributions is computed with Sinkhorn iterations. The term $\Delta(s,a_i)$ measures the transport imbalance associated with action $a_i$, namely how much mass linked to that action must be reassigned when passing from the current value profile to the target one: \textcolor{black}{with $\Gamma^*$ the entropic OT plan between $Q_s$ and $T_t$, \citet{shahrooei2025safety} define it as the absolute net flow imbalance at $a_i$, excluding self-transport, $\Delta(s,a_i)=\big|\sum_{b\neq a_i}\Gamma^*_{a_i,b}-\sum_{b\neq a_i}\Gamma^*_{b,a_i}\big|$. }
This uncertainty score is used to define the behaviour policy
\[
\pi^b(a\mid s) =
\begin{cases}
1-\epsilon+\frac{\epsilon}{|\mathcal{A}|} & \text{if } a=\argmax_{a'}[Q(s,a')-\beta U(s,a')]\\
\frac{\epsilon}{|\mathcal{A}|}  & \text{otherwise}.
\end{cases}
\]
\textcolor{black}{Here $\epsilon$ is the exploration rate of the $\epsilon$-greedy rule and is unrelated to the entropic coefficient $\varepsilon$.}
This provides an alternative route to risk-aware behaviour: instead of specifying a safe target distribution explicitly, the method penalises actions whose value estimates move erratically from one update to the next, with the strength controlled by $\beta$.
At a high level, the method is using OT as a detector of value instability across actions.

\subsection{Preference shift and Wasserstein DRO in RLHF}
\label{sec:robust-rlhf}

So far, $P_0$ has always been a transition law or a visitation distribution. The same ambiguity-set template transfers directly to the setting where the uncertain object is instead a \emph{preference distribution} from human annotators. Wasserstein ambiguity sets also arise in preference-based learning and RLHF for large language model (LLM) alignment (cf.\ the regularisation perspective in Section~\ref{sec:policy-trust}), where the deployed preference distribution may differ from the training distribution due to annotator heterogeneity, selection bias, or temporal drift. In such settings, distributionally robust optimisation (DRO) formulations treat the empirical preference distribution as nominal and optimise for worst-case preference shifts within an ambiguity set, including $p$-Wasserstein balls. The conceptual analogy with robust control is direct: instead of hedging against shifted transition laws, one now hedges against shifted preference data.

Concretely, \citet{xu2025robust} extend Direct Preference Optimisation (DPO) \citep{rafailov2023direct} and \citet{sahu2025dro} extend REBEL \citep{gao2024rebel} to a DRO objective of the form
\[
\min_{\theta}\max_{\mathbb{P}\in\mathcal{B}_{\delta}(\mathbb{P}^0;D)}
\E_{z\sim\mathbb{P}}[l_{\mathrm{DPO/REBEL}}(z;\theta)],
\]
where $\mathbb{P}^0$ is the nominal data-generating distribution, $\mathcal{B}_{\delta}(\mathbb{P}^0;D)$ is an ambiguity set defined by a distance $D$ (including Wasserstein choices), and $l(\cdot;\theta)$ is the preference-learning loss. From the OT perspective, this matches the same robustness template used for transition uncertainty (Section~\ref{sec:robust-ambiguity}): the modelling choices are the geometry defining plausible distribution shifts and the radius \textcolor{black}{$\delta$} controlling conservatism.

Beyond DRO, several recent works use OT more \emph{directly} as an alignment operator on preference or reward distributions. Distributional preference alignment via optimal transport formulates alignment updates by transporting between predicted and target preference/reward distributions rather than only constraining worst-case shifts \citep{melnyk2024aot}. Another fine-grained use of OT is at the \emph{token level}: OT couplings between token representations can be used to assign saliency/importance weights within preference pairs, biasing the optimisation toward the parts of responses that drive preference differences \citep{li2025otpo}. Finally, dual-reference “push--pull” distributional alignment uses two reference distributions (desired and undesired) and OT-based weighting to push the model toward the former while pulling it away from the latter during fine-tuning \citep{wang2026sot}.

While these works are not RL benchmarks in the classical sense, they are directly relevant to preference-based sequential decision-making because they provide concrete recipes for (i) modelling preference shift geometrically and (ii) propagating sparse feedback through an embedding space. An important open direction is how these OT-based alignment mechanisms extend to temporally extended preferences and trajectory-level feedback, where preferences are defined over sequences of states/actions rather than i.i.d.\ samples.

\subsection{Reward ambiguity and Wasserstein structure}
\label{sec:reward-ambiguity}

The preceding subsection hedged against uncertain preference \emph{data}. A closely related source of uncertainty lives one level up, in the reward function itself: even with clean preference data, many reward functions can explain the same observed behaviour. A central difficulty in IRL is therefore \emph{reward ambiguity}: many distinct reward (or cost) functions can induce the same expert policy (or an observationally indistinguishable occupancy). \citet{baheri2023understanding} proposes to study this ambiguity through a geometric lens by equipping the set of rewards consistent with the expert with a Wasserstein metric, so that ambiguity can be quantified by distances and dispersion in a transport geometry.

Let $\mathcal{R}(\pi_E)$ denote a set of reward functions consistent with the expert behaviour. The key modelling step is a representation map (denoted $\Phi$ in \citep{baheri2023understanding}; we write $\Xi$ here to avoid clashing with the survey-wide use of $\Phi$ for behavioural embeddings)
\[
\Xi:\mathcal{R}(\pi_E)\to \mathcal{P}(\mathcal{X}),\qquad R\mapsto \mu_R,
\]
\textcolor{black}{which embeds each candidate reward as a probability measure within the Wasserstein space $\mathcal{P}(\mathcal{X})$ on a ground space $\mathcal{X}$ equipped with a ground metric.}
Given candidates $\{R_i\}_{i=1}^N\subset\mathcal{R}(\pi_E)$, a ``centroid'' reward function representation can be identified as 
\[
\textcolor{black}{\eta^\star \in \argmin_{\eta\in\Xi(\mathcal{R}(\pi_E))}\ \sum_{i=1}^N \Wassp{p}\!\big(\eta,\mu_{R_i}\big).}
\]
\textcolor{black}{The terms in the sum are not squared, so this is a metric-median objective rather than the $p$-power barycentre of~\eqref{eq:barycentre}.}
The average pairwise distance $\frac{2}{N(N-1)}\sum_{i<j} \Wassp{p}(\mu_{R_i},\mu_{R_j})$ provides a quantitative notion of ``how ambiguous'' the rewards are under the chosen representation and ground metric, which shows the increasing importance of identifying ambiguity with the higher dimension of state-action space \citep{baheri2023understanding}.

This use of OT is conceptually orthogonal to the behaviour-matching pipelines in Section~\ref{sec:il}: here OT characterises \emph{reward-space structure}, whereas those methods use OT directly on \emph{behaviour distributions} (occupancies, trajectories, or embeddings). The notion of reward ambiguity connects directly to the Wasserstein ambiguity sets used for transition uncertainty (Section~\ref{sec:robust-ambiguity}) and preference shifts in RLHF (Section~\ref{sec:robust-rlhf}), though the latter concern transitions or preferences rather than reward functions. In practice, conclusions depend strongly on the choice of $\Xi$ and the metric on $\mathcal{X}$, which determine whether Wasserstein distances capture control-relevant modes of ambiguity in high-dimensional reward families \citep{baheri2023understanding}.
This ambiguity measure is representation-dependent and is not invariant to reward shaping unless \(\Xi\) encodes shaping equivalence.

\subsection{Transfer}
\label{sec:transfer}
Beyond robustness and safety, OT also appears in transfer across tasks and domains. Here the goal is no longer to hedge against worst-case perturbations, but to align source and target distributions so that policies, rewards, or representations transfer meaningfully under shift. The common thread is still geometric correspondence: OT specifies which source states, trajectories, or observations should count as close to which target ones. Robustness asks which shifts should be feared; transfer asks which correspondences should be trusted.

\paragraph{Goal/intent alignment.}
Zero-shot offline imitation \citep{pathak2018zero} aims to pretrain a task-agnostic policy and then adapt it to unseen tasks from only a single demonstration. \citet{rupf2024zero} do this by aligning state occupancies to a sequence of goals with OT. With the training set
\[
\mathcal{D}=\{(s_0,a_0,s_1,a_1,\dots)\}
\]
and a single rough (ordered but without time-step information) or partial (partial information is contained in each state) expert demonstration
\[
\mathcal{D}^e=(g_0,g_1,\dots ,g_M),
\]
where the goal $g$ is an abstract representation of state $s$, the optimisation problem is
\[
\pi^*=\argmin_{\pi}\mathrm{OT}_c(\hat{\rho}^{\pi}_{N},\hat{\rho}^{E}_{M}),
\]
where \textcolor{black}{$c(s,g)=-V^*(s,g)$ for a goal-conditioned value function $V^*$ trained from $\mathcal{D}$, and
  $\hat{\rho}^{\pi}_{N},\hat{\rho}^{E}_{M}$ are empirical estimates of state and goal occupancies. }
  \textcolor{black}{In the absence of goal abstraction, the induced state-space geometry is a quasimetric: nonnegative, satisfying the triangle inequality but not necessarily symmetry.}
  \textcolor{black}{The advantage of this formulation is that it aligns full occupancy patterns rather than greedily matching immediate goals, which reduces the myopia of earlier goal-conditioned methods. The paper also tests unbalanced OT, but does not find a significant gain in that setting.}

\paragraph{Preference transfer.}
Preference-based RL addresses reward design by learning from human comparisons between trajectories, but collecting such preferences for every new task is costly. \citet{liu2024pearl} therefore study zero-shot preference transfer. When tasks share an action space, they solve a GW problem between source and target trajectory sets and use the resulting alignment to transfer preference labels from source-task trajectory pairs to target-task pairs. GW is natural here because the state observations can differ across tasks, while their intra-trajectory relations may still be comparable.

Let us assume that there are trajectories $\{x_i\}_{i=1}^M$ for the source task with pairwise preferences $z(x_i,x_{i'})$ and trajectories $\{y_j\}_{j=1}^N$ for the target task. With weights $\sum_{i=1}^M u_i=1$ and $\sum_{j=1}^N v_j=1$, let $\Gamma$ be the optimal transport matrix for GW between two probability measures $\mu=\sum_{i=1}^M u_i\delta_{x_i}$ and $\nu=\sum_{j=1}^N v_j\delta_{y_j}$. Zero-shot Cross-task Preference Alignment and Robust Reward Learning (PEARL) \citep{liu2024pearl} defines the matching matrix for each pair of trajectories $(y_j,y_{j'})$ in the target task as $A^{jj'}=\Gamma_{\cdot j}\Gamma^T_{\cdot j'},$
where $\Gamma_{\cdot j}$ is the vector of transported mass from all source trajectories into $y_j$. The target trajectories are then labelled with preferences
\[
z(y_j,y_{j'})=\sum_i\sum_{i'\neq i}A^{jj'}_{ii'}z(x_i,x_{i'}),
\]
\textcolor{black}{so source pairs that align more strongly with a target pair contribute more heavily to the transferred label. In the practical algorithm, \citet{liu2024pearl} subsequently apply min-max normalisation to the transferred labels and binarise them at a threshold of 0.5.}

\paragraph{Visual/domain adaptation.}
In visual domains, the main challenge is to learn a representation that removes domain-specific appearance while preserving task-relevant content. Building on PPO and borrowing the critic-based Wasserstein idea from WGAN, Wasserstein Adversarial Proximal Policy Optimisation (WAPPO) \citep{roy2021visual} learns a representation function $h_\theta$ that approximates the optimal one by minimising the $\Wone$ distance between source and target feature distributions,
\[\Wone((h_\theta)_\#P_s,(h_\theta)_\#P_t)=\sup_{\|f\|_{\mathrm{Lip}}\le1}\{\E_{x\sim P_s}f(h_\theta(x))-\E_{x\sim P_t}f(h_\theta(x))\}\]
where $f$ is the adversarial critic (1-Lipschitz potential) trained to approximate the supremum.
Similarly, Wasserstein Distance Guided Representation Learning (WDGRL) \citep{shen2018wasserstein} adapts data to different domains by learning a shared feature space together with a task classifier on labelled source data. The domain discrepancy is measured by a Wasserstein critic in dual form, with the 1-Lipschitz constraint enforced by a gradient penalty.

Unbalanced minibatch OT has also been applied for domain adaptation to learn the representation function and the label function for the target domain \citep{fatras2021unbalanced} as mentioned in Section~\ref{sec:offline-ot}. Moreover, a Wasserstein-based reward has been proposed to select intermediate domains in continuous domain adaptation \citep{liu2025reinforced}.

\paragraph{Data filtering via OT scores.}
When a target domain has only limited offline data, OT-based alignment can identify which source-domain transitions are most compatible with the target dynamics before policy learning. Optimal Transport Data Filtering (OTDF) \citep{lyu2025cross} instantiates this in cross-domain offline policy adaptation by solving OT between empirical measures over transition tuples $u=(s,a,s')$ from a source dataset $\mathcal{D}_{\mathrm{src}}$ and $u'=(\bar s,\bar a,\bar s')$ from a (small) target dataset $\mathcal{D}_{\mathrm{tar}}$, using a cost $C(u,u')$ on $\mathcal{S}\times\mathcal{A}\times\mathcal{S}$. The optimal coupling $\Gamma^\ast$ yields a per-transition compatibility score (an expected cost under $\Gamma^\ast$), which is then used to filter or weight source transitions during critic learning. OTDF further adds an explicit dataset-constraint regulariser that encourages the learned policy to stay within the support of the target dataset, reducing bias toward the source behaviour when $\mathcal{D}_{\mathrm{tar}}$ is small \citep{lyu2025cross}. OT has also been explored in federated IRL settings, where expert demonstrations are distributed across clients, and OT aggregation can improve reward recovery under heterogeneity \citep{millard2026fedirl}.

\subsection{Summary}
\label{sec:robust-discussion}

Across the applications in this section, OT turns robustness and transfer questions into geometric ones. Ambiguity sets ask which transition or preference shifts should count as plausible. Visitation shaping asks how far the policy's visitation distribution is from a safer target distribution. Action-uncertainty scores ask which actions become unstable when value-induced transport changes sharply. Transfer reuses the same logic to align source and target distributions across tasks or domains. \textcolor{black}{Table~\ref{tab:summary7} collects} \textcolor{black}{representative} \textcolor{black}{methods discussed above and reports the main OT design choices.} In every case, the decisive modelling choices are the ground metric, which defines what kinds of shifts are plausible, and the scale parameter, such as \textcolor{black}{the ambiguity radius $\delta$, the penalty weight $\lambda$ or the
      uncertainty weight $\beta$}, which determines how conservative the method becomes. These methods fail when the ball radius, safety prior, or feature metric is poorly calibrated.

\section{Other applications}
\label{sec:extensions}

OT appears in RL well beyond the main themes of imitation, policy optimisation, world modelling, offline learning, and robustness. This section gathers those smaller but still instructive pockets of the literature: diversity and skill discovery, multi-agent coordination, curricula, and a handful of generative or evaluation settings. What links them is not a single algorithmic template, but a recurring move: replace an ad hoc similarity or diversity measure with a transport-based one on policy-induced distributions, rollout-derived empirical measures, or learned embeddings. The same diagnostic question applies throughout: which distributions are being compared, and what does the transport geometry add? This section is best read as frontier and adjacent applications; the evidence base is less mature than in imitation, offline RL, and policy optimisation. When a paper uses a bespoke distance denoted ``WD'', we interpret it according to the paper's definition as a Wasserstein distance $\Wassp{p}$ or a Sinkhorn-type surrogate.

\subsection{Multiple policies or skills}
\label{sec:diversity}
In unsupervised RL and skill discovery, the problem is often not how to imitate a target behaviour, but how to make several behaviours genuinely different from one another. OT offers a direct way to do this by separating the induced state distributions $p_l$ of concurrently trained policies. Here the transport distance itself becomes the diversity measure. \citet{he2022wasserstein} propose an intrinsic reward based on projected Wasserstein discrepancies \citep{rowland2019orthogonal} between state marginals, so that each policy is pushed away from its closest neighbour and state-space coverage improves. \citet{he2022wasserstein} define the final reward of an episode induced by policy $l$ as \textcolor{black}{$\min_{l'\neq l}\Wass(p_l,p_{l'})$} in practice, while the amortised reward for a single state $s_i$ is assigned as
\[
r(s_i) = \sum_{k=1}^m\sum_{j=1}^n c(s_j,s'_k)\Gamma^*_{i,k},
\]
where $\Gamma^*$ is the optimal transport matrix between two state distributions. \textcolor{black}{%
This credit-assignment rule is different from the standard row-wise decomposition of the OT cost in OTR. Moreover, OTR with the minus sign seeks closeness for imitation, while the reward defined here encourages separation for diverse skills. \citet{he2022wasserstein} use both a $\Wone$ dual estimator and a \emph{projected} Wasserstein discrepancy \citep{rowland2019orthogonal} for the amortised reward. In the projected variant, one-dimensional projections are used to determine the matching, while the matched samples are scored by their distances in the original state space. This differs from sliced Wasserstein distance, where the transport cost is evaluated after projection.}
Projecting states through a representation $f$ mitigates the curse of dimensionality and makes costs task-aware.
The practical interpretation is that a skill gets rewarded for visiting regions of state space that the other skills have not already claimed.

One can push the same idea further without explicitly constructing a per-step reward. \citet{yang2025task} use the Wasserstein distance to improve mutual-information skill learning by enforcing diversity and separability directly at the level of state distributions. Their objective, Wasserstein \textcolor{black}{SEParability} (WSEP), maximises
\[
\mathrm{WSEP}=\sum_{z_i\in\mathcal{Z}}\sum_{z_j\in \mathcal{Z}\backslash \{z_i\}}\Wass(p(S\mid z_i),p(S\mid z_j)),
\]
where $p(S\mid z)$ is the state distribution of a policy, and $\mathcal{Z}$ indexes skills. They further show that vertices of the feasible state-distribution polytope can be discovered via iterations of maximising projected WSEP,
\[
\mathrm{PWSEP}(i)=\min_{\lambda}\Wass\Big(p(S\mid z_i), \sum_{z_j\in\mathcal{Z}_i}\lambda^j p(S\mid z_j)\Big),
\]
where at iteration $i$, $\mathcal{Z}_i$ denotes previously learned skills and $\lambda$ are mixture coefficients with \textcolor{black}{$\lambda^j\ge 0$ and $\sum_{j\in\{0,1,\dots,i-1\}}\lambda^j=1$, so that the inner argument is a genuine
      mixture of the previously learned state distributions}. In practice, the sliced Wasserstein distance \citep{kolouri2019generalized} improves computational efficiency.
This can be read as a ``novelty against mixtures'' criterion: a new skill should not just differ from one old skill, but from combinations of the old ones.

Quality-diversity imitation learning aims to learn diverse behaviours from limited demonstrations. Wasserstein Quality Diversity Imitation Learning (WQDIL) \citep{yu2024imitation} improves reward-learning stability via Wasserstein adversarial training within the latent space of a Wasserstein auto-encoder (WAE). For WAE-WGAN, the objective is
\[
D_{\mathrm{WAE}}(P_X,P_G)=\inf_{Q(z|x)\in\mathcal{Q}}\E_{P_X}\E_{Q(z|x)}[c(x,\textcolor{black}{P_G(z)})]+\lambda \Wone(Q_z,P_z),
\]
where $P_X$ is the data distribution, \textcolor{black}{$P_G$ the decoder (generative model), $Q_z=\int Q(\cdot\mid
      x)\,\mathrm{d}P_X(x)$ the aggregate posterior over latents induced by encoding the demonstrations, and $P_z$
      the prior. WQDIL then adapts this latent-distribution matching idea to imitation learning to align the expert and policy latent distributions.} The first term is related to $\Wtwo^2$ with $c$ as the squared Euclidean cost. The corresponding loss for WAE-WGAIL is
\begin{align*}
\mathcal{L}(\phi,\psi)=&\E_{x^e\sim\mathcal{D},z^e\sim Q_{\phi}(z^e|x^e)}[c(x^e,G_{\psi}(z^e))]
+\E_{x^{\pi}\sim\pi,z^{\pi}\sim Q_{\phi}(z^{\pi}|x^{\pi})}[c(x^{\pi},G_{\psi}(z^{\pi}))] \\
&+\lambda \,\E_{z^e,z^{\pi}}[D_{\Wass}(z^e)-D_{\Wass}(z^{\pi})],
\end{align*}
where $Q_{\phi}$ is the encoder, $G_{\psi}$ the decoder, $D_{\Wass}$ the 1-Lipschitz potential function for the dual form of $\Wone$, and \textcolor{black}{$x=(s,a)$ for WAE-WGAIL or $x=(s,a,\delta(s))$ for measure-conditioned WAE-WGAIL, where $\delta(s)$ is \textcolor{black}{the Markovian Measure Proxy, a state-dependent single-step measure whose trajectory average yields the behavioural measure,}
instantiated in their locomotion tasks as the fraction of steps in which a given leg contacts the ground.}

Wasserstein distances are also applied to hierarchical RL to improve diversity among subpolicies \citep{li2023wasserstein}. For each subpolicy $\pi_k$, a regularisation term is added to the actor loss:
\[
\mathcal{L}(\pi_k)=-\min_{j\neq k}\Wass_\varepsilon(\mu_{\pi_k},\mu_{\pi_j}),
\]
where policies are represented as categorical distributions for discrete actions and Gaussians for continuous actions, with behavioural embeddings analogous to \citet{pacchiano2020learning} (cf.\ Section~\ref{sec:policy-bgpg}). Hierarchical policy blending as optimal transport (HiPBOT) \citep{le2023hierarchical} formulates blending multiple expert policies as an entropy-regularised unbalanced OT problem to derive mixture weights.

\subsection{Multi-agent RL}
\label{sec:marl}

Multi-agent RL (MARL) extends the single-agent setting to environments where several agents interact, each with its own reward function. A convenient formalism is a discounted Markov game
\[
\mathcal{G}=\bigl(\mathcal{S},\{\mathcal{A}_i\}_{i=1}^N, P, \{R_i\}_{i=1}^N,\gamma,p_0\bigr),
\]
with joint action space $\mathcal{A}=\prod_{i=1}^N \mathcal{A}_i$ and joint action $a=(a^1,\dots,a^N)$.
\textcolor{black}{We restrict attention to \emph{decentralised} (product) policies, for which the joint policy factorises as
      $\pi(a\mid s)=\prod_{i=1}^N \pi_i(a^i\mid s)$ in the fully observed case, or, under partial observability, $\pi(a\mid o)=\prod_{i=1}^N\pi_i(a^i\mid o^i)$ conditional on the joint observation $o=(o^1,\dots,o^N)$. Centralised or correlated joint policies do not factorise in this way.}
The joint discounted occupancy measure follows the single-agent definition~\eqref{eq:occupancy},
\[
\rho_\pi(s,a)=(1-\gamma)\sum_{t=0}^\infty \gamma^t\,\Pr(s_t=s,a_t=a\mid \pi),
\]
and per-agent marginals (useful when agents have local actions) are $\rho^i_\pi(s,a^i)=\sum_{a^{-i}}\rho_\pi(s,a)$.

\paragraph{Where OT can enter MARL objectives.}
Many OT ideas from the earlier sections extend naturally to MARL, for example trust regions on per-agent policies, Sinkhorn-regularised behaviour distributions, or visitation-shaping objectives. What changes is the object to which OT is applied. One must decide whether to regularise the joint occupancy $\rho_\pi$, per-agent marginals $\rho^i_\pi$, or some learned multi-agent embedding, and that choice determines both the geometry and the computational burden. This is the main conceptual fork in MARL: does OT compare the behaviour of the team as a whole, or the behaviour of each agent separately? \citet{baheri2024synergy} survey several such roles for OT in cooperative MARL, from aligning agent strategies to a collective objective, to treating resource allocation as a transport problem, to using Wasserstein drift between state distributions as a signal of nonstationarity.

\paragraph{Consensus via Sinkhorn barycentres.}
Coordination can be posed directly in distribution space rather than parameter space. \citet{baheri2025wasserstein} \textcolor{black}{computes} a Sinkhorn-divergence barycentre and \textcolor{black}{regularises} each agent toward it, encouraging consensus in visitation distributions while maintaining decentralised policies. A generic template consistent with this description is
\[
\bar\rho \in \argmin_{\rho}\ \sum_{i=1}^N \Seps{\varepsilon}\bigl(\rho_{\pi_i},\rho\bigr),
\qquad
\pi_i \leftarrow \argmax_{\pi_i}\ \E_{\rho_{\pi}}[R_{\text{common}}]\ -\ \lambda\,\Seps{\varepsilon}\bigl(\rho_{\pi_i},\bar\rho\bigr),
\]
where \textcolor{black}{$\rho_{\pi_i}:=\rho^i_\pi$ is the per-agent marginal defined above and $\bar\rho$ is the Sinkhorn
      barycentre acting as a consensus target. Note that $\rho^i_\pi$ depends on the full joint policy through the induced dynamics, not on $\pi_i$ alone. Where agents have different local action spaces, the barycentre requires a common space or explicit embeddings into one.}
The barycentre plays the role of a soft team average, and each agent is nudged toward that shared behavioural centre.

When the number of agents is large, mean-field formulations replace the joint interaction with a population distribution and can make OT geometry explicit: the population distribution becomes the evolving object, and Wasserstein/OT structure can be used to describe (and sometimes stabilise) its dynamics. Recent mean-field actor-critic flow approaches learn mean-field games by explicitly coupling policy/value updates to an evolution of the population distribution in a Wasserstein geometry, providing a concrete instance of ``learning dynamics over distributions’’ beyond single-agent RL \citep{zhou2025mfac}. This line aligns with the survey’s gradient-flow viewpoint (Section~\ref{sec:policy-wgf}), but with the population distribution playing the role of the transported measure. The Wasserstein distance can also express robustness in many-agent settings: ERNIE \citep{bukharin2023robust} proposes adopting Wasserstein constraints in a mean-field approximation to handle many-agent uncertainty, turning nonstationarity into a distributional uncertainty set over population measures and suggesting a natural bridge to the ambiguity-set constructions of Section~\ref{sec:robust-ambiguity}. A closely related federated-RL setting uses Wasserstein barycentres to aggregate heterogeneous client policies \citep{pereira2025fedwb}.

\subsection{Curricula via distributional interpolation or constraints}
\label{sec:curricula}
OT can generate intermediate tasks that bridge an easy source to a difficult target, replacing direct transfer with a graduated curriculum. Here OT is being used to answer a curriculum-design question: which task should come next, given both where the target lies and what the current policy can already solve?

Contextual RL can be modelled via a contextual MDP $\mathcal{M}(c)=(\mathcal{S},\mathcal{A}, P_c,R_c,\gamma_c, \rho_{0,c})$, \textcolor{black}{where $c$ is a context variable, $\rho_{0,c}$ is the context-dependent initial-state distribution (the
      analogue of the global $p_0$ of Section~\ref{sec:rl-brief}, which is recovered as
      $p_0=\int\rho_{0,c}\,\mathrm{d}\mu(c)$),} and the training objective becomes $\E_{\mu(c)}[J(\pi, c)]$, with $\mu(c)$ defining the target task distribution. Curriculum RL generates a sequence of intermediate task distributions $p_i$ designed to improve exploration and final performance on $\mu(c)$.
The key tension is that the next task should be close enough to the target to be useful, but not so hard that the current policy cannot learn from it.

Constrained OT with a $\Wtwo$ proximity constraint generates intermediate task distributions that are close to the target while remaining solvable by the current policy \citep{klink2022curriculum,klink2024benefit}:
\begin{equation*}
\textcolor{black}{p^\ast \in \argmin_{p \in \mathcal{P}(\mathcal{C})} \; \Wtwo(p, \mu)
\quad \text{s.t.} \quad
J(\pi, c) \geq \delta_J \;\; \text{for } p\text{-almost every } c \in \mathcal{C},
\quad \Wtwo(p, q) \leq \delta_c,}
\end{equation*}
\textcolor{black}{where the transport distances are between the context \emph{measures} $p,\mu,q\in\mathcal{P}(\mathcal{C})$
      rather than between the scalar density values $p(c),\mu(c)$; $\delta_J$ is the desired performance threshold,
      required $p$-almost everywhere, which is the measure-theoretic reading of the original implication $p(c)>0\Rightarrow J(\pi,c)\ge\delta_J$ (it strengthens to a condition on $\operatorname{supp} p$ when $J(\pi,\cdot)$ is continuous); and $\delta_c$ controls deviation from the
      previous curriculum $q$, written with a subscript to distinguish it from the ambiguity radius $\delta$ of
      Section~\ref{sec:robust-ambiguity}.}

\citet{huang2022curriculum} propose GRADIENT, which defines a $\pi$-contextual distance that accounts for policy performance to interpolate between context distributions while maintaining feasibility; in practice, they approximate the contextual metric when exact computation is costly. They also use 1-Wasserstein barycentres between distributions $\mu_0$ and $\mu_1$:
$\eta_{\alpha} = \argmin_{\eta} \, (1-\alpha) \Wone(\mu_{0}, \eta)
+ \alpha \Wone(\eta, \mu_{1})$, where $\alpha$ controls the weights.

\paragraph{Alternative curricula: HGG.}
In goal-conditioned settings, Hindsight Goal Generation (HGG) \citep{ren2019exploration} selects intermediate goals that are closer (in a transport-aware sense) to the target goal while remaining achievable by the current policy, yielding more sample-efficient curricula than simple Hindsight Experience Replay (HER) \citep{andrychowicz2017hindsight} which relabels transitions with achieved goals to mitigate the challenge of sparse rewards. \textcolor{black}{Let $\mathcal{G}$ be the goal space. HGG works with a finite \emph{selected set}
      $Z\subset\mathcal{S}\times\mathcal{G}$ of $|Z|=K$ initial-state/goal pairs and its associated empirical measure
      $\zeta_Z=\frac{1}{K}\sum_{(s,g)\in Z}\delta_{(s,g)}\in\mathcal{P}(\mathcal{S}\times\mathcal{G})$; the target is
      a reference measure $\zeta^*\in\mathcal{P}(\mathcal{S}\times\mathcal{G})$. Keeping the set and the measure
      distinct matters, because the objective below both transports $\zeta_Z$ and tests membership in $Z$, and a
      general density on $\mathcal{S}\times\mathcal{G}$ need not be bounded by $1$.} The agent seeks to maximise $J(\pi;\zeta^*)=\E_{(s,g)\sim\zeta^*}[J(\pi,s,g)]$. To generate goals $\zeta$, HGG optimises
\begin{align*}
    \textcolor{black}{\max_{\pi,\; Z : |Z| = K} J(\pi;\zeta_Z) - L \cdot \mathrm{OT}_c(\zeta_Z, \zeta^*)} \quad
    \text{s.t.} \quad
    & \sum_{s_0, s_t \in \tau} \mathbf{1}\!\left[(s_0, \phi(s_t)) \in \textcolor{black}{Z}\right] \leq 1, \quad \forall \tau \in \mathcal{B}, \\
    & \sum_{\tau \in \mathcal{B}} \sum_{s_0, s_t \in \tau} \mathbf{1}\!\left[(s_0, \phi(s_t)) \in \textcolor{black}{Z}\right] = K,
\end{align*}
where $K$ trajectories are sampled from the replay buffer $\mathcal{B}$, $\phi$ maps states to goals, \textcolor{black}{$L>0$ is the Lipschitz constant of the value function with respect to the metric $c$ on initial-state/goal pairs, so that $|V^{\pi}(s,g)-V^{\pi}(s',g')|\le L\,c((s,g),(s',g'))$.} \textcolor{black}{\citet{ren2019exploration} assume this and use it to obtain $J(\pi;\zeta')\ge J(\pi;\zeta)-L\,\mathrm{OT}_c(\zeta,\zeta')$, which motivates the transport-regularised surrogate, although $L$ is unknown in practice and treated as a hyperparameter. Here} $\mathrm{OT}_c$ is the OT distance with cost
\[
c((s,g),(s',g'))=\alpha\lVert \phi(s) - \phi(s') \rVert_{2} + \lVert g - g' \rVert_{2},
\]
where $\alpha>0$ is a weight for the trade-off.
This optimisation is solved by iterating between improving the policy and updating the hindsight distribution $\zeta$ (solving a Wasserstein barycentre problem via bipartite matching). Safety-prioritising curricula extend this framework to constrained RL, generating intermediate task distributions that explicitly balance learning progress against safety constraints \citep{koprulu2025scg}.

\subsection{Further OT-RL intersections}
\label{sec:other-apps}

Several OT uses fall outside the main templates of the survey but are still worth recording. In generative and sequence-based settings, for example, Sinkhorn-based decoders have been used for non-autoregressive trajectory generation in combinatorial domains such as the travelling salesman problem, where Sinkhorn acts as a differentiable approximation to a permutation or assignment operator \citep{goh2022combining}. \citet{chen2020sequence} add a Wasserstein regulariser between generated and reference sequence embeddings in a learned semantic space, and \citet{sun2025score} use OT-guided matching inside a score-based diffusion policy. The common pattern is that OT is not comparing occupancies or Bellman targets directly; it is supplying a structured discrepancy inside a generative model.

OT also provides useful signals for evaluation and exploration. \citet{nkhumise2025studying} use the $\Wone$ distance between successive occupancy measures $\Wone(\rho_{\pi_k},\rho_{\pi_{k+1}})$ to assess the learning efficiency of RL algorithms, defining optimality measures based on the ratio of total transport effort to final policy improvement. As an intrinsic control signal, \citet{durugkar2021wasserstein} propose Wasserstein Distance Maximising Intrinsic Control, where each skill is trained to maximise the Wasserstein distance between its state-visitation distribution and its start-state distribution, encouraging controllable, wide-coverage behaviour without hand-crafted reward shaping, complementing the mutual-information skill discovery methods of Section~\ref{sec:diversity}. 

Finally, OT-based robustness and structural tools extend to settings adjacent to standard RL. In contextual bandits, which can be read as a one-step special case of RL, \citet{shen2024wasserstein} develop Wasserstein distributionally robust policy evaluation and learning through tractable OT-dual objectives. The same ambiguity-set logic could naturally feed into robust off-policy evaluation in sequential settings. Two further directions show how far OT now reaches beyond the standard RL template: dynamic OT has been used to derive conditions for counterfactual identifiability via continuous-time flows \citep{ribeiro2025counterfactual}, with possible relevance to causal RL and offline evaluation, and optimal transport has been used to fuse task-specific models by aligning feature-space geometry under OT plans rather than interpolating parameters directly \citep{pan2025otmf}, enabling continual fusion without revisiting past tasks.

\subsection{Summary}
\label{sec:extensions-discussion}

\textcolor{black}{Table~\ref{tab:summary8} summarises} \textcolor{black}{main methods reviewed in this section.} Even in these more specialised settings, the same two questions keep reappearing: \emph{what} distributions are being compared, and \emph{what} geometry makes that comparison meaningful? Once those are fixed, the rest of the design usually follows. In multi-agent settings, one further choice becomes unavoidable: should OT act on joint distributions, which capture coordination but scale poorly, or on marginals, which are cheaper but may miss interactions between agents? The methods in this section should be read as promising but less settled: several trade statistical efficiency or wall-clock cost for a more structured notion of diversity, alignment, or evaluation.

\section{Discussion}
\label{sec:discussion}

Across Sections~\ref{sec:il}--\ref{sec:extensions}, the same modelling choices and failure modes recur. This section gathers them in one place: we discuss the main trade-offs, practical guidance, 
and open problems. Method-level details remain in Tables~\ref{tab:summary3}--\ref{tab:summary8}; the goal here is synthesis rather than cataloguing.

\subsection{Cross-cutting trade-offs}
\label{sec:discussion-tradeoffs}

The same tensions keep reappearing across applications. They are best read as design trade-offs rather than universal rules, and they cluster around three recurring themes: sharpness versus stability, the role of representation and temporal structure, and the calibration of the hyperparameters that mediate both.

The first tension is between \emph{sharpness} and \emph{stability}. $\Wone$ dual critics remain informative when distributions barely overlap, which is one reason OT is introduced in adversarial imitation, offline behaviour constraints, and robust ambiguity sets. The price is fragility: Lipschitz enforcement is delicate, and critics can overfit small minibatches \citep{arjovsky2017wasserstein,gulrajani2017improved}. Entropic OT is smoother, easier to minibatch, computationally efficient, and often more stable, but biased unless debiased through the Sinkhorn divergence $\Seps{\varepsilon}$ \citep{genevay2018learning}. The same tension propagates to the algorithmic level: dual-critic $\Wone$ methods such as WAIL and BRAC-W depend heavily on critic capacity and optimisation balance; map-based $\Wtwo$ methods avoid adversarial training but impose convexity constraints through ICNN parameterisations \citep{brenier1991polar,omura2025offline}; primal Sinkhorn objectives sidestep both at the price of making the entropic parameter a sensitive modelling choice. OT does not remove the usual optimisation trade-offs; it relocates them into the choice of solver and geometry, and in practice the way OT is \emph{solved} often matters as much as the RL algorithm it is inserted into.

The second tension concerns what the ground cost actually encodes. The effectiveness of OT geometry is primarily determined by the geometry encoded by the ground cost and the learned features when the cost is defined via a representation space. Task-specific learned features can improve performance in high-dimensional or unstructured observation spaces. For example, \citet{bobrin2024align} train the temporal goal-conditioned value function through ICVF (intent-conditioned value function) \citep{ghosh2023reinforcement}, and use the backward embedding $\varphi$ as representations for OT, where $\|\varphi(s_i)-\varphi(s_{j})\|^2$ estimates the average path length from state $s_i$ to $s_j$, which measures the reachability and is robust to noise in the original state space. \citet{pacchiano2020learning} instead suggest a Behavioural Embedding Map (BEM) to abstract behaviours from trajectories or states and apply different BEMs in different tasks and settings.
The representation becomes a critical design choice for optimisation, and the performance may degrade under embedding-distribution shift even with frozen representations during training.
Temporal structure raises a parallel question of granularity: bag-of-states matching is cheap and effective for short, well-aligned episodes (e.g.\ OTR, Section~\ref{sec:il-otr}), but creates spurious correspondences under speed variation or partial observability. Banded masks and context windows recover local order at moderate cost (Section~\ref{sec:temporal}), while fully dynamic couplings remain expensive. The same geometry-versus-ordering tension resurfaces wherever OT operates on sequential data: offline reward labelling (Section~\ref{sec:offline}), curricula (Section~\ref{sec:curricula}), and cross-domain transfer (Section~\ref{sec:transfer}).
Empirical Wasserstein estimates can converge slowly in high dimensions, so raw observations are rarely the right transport space. This motivates feature costs, slicing, minibatches, and entropic or unbalanced relaxations, but each changes the geometry being optimised.

Overlaying both is a \emph{calibration} problem that the OT-RL literature largely avoids. The entropic parameter $\varepsilon$, the radius $\delta$ in ambiguity sets (Section~\ref{sec:robust}), the transported mass $m$ in partial OT (Section~\ref{sec:offline}), penalty strengths in unbalanced OT, and proximity radius in curricula (Section~\ref{sec:curricula}) all require calibration. \textcolor{black}{There is no \emph{unified} calibration procedure across the surveyed OT-RL settings, and practical calibration with dependent, adaptively collected, high-dimensional RL data and approximate models remains substantially less developed.}

\subsection{When OT helps---and when it does not}
\label{sec:discussion-when}

\textcolor{black}{\paragraph{Cross-cutting implications.}
The taxonomy separates distributional objectives that are easy to conflate. \textcolor{black}{State-conditioned action-distribution penalties constrain local policy changes; occupancy measure comparisons assess visitation and sequential behaviour;}
return-distribution losses model the variability of the return, whereas Q-posteriors represent uncertainty in the value estimate. Guarantees about performance or behavioural equivalence do not follow from the choice of
transported object alone, but from the additional assumptions imposed and design choices for each method. This helps explain why bisimulation-style results appear for transition-level objectives (Sections~\ref{sec:bisimulation} and~\ref{sec:modelbased-latent}) and why trajectory-level tasks require temporal
structure in the coupling or the cost (Section~\ref{sec:temporal}). Across these settings, geometry and temporal
structure are modelling choices rather than consequences of using OT. The proximity-baseline results of
\citet{dong2025optimal} make the narrower empirical point that some reward-labelling gains do not require an
optimal coupling, which motivates comparisons holding the representation, reward transformation and downstream
algorithm fixed. Entropic regularisation, debiasing, slicing, minibatch objectives and restricted coupling
families can each change the optimisation problem, unlike exchanging one exact solver for another at a fixed
objective. Useful comparisons therefore isolate the transported object, the ground cost, the temporal structure
and the numerical approximation, rather than treating OT as a single interchangeable component.}

OT is most useful when distributions overlap weakly, as in early-training imitation or under severe distribution shift; when the state or action space has a geometry that KL divergence cannot express; or when domains are heterogeneous enough that no sensible pointwise cross-domain cost exists and Gromov-Wasserstein alignment becomes the natural alternative \citep{memoli2011gromov,fickinger2021cross}. It is also a natural fit when the algorithm needs an explicit transport plan, for example for reward labelling, data filtering, or transport-map policies, and when targets are multi-modal enough that the mode-seeking behaviour of a reverse KL objective, $\KL(\pi\|\pi_{\mathrm{target}})$, becomes unhelpful.

OT is also easy to overuse. When policy and target distributions already overlap well, for example once an imitation policy is already close to the expert, KL- and JS-based gradients are often informative enough and cheaper. When the action space carries no meaningful geometry, a KL-based trust-region or trust-region-like updates such as TRPO and PPO often work perfectly well without the extra modelling burden \citep{schulman2015trust,schulman2017proximal}. And if the ground cost does not encode anything task-relevant, the geometric machinery mostly adds overhead. Recent work makes that point explicitly: proximity-based and lightweight temporal baselines can match OT reward labelling on broad benchmarks \citep{dong2025optimal}, which raises the practical question of when coupling structure is really necessary rather than simply elegant. Computation sharpens the same issue. In very-small-sample regimes, minibatch Wasserstein estimates are biased and high-variance \citep{fatras2019learning,bellemare2017cramer}, and in compute-limited settings the inner-loop cost of Sinkhorn iterations, critic training, or ICNN updates may not justify itself. OT is least attractive when demonstrations already overlap with the learner, action geometry is uninformative, behaviour support is too narrow to leave safely, or a Wasserstein radius cannot be calibrated.

\subsection{Practitioner's guide}
\label{sec:discussion-guide}

Table~\ref{tab:decision-guide} is meant as a starting point rather than a prescription. It asks a practical question: if the RL problem looks like this, which OT default is the least bad place to begin? The point is not that these defaults are optimal, but that they provide a reasonable first experiment before heavier tuning. These defaults assume a meaningful ground cost and enough samples to estimate the transport signal.

\begin{table}[t]
\centering
\caption{Decision guide: \textcolor{black}{the authors' suggested starting points} by application area\textcolor{black}{. These are reasonable first experiments supported by the evidence discussed in Sections~\ref{sec:il}--\ref{sec:extensions}, not empirically established best defaults across problem classes}.}
\label{tab:decision-guide}
\small
\begin{tabular}{@{}p{2.5cm}p{2.5cm}p{2.5cm}p{2.2cm}p{3.3cm}@{}}
\toprule
\textbf{Application} & \textbf{Start with} & \textbf{Distributions} & \textbf{Cost default} & \textbf{Key pitfall} \\
\midrule
Online imitation & $\Seps{\varepsilon}$ (Sinkhorn div.) & occupancies $\rho_\pi$ vs $\rho_{\pi_E}$ & learned features & $\varepsilon$ sensitive to cost scale \\
Offline reward labelling & entropic OT + mask & trajectory segments & cosine on encoder & spurious matches without temporal mask \\
Policy trust region & per-state Sinkhorn & $\pi(\cdot|s)$ vs $\pi_{\mathrm{old}}(\cdot|s)$ & Euclidean (actions) & radius too tight $\to$ no improvement \\
Offline constraint & $\Wone$ penalty & $\pi(\cdot|s)$ vs $\pi_{\beta}(\cdot|s)$ & action-space metric & overly strong $\to$ clones $\pi_{\beta}$ \\
World model & $\Wone$ or $\Seps{\varepsilon}$ & $P_{\phi}(\cdot|s,a)$ vs $P(\cdot|s,a)$ & latent features & optimising prediction, not control \\
Robustness & $\Wassp{p}$ ball & transitions $P(\cdot|s,a)$ around $P_0$ & env.\ geometry & uncalibrated \textcolor{black}{$\delta$} $\to$ over-conservatism \\
Transfer / filtering & OT scores + soft reweight & behaviour embeddings & shared repr.\ space & hard thresholding with noisy costs \\
Curriculum & barycentres / OT path & task distributions & task metric & infeasible interpolated tasks \\
\bottomrule
\end{tabular}
\end{table}

\textcolor{black}{The computational form of OT depends on whether the compared distributions have finite, structured continuous, or general continuous support. For finite discrete distributions, the exact transport could be solved by linear programming over a transport matrix, although its cost grows rapidly with the support size \citep{peyre2019computational}. Continuous distributions do not necessarily require numerical discretisation. In one dimension, $\Wassp{p}$ admits an analytic representation based on quantile functions, avoiding the need to solve the full OT problem \citep{dabney2018distributional}. Another special case is $\Wtwo$, which has a closed-form expression for structured distribution families such as Gaussian distributions under the squared Euclidean ground cost \citep{terpin2022trust, likmeta2023wasserstein}. The exact OT problem between general continuous distributions is infinite-dimensional, and OT is typically estimated from samples using empirical measures, trading computational tractability for sampling error \citep{song2023provably, sun2024distributional}. Methods such as entropic regularisation and sliced Wasserstein distances are widely used to improve computational efficiency, at the cost of modifying or approximating the original OT problem \citep{papagiannis2022imitation, sebag2023combining}.}

\textcolor{black}{
\paragraph{An environment-level view of the evidence.}
The comparisons above are methodological. Readers who want to know \emph{where} OT has actually been applied, and
with what reported effect, may find Table~\ref{tab:evidence} more useful. It lists a small number of representative
studies and records, for each, the environment and data regime, \textcolor{black}{what OT is used for and how it is implemented, the reported outcome and comparison \emph{inside} that paper, and the noted limitation. Outcomes are summarised qualitatively in the context of each paper's own experimental or theoretical setting rather than across studies, since differences in tasks, data, experimental setups and evaluation protocols make direct cross-paper comparison inappropriate. }
The Evidence column marks whether a contribution has empirical (E) or theoretical (T) support.
The role labels classify the representative studies in this table, while the per-section summary tables record method characteristics rather than evidence status.
}

Two practical themes cut across these defaults. The first is that the ground cost deserves as much attention as the OT solver. Task-specific features can induce a more appropriate geometry than raw observations, but mis-specified features may distort the metric structure and degrade downstream performance.
Heterogeneous units (e.g.\ joint angles and end-effector positions) should be normalised before being combined, and costs should be scaled before setting $\varepsilon$ because entropic regularisation is sensitive to magnitude. For action spaces, it is usually more important to encode task semantics and anisotropy than to default to the isotropic Euclidean distance. Gromov-Wasserstein is best reserved for cases where a pointwise cross-domain cost is genuinely ill-posed; otherwise, a shared representation with standard $\Wassp{p}$ or Sinkhorn objectives is simpler and usually more stable \citep{shen2018wasserstein,lyu2025cross}.

The second theme is that most failures in OT-based RL are numerical or representational rather than algorithmic. Sinkhorn objectives benefit from log-domain iterations and from scaling $\varepsilon$ relative to a typical pairwise cost, and warm-starting dual variables helps when the compared distributions move only gradually. Dual $\Wone$ critics need careful Lipschitz control through gradient penalties or spectral normalisation, and should not be grossly over-parameterised relative to the minibatch size \citep{arjovsky2017wasserstein,gulrajani2017improved}; $\Wtwo$ and ICNN-based maps require nonnegative weights for convexity and some regularisation for smoothness \citep{makkuva2020icnn}. At the trajectory level, masks and context windows are usually safer than unconstrained long-horizon couplings, with the window size chosen according to the dynamics of the environment. Minibatch OT should generally be run on features rather than raw observations, and unbalanced losses are often more robust to subsampling than balanced ones \citep{fatras2021unbalanced}. When OT appears as an auxiliary term, e.g. an intrinsic reward, diversity bonus, or filtering weight, annealing its contribution over training can be beneficial as coverage improves. When costs are noisy, soft reweighting is safer than hard thresholding.

\subsection{Open problems}
\label{sec:discussion-open}

Four open directions stand out.

\paragraph{When does coupling structure actually matter?}
As noted in Section~\ref{sec:discussion-when}, good geometry can matter more than solving OT: many papers gain as much from representation design as from the transport formulation, and proximity baselines can match OT reward labelling on broad benchmarks \citep{dong2025optimal}.  What is missing are controlled experiments that isolate the transport component from the representation component, and theoretical characterisations of the regimes, including multi-modal demonstrations, systematic mass mismatch, severe temporal misalignment and heterogeneous domains, where coupling structure is genuinely necessary rather than merely elegant.

\paragraph{Scalability and amortisation.}
Current OT-RL methods invoke transport solvers per update, per state, per agent, or per minibatch.  Sliced and projected approximations reduce this cost while retaining meaningful geometry \citep{kolouri2019generalized,mahey2023fastswgg,chapel2025dgswplans}, and amortised or neural OT methods learn transport maps or potentials once and reuse them across iterations \citep{geuter2025unot,chen2025fast,makkuva2020icnn,korotin2023neuralot}.  Flow-matching constructions offer a complementary route, estimating dynamics gaps via Wasserstein distances between source and target transitions \citep{kong2025compflow}.  But the deeper challenge is temporal: scaling OT beyond short windows and masks to long-horizon and high-dimensional control remains largely unexplored.

\paragraph{Mass mismatch as a default.}
Real offline datasets and demonstrations are almost always incomplete, mixed-quality, or multi-domain, so balanced OT is often the wrong model.  Partial and unbalanced OT are natural alternatives \citep{asadulaev2024rethinking,chizat2018scaling,lyu2025cross}, and minibatch unbalanced OT mitigates some of the bias inherent in minibatch estimation \citep{fatras2021unbalanced}.  The missing piece is principled calibration: selecting transported mass and marginal-penalty strengths from observable diagnostics, such as coverage statistics, critic uncertainty and held-out value error, with guarantees analogous to those available for KL regularisation.

\paragraph{Theory for OT-regularised updates.}
Guarantees for Wasserstein trust regions, Wasserstein natural gradients, and OT-based mirror descent remain thin under realistic function approximation \citep{zhang2018policy,moskovitz2021efficient,li2018natural,pfau2025wasserstein}.  Recent algorithms that place Wasserstein geometry at the core of the update, e.g.\ Wasserstein Policy Optimisation, make this gap more pressing.  Understanding stability, sample complexity, and approximation error for Wasserstein-proximal updates, relative to the mature theory for KL-based TRPO and \textcolor{black}{natural policy gradient (NPG)} methods, likely requires analyses that separate representation error (ground-cost misspecification) from optimisation error (critic, Sinkhorn, or map approximation).

\medskip
OT has given RL a useful geometric vocabulary for comparing distributions. The next step is to understand when that geometry genuinely changes the problem for the better, and when it mainly adds machinery. Answering that requires better calibration tools, better scalability, and theory that separates the effect of transport geometry from the effect of representation learning or optimisation heuristics.

\newpage

\bibliographystyle{tmlr}
\bibliography{main}

\newpage

\appendix

\section*{Appendix: notation and summary tables} 

\begingroup
\small
\setlength{\LTleft}{0pt}
\setlength{\LTright}{0pt}
\setlength{\LTpre}{0pt}
\setlength{\LTpost}{0pt}
\setlength{\tabcolsep}{6pt}
\renewcommand{\arraystretch}{1.12}
\newcolumntype{Y}[1]{>{\raggedright\arraybackslash}p{#1}}

\begin{longtable}{@{}>{$}p{0.20\textwidth}<{$} Y{0.75\textwidth}@{}}
\caption{Notation used throughout the survey.}
\label{tab:notation}\\
\toprule
\multicolumn{2}{@{}l@{}}{\textbf{MDP and RL}}\\
\midrule
\endfirsthead

\toprule
\multicolumn{2}{@{}l@{}}{\textbf{MDP and RL (continued)}}\\
\midrule
\endhead

\midrule
\multicolumn{2}{r@{}}{\textit{Continued on next page}}\\
\endfoot

\bottomrule
\endlastfoot

\mathcal{S},\mathcal{A} & State and action spaces.\\
\mathcal{M}=(\mathcal{S},\mathcal{A},P,R,\gamma,p_0) & Discounted Markov decision process (MDP).\\
\gamma & Discount factor.\\
p_0\ (\text{or }\nu) & Initial-state distribution (default $p_0$; $\nu$ when varying the start distribution).\\
P(\cdot\mid s,a) & Environment transition kernel.\\
P_\pi & Trajectory law generated by $(p_0, \pi, P)$.\\
P_0(\cdot\mid s,a),\ P_\phi(\cdot\mid s,a) & Nominal and learned transition model / world model (parameters $\phi$).\\
R(s,a),\ r_t & Reward function and realised reward $r_t:=R(s_t,a_t)$ \textcolor{black}{(deterministic-reward convention; see Section~\ref{sec:rl-brief} and the distributional Bellman operator of Section~\ref{sec:policy-critics}, where $R(s,a)$ is read as a random reward)}.\\
\pi,\ \pi_{\theta} & Policy (possibly parametrised by $\theta$).\\
\pi_{\beta},\ \pi_E & Behaviour/data-collection policy; expert policy.\\
\tau & Trajectory, e.g.\ $(s_0,a_0,r_0,s_1,\dots)$.\\
t,\ T & Time index; horizon/trajectory length (as used in finite-horizon transports).\\
G_t & Discounted return from time $t$.\\
J(\pi) & Expected return objective.\\
V^\pi(s),\ Q^\pi(s,a) & State-value and action-value under policy $\pi$.\\
V_{\theta_v}(s),\ Q_{\theta_q}(s,a) & Learned approximations (critics) to $\textcolor{black}{V^\pi,Q^\pi}$.\\
A^\pi(s,a) & Advantage function.\\
\rho_\pi(s,a) & Normalised discounted occupancy measure; $\rho_\pi(s)=\sum_a\rho_\pi(s,a)$.\\
\hat{p}(s) & Empirical state occupancy measure of a single finite trajectory.\\
\mathcal{D} & Offline dataset of transitions $(s,a,r,s')$.\\

\midrule
\multicolumn{2}{@{}l@{}}{\textbf{Optimal transport}}\\
\midrule
\mathcal{P}(\mathcal{X}) & Probability measures on a space $\mathcal{X}$.\\
\mu,\nu & Generic probability measures.\\
F & \textcolor{black}{Measurable map $\mathcal{X}\to\mathcal{Y}$. A \emph{transport map} additionally satisfies $F_{\#}\mu=\nu$.}\\ %
\Gamma & Coupling / transport plan on $\mathcal{X}\times\mathcal{Y}$.\\
\Pi(\mu,\nu) & Set of couplings with marginals $\mu$ and $\nu$.\\
c(x,y),\ d(x,y) & Ground cost and ground metric.\\
\Wassp{p}\ (\text{e.g. } \Wone,\Wtwo) & $p$-Wasserstein distance (in particular $\Wone,\Wtwo$).\\
\varepsilon & Entropic regularisation parameter.\\
\Wassp{\varepsilon} & \textcolor{black}{Entropy-regularised (unrooted) transport cost. } \\ %
\Sink_\varepsilon & Sinkhorn divergence (debiased entropic OT)\textcolor{black}{; positive-definite under conditions on $c$, but not a metric}.\\
\KL(\cdot\|\cdot) & Kullback-Leibler divergence.\\
u,\ v & OT dual potentials (Kantorovich dual).\\
\textcolor{black}{f} & \textcolor{black}{Single OT dual potential: the $c$-transform potential (Section~\ref{sec:ot}), the $1$-Lipschitz critic in $\Wone$ duals (Sections~\ref{sec:il-adversarial},~\ref{sec:offline-ot}), and the nonnegative multiplier in the partial-OT dual (Section~\ref{sec:offline-ot}). }\\ %
m & Transported mass in partial OT.\\
\lambda_1,\lambda_2 & Marginal-penalty weights in unbalanced OT.\\
\eta & Wasserstein barycentre\textcolor{black}{; $\lambda_k\ge 0$, $\sum_k\lambda_k=1$ are its weights}.\\
\textcolor{black}{\mathcal{H}(\Gamma)} & \textcolor{black}{Coupling entropy, $-\sum_{ij}\Gamma_{ij}\log\Gamma_{ij}$. Distinct from the causal policy entropy $H(\pi)$.}\\
d_{\mathcal{X}},d_{\mathcal{Y}};\ \mathrm{GW} & Intra-space dissimilarities; Gromov-Wasserstein objective.\\

\midrule
\multicolumn{2}{@{}l@{}}{\textbf{Imitation / embeddings / distributional RL}}\\
\midrule
H(\pi) & Discounted causal entropy (max-ent IRL).\\
\psi;\ \psi^\ast & IRL cost regulariser and convex conjugate (as in $\mathrm{IRL}_\psi$).\\
\Phi,\ \Phi_{\#} & Embedding map and pushforward.\\
\mu_\pi & Behavioural distribution $\mu_\pi=\Phi_{\#}P_\pi$.\\
\hat p,\ \hat p_\pi,\ \hat p_E & Empirical state marginals of trajectories (learner/expert).\\
Z^\pi(s,a) & Return distribution (distributional RL).\\
\mathcal{T}^\pi & (Distributional) Bellman operator.\\
\textcolor{black}{\mathfrak{T}} & \textcolor{black}{Space of trajectories $\tau$.}\\
\textcolor{black}{\mathcal{B}_{\mathrm{lat}}} & \textcolor{black}{A latent behaviour space.}\\
\end{longtable}
\endgroup

\vspace{15pt}

\begingroup
\setlength{\extrarowheight}{3pt}
\footnotesize
\setlength{\LTleft}{0pt}
\setlength{\LTright}{0pt}
\setlength{\LTpre}{0pt}
\setlength{\LTpost}{0pt}
\setlength{\tabcolsep}{6pt}
\renewcommand{\arraystretch}{1.10}
\newcolumntype{L}[1]{>{\raggedright\arraybackslash}p{#1}}
\newcolumntype{C}[1]{>{\centering\arraybackslash}p{#1}}

\begin{longtable}{@{}L{0.22\textwidth} L{0.13\textwidth} L{0.20\textwidth} C{0.10\textwidth} L{0.23\textwidth}@{}}
\caption{Representative OT uses for imitation learning and reward inference.}
\label{tab:summary3}\\
\toprule
\textbf{Method (area)} & \textbf{OT role} & \textbf{Compared objects} & \textbf{OT form} & \textbf{Implementation / notes}\\
\midrule
\endfirsthead

\toprule
\textbf{Method (area)} & \textbf{OT role} & \textbf{Compared objects} & \textbf{OT form} & \textbf{Implementation / notes}\\
\midrule
\endhead

\midrule
\multicolumn{5}{r@{}}{\textit{Continued on next page}}\\
\endfoot

\bottomrule
\endlastfoot

\multicolumn{5}{@{}l@{}}{\textit{Adversarial occupancy matching and discriminator-based IL}}\\
\parbox[t]{\linewidth}{WAIL \citep{xiao2019wasserstein}, WDAIL \citep{zhang2020wasserstein}\\
(\,generative adversarial IL\,)}
& Discriminator / reward shaping
& State-action occupancy measures
& $\Wone$ (Euclidean cost)
& Kantorovich duality with 1-Lipschitz critic (WGAN-style training).\\

\parbox[t]{\linewidth}{W-IRL \citep{wu2023w}\\
(\,IRL\,)}
& Discriminator / reward shaping
& State marginals
& $\Wone$ (Euclidean cost)
& Kantorovich duality with a Lipschitz-constrained critic. \\

\parbox[t]{\linewidth}{LWAIL \citep{yang2026latent}\\
(\,adversarial IL; \textcolor{black}{twin delayed DDPG (TD3)}\,)}
& Latent discriminator + reward
& Latent state-pair (transition) occupancies (ICVF-pretrained features)
& $\Wone$ (Euclidean cost)
& Duality-based critic in latent space; pretraining via ICVF as proposed.\\

\parbox[t]{\linewidth}{SIL \citep{papagiannis2022imitation}\\
(\,adversarial IL\,)}
& Discriminator / reward shaping via primal OT
& State-action occupancy measures
& Entropy-regularised $\Wassp{\varepsilon}$ (OT-GAN-style cost \citep{salimans2018improving})
& Sinkhorn iterations in minibatches; entropic OT objective.\\

\parbox[t]{\linewidth}{PWIL \citep{dadashi2020primal}\\
(\,IL\,)}
& Reward-from-transport (non-adversarial)
& State-action occupancy measures
& Upper bound on $\Wone$
& Uses approximate primal coupling (e.g.\ POT \citep{flamary2024pot}); transport cost converted to rewards.\\

\addlinespace[0.35em]
\multicolumn{5}{@{}l@{}}{\textit{Offline reward labelling and temporal/intent extensions}}\\
\parbox[t]{\linewidth}{OTR \citep{luo2023optimal} (\citep{chang2023imitation,haldar2023watch, wang2023offline, zare2023leveraging})\\
(\,Offline RL\,)}
& Reward labelling
& Trajectory state marginals / occupancies
& Entropic OT (cosine cost)
& Sinkhorn algorithm on state features; labels used for downstream RL.\\

\parbox[t]{\linewidth}{\citet{dong2025optimal}\\
(\,IRL benchmark\,)}
& Empirical comparison / baseline analysis
& Trajectory state marginals (replicates OTR-style labelling)
& Various (reproduces $\Wone$, entropic, cosine)
& Argues proximity baselines can match OT-based reward labelling; provides controlled comparisons across cost and solver choices.\\

\parbox[t]{\linewidth}{OOPS \citep{chang2023imitation} \\(\,ILfO\,)}
& Reward labelling
& Trajectory transition occupancies
& Entropic $\Wone$ (square root of the Euclidean distance)
& Sinkhorn algorithm.\\

\parbox[t]{\linewidth}{PW-DICE \citep{yan2024pwdice}\\(\,offline Learning from Observations (LfO)\,)}
& Weighted BC with potentials solved by OT
& State occupancy measures
& primal $\Wone$ (a weighted sum of state occupancy with the Euclidean distance between state embeddings as the cost metric)
& A smooth convex dual objective \\

\parbox[t]{\linewidth}{SMMOTIL \citep{sebag2023combining} \\(\,IRL\,)}
& Reward labelling
& Agent and multiple expert trajectories (state occupancies)
& Sliced multi-marginal $\Wtwo$ 
& Monte Carlo estimation.\\

\parbox[t]{\linewidth}{TemporalOT \citep{fu2024robot}\\
(\,\textcolor{black}{online IL from observation}\,)}
& Reward labelling with temporal constraints
& Visual embeddings with temporal mask/context cost
& Masked entropic OT
& Sinkhorn algorithm with masked coupling and smoothed costs.\\

\parbox[t]{\linewidth}{AILOT \citep{bobrin2024align}\\
(\,Offline RL\,)}
& Reward labelling with intent-aware costs
& Intention/state representations
& Entropy-regularised $\Wassp{\varepsilon}$
& Sinkhorn algorithm with temporal-dependent cost as proposed.\\

\addlinespace[0.35em]
\multicolumn{5}{@{}l@{}}{\textit{Cross-domain imitation}}\\
\parbox[t]{\linewidth}{GWIL \citep{fickinger2021cross}\\
(\,cross-domain IL\,)}
& Cross-domain reward / matching
& State-action occupancy measures via relations
& GW distance
& GW solvers with entropic regularisation \citep{peyre2016gromov}, building on regularised standard OT \citep{ferradans2014regularized}.\\

\end{longtable}
\endgroup

\vspace{15pt}

\begingroup
\setlength{\extrarowheight}{3pt}
\footnotesize
\setlength{\LTleft}{0pt}
\setlength{\LTright}{0pt}
\setlength{\LTpre}{0pt}
\setlength{\LTpost}{0pt}
\setlength{\tabcolsep}{6pt}
\renewcommand{\arraystretch}{1.10}
\newcolumntype{L}[1]{>{\raggedright\arraybackslash}p{#1}}
\newcolumntype{C}[1]{>{\centering\arraybackslash}p{#1}}

\begin{longtable}{@{}L{0.22\textwidth} L{0.13\textwidth} L{0.20\textwidth} C{0.10\textwidth} L{0.23\textwidth}@{}}
\caption{Representative OT methods for online policy optimisation and closely related objectives.}
\label{tab:summary4}\\
\toprule
\textbf{Method (setting)} & \textbf{OT role} & \textbf{Compared objects} & \textbf{OT form} & \textbf{Implementation / notes}\\
\midrule
\endfirsthead

\toprule
\textbf{Method (setting)} & \textbf{OT role} & \textbf{Compared objects} & \textbf{OT form} & \textbf{Implementation / notes}\\
\midrule
\endhead

\midrule
\multicolumn{5}{r@{}}{\textit{Continued on next page}}\\
\endfoot

\bottomrule
\endlastfoot

\multicolumn{5}{@{}l@{}}{\textit{Policy update constraints and regularisers}}\\
\parbox[t]{\linewidth}{WPO, SPO \citep{song2023provably}\\
(\,on-policy; discrete\,)}
& Trust-region constraint
& Per-state action distributions $\pi'(\cdot\mid s)$ vs.\ $\pi(\cdot\mid s)$
& $\Wone$ / Sinkhorn
& Per-state Lagrangian dual; closed-form distribution update under task-dependent action costs (e.g.\ $0$--$1$, $\mathrm{L}_1$).\\

\parbox[t]{\linewidth}{OT-TRPO \citep{terpin2022trust}\\
(\,TRPO-style\,)}
& Trust-region constraint
& Action distributions (e.g.\ Gaussians)
& $\Wtwo$
& Closed-form $\Wtwo$ for Gaussians under quadratic costs; dual optimisation over the multiplier.\\

\parbox[t]{\linewidth}{BGPG, BGES \citep{pacchiano2020learning}\\
(\,PG / ES\,)}
& Behaviour regulariser
& Embedding distributions $\mu_{\pi_{\theta}}$ vs.\ $\mu_{\mathrm{ref}}$
& Entropy-regularised $\Wassp{\varepsilon}$
& Dual OT test functions on embedding space; OT signal used as shaping/weights in PG or ES updates.\\

\parbox[t]{\linewidth}{WPR \citep{na2026semantic}\\
(\,RLHF\,)}
& Policy regularisation 
& learned policy $\pi_{\theta}$ vs. reference policy $\pi_{ref}$
& Entropy-regularised $\Wassp{\varepsilon}$
& Dual formulation with Sinkhorn-Knopp iterations yields a tractable semantic regulariser compatible with PPO-style updates.\\

\addlinespace[0.35em]
\multicolumn{5}{@{}l@{}}{\textit{Wasserstein geometry: gradient flows and natural gradients}}\\
\parbox[t]{\linewidth}{WGF \citep{richemond2017wasserstein}\\
(\,bandit / 1-step setting\,)}
& Gradient flow
& Action distributions (bandit / 1-step setting)
& $\Wtwo$
& JKO/proximal viewpoint; policy updates via Wasserstein measure evolution.\\

\parbox[t]{\linewidth}{WGF \citep{zhang2018policy}\\
(\,trajectory-measure view\,)}
& Gradient flow
& Policy-induced measures (e.g.\ trajectory laws)
& $\Wtwo$
& JKO/proximal viewpoint; policy updates via Wasserstein measure evolution.\\

\parbox[t]{\linewidth}{WNG (WNPG / WNES) \citep{moskovitz2021efficient}\\
(\,parametric policies\,)}
& Natural gradient
& Policy-induced measures / behaviours
& $\Wtwo$
& Wasserstein information geometry; kernel/low-rank approximations yield practical natural-gradient steps (policy-gradient and evolution-strategies variants).\\

\parbox[t]{\linewidth}{Wasserstein policy optimisation \citep{pfau2025wasserstein}\\
(\,actor-critic\,)}
& Gradient-flow update
& Policy distributions
& $\Wtwo$
& Discretises a Wasserstein gradient flow and projects to a parametric policy class; the implemented algorithm keeps KL regularisation in practice.\\

\multicolumn{5}{@{}l@{}}{\textit{Distributional RL}}\\
\parbox[t]{\linewidth}{QR-DQN \citep{dabney2018distributional}\\
(\,distributional RL\,)}
& Bellman updates
& Return distributions $Z^\pi(s,a)$ (action-return)
& $\Wone$
& Quantile-based approximation via fixed quantiles / quantile regression.\\

\parbox[t]{\linewidth}{SinkhornDRL \citep{sun2024distributional}\\
(\,distributional RL\,)}
& Return-distribution update
& Particle/empirical return distributions
& Sinkhorn divergence
& Sinkhorn iterations on particle sets (debiased entropic OT).\\

\addlinespace[0.35em]
\multicolumn{5}{@{}l@{}}{\textit{Value learning and OT-regularised critics}}\\
\parbox[t]{\linewidth}{WQL \citep{metelli2019propagating}\\
(\,value learning\,)}
& Wasserstein TD
& $Q$-posteriors vs.\ targets
& $\Wtwo$ barycentres
& Closed-form barycentre updates under Gaussian/particle approximations; uncertainty propagation via Wasserstein TD.\\

\parbox[t]{\linewidth}{WAC \citep{likmeta2023wasserstein}\\
(\,\textcolor{black}{soft actor-critic (SAC)}\,)}
& Critic update
& Gaussian $Q$-posterior vs.\ target
& $\Wtwo$
& Closed-form $\Wtwo$ for Gaussian families; actor exploits posterior structure.\\

\parbox[t]{\linewidth}{WAVE \citep{baheri2025wave}\\
(\,actor-critic\,)}
& Stabilise critic drift
& Critic-output distributions across iterations
& Entropic OT
& Sinkhorn OT penalty between successive critic outputs; adaptive weighting.\\

\end{longtable}
\endgroup

\vspace{15pt}

\begingroup
\setlength{\extrarowheight}{3pt}
\footnotesize
\setlength{\LTleft}{0pt}
\setlength{\LTright}{0pt}
\setlength{\LTpre}{0pt}
\setlength{\LTpost}{0pt}
\setlength{\tabcolsep}{6pt}
\renewcommand{\arraystretch}{1.10}
\newcolumntype{L}[1]{>{\raggedright\arraybackslash}p{#1}}
\newcolumntype{C}[1]{>{\centering\arraybackslash}p{#1}}

\begin{longtable}{@{}L{0.22\textwidth} L{0.13\textwidth} L{0.20\textwidth} C{0.10\textwidth} L{0.23\textwidth}@{}}
\caption{Representative OT methods for model-based RL and Wasserstein world models.}
\label{tab:summary5}\\
\toprule
\textbf{Method (setting)} & \textbf{OT role} & \textbf{Compared objects} & \textbf{OT form} & \textbf{Implementation / notes}\\
\midrule
\endfirsthead

\toprule
\textbf{Method (setting)} & \textbf{OT role} & \textbf{Compared objects} & \textbf{OT form} & \textbf{Implementation / notes}\\
\midrule
\endhead

\midrule
\multicolumn{5}{r@{}}{\textit{Continued on next page}}\\
\endfoot

\bottomrule
\endlastfoot

\multicolumn{5}{@{}l@{}}{\textit{Value-awareness}}\\
\parbox[t]{\linewidth}{VAML \citep{farahmand2017value,asadi2018equivalence}\\
(\,model-based\,)}
& Value-aware transition model loss
& $P(\cdot\mid s,a)$ vs.\ $P_\phi(\cdot\mid s,a)$
& $\Wone$
& Equivalence between Lipschitz test-function losses and a $\Wone$ discrepancy on next-state distributions.\\

\multicolumn{5}{@{}l@{}}{\textit{Link to bisimulation metric}}\\
\parbox[t]{\linewidth}{\citep{calo2024bisimulation}\\(dynamic programming)}
& Efficient OT distances for Markov processes
& Stationary Markov processes (occupancy of state pairs)
& Discounted $\Wone$
& Sinkhorn value iteration; Markov-structured couplings improve computation and convergence.\\

\multicolumn{5}{@{}l@{}}{\textit{Latent MDP training}}\\
\parbox[t]{\linewidth}{WAE-MDP \citep{delgrange2023wasserstein}\\
(\,latent world model\,)}
& OT-based model learning
& Trace/trajectory distributions (encoded/decoded)
& $\Wone$-type
& WAE/WGAN-style objective; dual-critic realisation with gradient penalty in practice.\\

\parbox[t]{\linewidth}{OT-MDP \citep{ropke2025integrating}\\
(\,latent world model\,)}
& Regularised OT objective
& Real vs.\ latent trajectories / transitions
& OT-based
& Approximates via total variation; discretised latent states with straight-through estimators (as proposed).\\

\end{longtable}
\endgroup

\vspace{15pt}

\begingroup
\setlength{\extrarowheight}{3pt}
\footnotesize
\setlength{\LTleft}{0pt}
\setlength{\LTright}{0pt}
\setlength{\LTpre}{0pt}
\setlength{\LTpost}{0pt}
\setlength{\tabcolsep}{6pt}
\renewcommand{\arraystretch}{1.10}
\newcolumntype{L}[1]{>{\raggedright\arraybackslash}p{#1}}
\newcolumntype{C}[1]{>{\centering\arraybackslash}p{#1}}

\begin{longtable}{@{}L{0.22\textwidth} L{0.13\textwidth} L{0.20\textwidth} C{0.10\textwidth} L{0.23\textwidth}@{}}
\caption{Representative OT uses for offline policy optimisation. 
}
\label{tab:summary6}\\
\toprule
\textbf{Method (area)} & \textbf{OT role} & \textbf{Compared objects} & \textbf{OT form} & \textbf{Implementation / notes}\\
\midrule
\endfirsthead

\toprule
\textbf{Method (area)} & \textbf{OT role} & \textbf{Compared objects} & \textbf{OT form} & \textbf{Implementation / notes}\\
\midrule
\endhead

\midrule
\multicolumn{5}{r@{}}{\textit{Continued on next page}}\\
\endfoot

\bottomrule
\endlastfoot

\parbox[t]{\linewidth}{BRAC \citep{wu2019behavior}\\
(\,offline RL\,)}
& Policy regularisation / value penalty
& Action distributions $\pi(\cdot\mid s)$ vs.\ $\pi_{\beta}(\cdot\mid s)$
& $\Wone$
& Dual form with a discriminator (WGAN-style critic).\\

\parbox[t]{\linewidth}{Q-DOT \citep{omura2025offline}\\
(\,offline RL\,)}
& Policy regularisation via transport map
& Action distributions (state-conditional)
& $\Wtwo$
& ICNN parameterisation of a convex potential / transport map.\\

\parbox[t]{\linewidth}{VGF \citep{xu2026reinforcement}\\
(\,offline RL\,)}
& Gradient flow
& Action distributions (state-conditional)
& $\Wtwo$
& \textcolor{black}{JKO scheme, realised by an SVGD-style RKHS-restricted particle update rather than exact JKO minimisation; implicit behaviour control via a transport budget.} \\

\parbox[t]{\linewidth}{PPL \citep{asadulaev2024rethinking}\\
(\,offline RL\,)}
& Partial behaviour cloning / selective matching
& Action distributions under partial OT constraint
& Partial OT
& Maximin formulation with a tunable partiality parameter.\\

\parbox[t]{\linewidth}{\citet{givchi2021distributionally}\\
(\,RL (policy optimisation)\,)}
& Unbalanced OT formulation
& Projected state and action marginals of occupancies
& Unbalanced OT
& Multiple convex constraints handled via Dykstra’s algorithm; Bregman (KL) divergences used in practice.\\

\parbox[t]{\linewidth}{\citet{fatras2021unbalanced}\\
(\,Domain adaptation\,)}
& Domain-invariant representation learning
& Source vs. target distributions
& Unbalanced minibatch OT
& POT and Geomloss packages\\

\end{longtable}
\endgroup

\vspace{15pt}

\begingroup
\setlength{\extrarowheight}{3pt}
\footnotesize
\setlength{\LTleft}{0pt}
\setlength{\LTright}{0pt}
\setlength{\LTpre}{0pt}
\setlength{\LTpost}{0pt}
\setlength{\tabcolsep}{6pt}
\renewcommand{\arraystretch}{1.10}
\newcolumntype{L}[1]{>{\raggedright\arraybackslash}p{#1}}
\newcolumntype{C}[1]{>{\centering\arraybackslash}p{#1}}

\begin{longtable}{@{}L{0.22\textwidth} L{0.13\textwidth} L{0.20\textwidth} C{0.10\textwidth} L{0.23\textwidth}@{}}
\caption{Representative OT uses for control shift and transfer.}
\label{tab:summary7}\\
\toprule
\textbf{Method (area)} & \textbf{OT role} & \textbf{Compared objects} & \textbf{OT form} & \textbf{Implementation / notes}\\
\midrule
\endfirsthead

\toprule
\textbf{Method (area)} & \textbf{OT role} & \textbf{Compared objects} & \textbf{OT form} & \textbf{Implementation / notes}\\
\midrule
\endhead

\midrule
\multicolumn{5}{r@{}}{\textit{Continued on next page}}\\
\endfoot

\bottomrule
\endlastfoot

\multicolumn{5}{@{}l@{}}{\textit{Distributionally robust control}}\\
\parbox[t]{\linewidth}{WR$^2$L \citep{abdullah2019wasserstein}\\
(\,robust RL\,)}
& Minimax constraint
& Transition kernels $P(\cdot\mid s,a)$ within a Wasserstein ball
& $\Wtwo$
& Approximates $\Wtwo$ via a Hessian-of-expectation surrogate; zero-order optimisation as proposed.\\

\parbox[t]{\linewidth}{DRMDP \citep{kordabad2022safe,yang2020wasserstein,hou2020robust,queeney2024optimal,yang2017convex}\\
(\,robust control / robust RL\,)}
& Ambiguity set (Wasserstein ball)
& Disturbance or transition distributions
& OT-cost uncertainty set (generalises $\Wassp{p}$)
& Kantorovich duality yields tractable robust Bellman operators / reformulations.\\

\parbox[t]{\linewidth}{FOM for DRMDP \citep{clement2021first}\\
(\,robust control / robust RL\,)}
& Ambiguity set (Wasserstein ball)
& Disturbance or transition distributions
& $\Wtwo$
& First-order methods.\\

\addlinespace[0.35em]
\multicolumn{5}{@{}l@{}}{\textit{Safety}}\\
\parbox[t]{\linewidth}{\citep{baheri2023risk, shahrooei2024optimal}\\
(\,safe RL\,)}
& Visitation shaping / regularisation
& Risk distribution vs.\ state visitation distribution
& Wasserstein with squared Euclidean cost
& OT-derived costs used as penalties in value-based updates (as proposed).\\

\parbox[t]{\linewidth}{\citet{shahrooei2025safety}\\
(\,safe RL (SARSA)\,)}
& Action uncertainty scoring
& Action distributions induced by current vs.\ target Q estimates
& Entropic OT
& Sinkhorn iterations compute uncertainty scores via an OT map; used to define safer behavioural policy.\\

\addlinespace[0.35em]
\multicolumn{5}{@{}l@{}}{\textit{RLHF / preference alignment}}\\
\parbox[t]{\linewidth}{\citet{xu2025robust}, \citet{sahu2025dro}\\
(\,RLHF\,)}
& Wasserstein DRO on preferences
& Preference distributions (nominal vs.\ shifted)
& $\Wassp{p}$ ball
& Robust DPO/REBEL with Wasserstein ambiguity sets over preference distributions.\\

\parbox[t]{\linewidth}{OTPO \citep{li2025otpo}\\
(\,RLHF\,)}
& Token-level saliency weighting
& Token representations within preference pairs
& Unbalanced entropic OT
& OT couplings assign importance weights to tokens; biases optimisation toward preference-driving content.\\

\parbox[t]{\linewidth}{SOT \citep{wang2026sot}\\
(\,safety-aware \textcolor{black}{supervised fine-tuning (SFT)}, not RLHF\,)}
& Push-pull distributional alignment
& Model output vs.\ desired/undesired reference distributions
& OT-based weighting
& Dual-reference OT: pushes toward desired, pulls from undesired distribution during fine-tuning.\\

\addlinespace[0.35em]
\multicolumn{5}{@{}l@{}}{\textit{Reward ambiguity}}\\
\parbox[t]{\linewidth}{\citet{baheri2023understanding}\\
(\,IRL\,)}
& Reward ambiguity analysis
& Representations of feasible reward functions
& Wasserstein distance
& Computes Wasserstein structure on reward representations; no implementation reported.\\

\addlinespace[0.35em]
\multicolumn{5}{@{}l@{}}{\textit{Transfer}}\\
\parbox[t]{\linewidth}{ZILOT \citep{rupf2024zero}\\
(\,zero-shot offline IL\,)}
& Goal/intent alignment for transfer
& State and goal occupancy measures
& \textcolor{black}{$\mathrm{OT}_c$ (learned goal-conditioned cost)}
& Sinkhorn algorithm with \textcolor{black}{the negated goal-conditioned value function} used as ground cost\textcolor{black}{; possibly asymmetric, so not a Wasserstein metric}.\\

\parbox[t]{\linewidth}{PEARL \citep{liu2024pearl}\\
(\,preference-based RL\,)}
& Preference transfer
& Trajectory sets via preference relations
& GW
& Sinkhorn-style solver for GW alignment; transfers preferences by matching intra-trajectory structure.\\

\parbox[t]{\linewidth}{WDGRL \citep{shen2018wasserstein}\\
(\,domain adaptation\,)}
& Domain-invariant representation learning
& Feature distributions (source vs.\ target)
& $\Wone$
& Adversarial Wasserstein alignment (WGAN-style duality).\\

\parbox[t]{\linewidth}{WAPPO \citep{roy2021visual}\\
(\,RL under visual shift\,)}
& Domain-agnostic representation for control
& Feature distributions across domains
& $\Wone$
& Dual formulation; minimises feature-distribution Wasserstein distance during policy learning.\\

\parbox[t]{\linewidth}{\citet{liu2025reinforced}\\
(\,domain selection\,)}
& Reward shaping for intermediate-domain choice
& Domain-specific feature distributions
& Wasserstein distance
& Uses Wasserstein-based reward to guide selection of intermediate domains (as proposed).\\

\parbox[t]{\linewidth}{OTDF \citep{lyu2025cross}\\
(\,cross-domain offline RL\,)}
& Data filtering / reweighting
& State-action-next-state triples
& Entropic $\Wassp{\varepsilon}$ / Sinkhorn
& Sinkhorn-based scoring used to filter or reweight data (OTR-style).\\

\end{longtable}
\endgroup

\vspace{15pt}

\begingroup
\setlength{\extrarowheight}{3pt}
\footnotesize
\setlength{\LTleft}{0pt}
\setlength{\LTright}{0pt}
\setlength{\LTpre}{0pt}
\setlength{\LTpost}{0pt}
\setlength{\tabcolsep}{6pt}
\renewcommand{\arraystretch}{1.10}
\newcolumntype{L}[1]{>{\raggedright\arraybackslash}p{#1}}
\newcolumntype{C}[1]{>{\centering\arraybackslash}p{#1}}

\begin{longtable}{@{}L{0.22\textwidth} L{0.13\textwidth} L{0.20\textwidth} C{0.10\textwidth} L{0.23\textwidth}@{}}
\caption{Representative OT uses in multi-agent RL and other applications.}
\label{tab:summary8}\\
\toprule
\textbf{Method (area)} & \textbf{OT role} & \textbf{Compared objects} & \textbf{OT form} & \textbf{Implementation / notes}\\
\midrule
\endfirsthead

\toprule
\textbf{Method (area)} & \textbf{OT role} & \textbf{Compared objects} & \textbf{OT form} & \textbf{Implementation / notes}\\
\midrule
\endhead

\midrule
\multicolumn{5}{r@{}}{\textit{Continued on next page}}\\
\endfoot

\bottomrule
\endlastfoot

\multicolumn{5}{@{}l@{}}{\textit{Diversity and hierarchical RL}}\\
\parbox[t]{\linewidth}{WURL \citep{he2022wasserstein}\\
(\,unsupervised RL\,)}
& Intrinsic reward for diversity
& State marginal distributions across policies/skills
& \textcolor{black}{$\Wone$}
& \textcolor{black}{Dual form and projected Wasserstein discrepancy amortise rewards from OT plans.}\\

\parbox[t]{\linewidth}{PWSEP \citep{yang2025task}\\
(\,unsupervised RL\,)}
& Skill separation / discovery
& State distributions conditioned on skills
& Wasserstein (sliced approximation)
& Uses sliced Wasserstein distance for efficient estimation of separability objective.\\

\parbox[t]{\linewidth}{WQDIL \citep{yu2024imitation}\\
(\,quality-diversity IL\,)}
& Stable reward learning in latent space
& Latent distributions of demonstrations vs.\ policy trajectories
& WAE-WGAN
& Duality-based adversarial training within WAE latent space (as proposed).\\

\parbox[t]{\linewidth}{WDER \citep{li2023wasserstein}\\
(\,hierarchical RL\,)}
& Diversity regulariser for subpolicies
& Behavioural embeddings / action distributions
& Smoothed $\Wassp{\varepsilon}$ with random-feature approximation
& Duality with random-feature approximations enabling scalable updates.\\

\parbox[t]{\linewidth}{HiPBOT \citep{le2023hierarchical}\\
(\,hierarchical control\,)}
& Policy blending (mixture weights)
& Expert-policy and agent-policy priors (temperature-weighted)
& Unbalanced entropic OT
& Sinkhorn-like scaling algorithm \citep{chizat2018scaling}.\\

\multicolumn{5}{@{}l@{}}{\textit{Multi-agent RL}}\\
\parbox[t]{\linewidth}{\citet{baheri2024synergy}\\
(\,MARL\,)}
& Conceptual integration points
& Policy-induced distributions over state/action (various)
& $\Wass_p$ / Sinkhorn (various)
& Overview of potential OT roles for coordination, scalability, and robustness (position paper).\\

\parbox[t]{\linewidth}{\citet{baheri2025wasserstein}\\
(\,cooperative MARL\,)}
& Coordination via barycentre regularisation
& Joint state-action visitation measures
& Sinkhorn divergence barycentre
& Iterative barycentre computation; sliced-Wasserstein approximation used for scalability.\\

\parbox[t]{\linewidth}{ERNIE \citep{bukharin2023robust}\\
(\,MARL\,)}
& Robustness (mean-field idea)
& Distributions over states (mean-field approximation)
& Wasserstein (proposed)
& Proposes adopting Wasserstein distance in the mean-field constraint; not implemented in practice (as noted by authors).\\

\addlinespace[0.35em]
\multicolumn{5}{@{}l@{}}{\textit{Curricula}}\\
\parbox[t]{\linewidth}{CURROT \citep{klink2022curriculum, klink2024benefit}\\
(\,curriculum RL\,)}
& Curriculum generation
& Task/context distributions
& Constrained $\Wtwo$ distance
& Linear assignment between particle sets; shortest-path / performance pseudo-distances as ground costs.\\

\parbox[t]{\linewidth}{GRADIENT \citep{huang2022curriculum}\\
(\,curriculum RL\,)}
& Curriculum generation
& Context/task distributions
& Wasserstein barycentre ($\Wone$)
& Sinkhorn algorithm under a $\pi$-contextual distance.\\

\parbox[t]{\linewidth}{HGG \citep{ren2019exploration}\\
(\,goal-conditioned RL\,)}
& Goal generation for exploration
& Joint distribution of initial state and goal
& $\Wone$
& Maximum-weight bipartite matching for goal selection under a Wasserstein objective.\\

\addlinespace[0.35em]
\multicolumn{5}{@{}l@{}}{\textit{Other applications}}\\
\parbox[t]{\linewidth}{\citet{goh2022combining}\\
(\,combinatorial RL\,)}
& OT-based decoder in RL pipeline
& Permutation/assignment structure (e.g.\ tours)
& Sinkhorn-based
& Sinkhorn decoder for non-autoregressive generation in a travelling salesman setting.\\

\parbox[t]{\linewidth}{OTRL \citep{chen2020sequence}\\
(\,sequence generation\,)}
& Regularisation for policy updates
& Sequence embeddings vs.\ reference embeddings
& \textcolor{black}{$\mathrm{OT}_c$ (cosine cost)}
& IPOT algorithm \citep{xie2020fast}.\\

\parbox[t]{\linewidth}{OTPR \citep{sun2025score}\\
(\,diffusion policy (IQL)\,)}
& Action reweighting
& Masked OT between state/action marginals with $Q$-based cost
& $\mathrm{L}_2$-reg.\ masked OT
& OT-derived compatibility weights reweight the diffusion score loss; masked dual formulation.\\

\parbox[t]{\linewidth}{\citet{nkhumise2025studying}\\
(\,RL evaluation\,)}
& Measure exploration / learning efficiency
& State-action occupancy measures across policies
& $\Wone$ ($\mathrm{L}_1$ or $\mathrm{L}_2$ ground norm)
& Uses OT dataset distance (OTDD) \citep{alvarez2020geometric} to quantify policy update effort and effectiveness.\\

\parbox[t]{\linewidth}{WIC \citep{durugkar2021wasserstein}\\
(\,Unsupervised RL\,)}
& Reward shaping for skill-conditioned policies learning
& State visitation distribution vs. start state distribution
& $\Wone$ (time-step metric)
& Duality with a 1-Lipschitz potential. \\

\parbox[t]{\linewidth}{Wasserstein DRO \citet{shen2024wasserstein}\\
(\,Contextual bandits\,)}
& Ambiguity set for robust policy evaluation/learning
& Candidate vs. empirical context distributions
& $\Wtwo$/$\Wassp{\varepsilon}$
& Dual problem. \\

\end{longtable}
\endgroup

\begingroup
\centering
\footnotesize
\setlength{\tabcolsep}{4pt}
\renewcommand{\arraystretch}{1.15}

\begin{longtable}{@{}p{1.7cm} p{1.05cm} p{2.15cm} p{2.5cm} p{3.0cm} p{2.6cm}@{}}

\caption{\textcolor{black}{Experimental results for representative studies. Evidence type is denoted by E for empirical evaluation and T for explicit method-specific theoretical analysis. Labels refer to the cited study, not necessarily to each reported outcome.}}
\label{tab:evidence}\\

\toprule
\textcolor{black}{\textbf{Study (role)}} & \textcolor{black}{\textbf{Evid.}} & \textcolor{black}{\textbf{Environment \& setting}} & \textcolor{black}{\textbf{OT form \& implementation}} & \textcolor{black}{\textbf{Within-study comparison \& outcome}} & \textcolor{black}{\textbf{Limitation noted}}\\
\midrule
\endfirsthead

\toprule
\textcolor{black}{\textbf{Study (role)}} & \textcolor{black}{\textbf{Evid.}} & \textcolor{black}{\textbf{Environment / regime}} & \textcolor{black}{\textbf{OT construction \& solver}} & \textcolor{black}{\textbf{Within-study comparison \& outcome}} & \textcolor{black}{\textbf{Limitation noted}}\\
\midrule
\endhead

\midrule
\multicolumn{6}{r}{\textit{Continued on next page}}\\
\endfoot

\bottomrule
\endlastfoot

\textcolor{black}{OTR \citep{luo2023optimal} (reward labelling)}
& \textcolor{black}{E}
& \textcolor{black}{D4RL locomotion, AntMaze and Adroit; offline, reward-free, one expert demonstration}
& \textcolor{black}{Entropic OT between learner/expert state marginals; Sinkhorn; cosine cost on raw states}
& \textcolor{black}{OTR+IQL against oracle-reward IQL and prior reward-learning baselines on D4RL locomotion (HalfCheetah, Hopper, Walker2d): near-oracle \emph{aggregate} performance from one demonstration, avoiding the severe drops seen in some alternatives; not equal or superior on every task}
& \textcolor{black}{Labels are trajectory-context dependent; state-marginal OT ignores ordering}\\
\addlinespace[0.3em]

\textcolor{black}{\citet{dong2025optimal} (critique)}
& \textcolor{black}{E, T}
& \textcolor{black}{32 benchmarks, offline and online (D4RL locomotion, AntMaze, Adroit); four learners: IQL, ReBRAC, TD3+BC, DrQ-v2}
& \textcolor{black}{Ablates two axes: Wasserstein vs.\ a point-to-set distance (MinDist), and with vs.\ without temporal constraints (SegMatch)}
& \textcolor{black}{\emph{Negative/mixed}: offline, the gains from Wasserstein approximation and temporal constraints largely vanish against a well-tuned learner; online, temporal constraints are essential while the proximity approximation is secondary}
& \textcolor{black}{Temporal alignment does matter offline as the number of demonstrations grows}\\
\addlinespace[0.3em]

\textcolor{black}{SinkhornDRL \citep{sun2024distributional} (critic loss)}
& \textcolor{black}{E, T}
& \textcolor{black}{55 Atari 2600 games, online, fixed model capacity}
& \textcolor{black}{Sinkhorn divergence between particle-based return distributions}
& \textcolor{black}{\emph{Mixed}: against QR-DQN and MMD-DQN, slower convergence early but better aggregate human-normalised score; largest gains with larger action spaces and multi-dimensional rewards}
& \textcolor{black}{$\approx$20\% more compute than MMD-DQN (a different overhead is reported against C51 and QR-DQN); sensitive to $\varepsilon$, iteration count and particle number}\\
\addlinespace[0.3em]

\textcolor{black}{TemporalOT \citep{fu2024robot} (reward labelling)}
& \textcolor{black}{E}
& \textcolor{black}{Nine Meta-world image-based manipulation tasks; \emph{online} imitation from expert video, DrQ-v2 backbone}
& \textcolor{black}{Masked (banded) entropic OT with context-window cosine cost on encoder features}
& \textcolor{black}{Against an online OTR variant, ADS, GAIfO, BC and an oracle task reward: higher success rate without task rewards; an ablation shows both the context cost and the mask contribute}
& \textcolor{black}{Requires online interaction; a temporally informed proxy rather than a full dynamic-OT formulation}\\
\addlinespace[0.3em]

\textcolor{black}{SMMOTIL \citep{sebag2023combining} (reward labelling)}
& \textcolor{black}{E}
& \textcolor{black}{Pendulum-v0 and CartPole-v0, with five experts differing in pole length or mass}
& \textcolor{black}{Sliced multi-marginal OT; closed-form 1D solvers}
& \textcolor{black}{Against concatenating the demonstrations into a single empirical measure: higher mean episodic reward under both length and mass diversity}
& \textcolor{black}{Two low-dimensional control tasks only; slicing approximates the geometry}\\
\addlinespace[0.3em]

\textcolor{black}{WPO / SPO \citep{song2023provably} (trust region)}
& \textcolor{black}{E, T}
& \textcolor{black}{Tabular MDPs; continuous actions via an IQN target}
& \textcolor{black}{Per-state $\Wone$ / Sinkhorn trust region; 1D dual solve, no parametric form assumed}
& \textcolor{black}{Against KL trust regions: monotonic improvement with exact advantages, and global convergence in the tabular setting under a decaying multiplier}
& \textcolor{black}{Tabular guarantee needs finite spaces, non-negative rewards and full initial-state support; the bound degrades by $2\epsilon/(1-\gamma)$ under advantage error}\\
\addlinespace[0.3em]

\textcolor{black}{Q-DOT \citep{omura2025offline} (offline constraint)}
& \textcolor{black}{E}
& \textcolor{black}{D4RL, offline}
& \textcolor{black}{Explicit $\Wtwo$ Brenier map via ICNN; no discriminator}
& \textcolor{black}{Against adversarial Wasserstein penalties: removes the inner min-max loop while retaining $\Wtwo$ geometry}
& \textcolor{black}{Expressiveness limited by the finite ICNN parameterisation rather than by $\Wtwo$ structure itself; minibatch OT bias}\\
\addlinespace[0.3em]

\textcolor{black}{GWIL \citep{fickinger2021cross} (cross-domain IL)}
& \textcolor{black}{E, T}
& \textcolor{black}{Rigid transformation-PointMass Maze, pendulum-cartpole and cheetah-walker; online imitation from one demonstration}
& \textcolor{black}{GW between expert and learner occupancy measures using within-domain Euclidean costs}
& \textcolor{black}{GWIL recovers optimal behaviours for first two tests; with significant mismatched spaces, cheetah $\to$ walker yields forward or backward locomotion depending on the seed}
& \textcolor{black}{Identification only up to isometries can produce suboptimal behaviour; the empirical evaluation contains limited quantitative head-to-head baseline comparisons.}\\
\addlinespace[0.3em]

\textcolor{black}{WR$^2$L \citep{abdullah2019wasserstein} (robustness)}
& \textcolor{black}{E, T}
& \textcolor{black}{MuJoCo (InvertedPendulum, HalfCheetah, Swimmer, Hopper, Walker2D) with perturbed physical parameters}
& \textcolor{black}{Expected $\Wtwo^2$ ball around a reference model; curvature-based approximation}
& \textcolor{black}{Against non-robust training under the same perturbations: improved worst-case return}
& \textcolor{black}{Radius $\delta$ must be chosen by hand; conservatism grows with it}\\

\end{longtable}

\endgroup

\end{document}